\documentclass[10pt,journal,compsoc]{IEEEtran}
\usepackage[utf8]{inputenc} 
\usepackage[T1]{fontenc}    
\usepackage{hyperref}       
\usepackage{url}            
\usepackage{booktabs}       
\usepackage{amsfonts}       
\usepackage{nicefrac}       
\usepackage{microtype}      
\usepackage[dvipsnames, svgnames, x11names]{xcolor}
\usepackage{graphicx}
\usepackage{makecell}
\usepackage{float}
\usepackage{subfig}
\usepackage{amsmath}
\usepackage{stfloats}
\usepackage{tcolorbox}
\usepackage{multirow}
\usepackage{pifont}
\usepackage{enumitem}
\usepackage{amssymb}

\usepackage{enumitem}
\usepackage[misc]{ifsym}

\usepackage[ruled]{algorithm2e}

\ifCLASSOPTIONcompsoc
  \usepackage[nocompress]{cite}
\else
  \usepackage{cite}
\fi
\ifCLASSINFOpdf
\else
\fi

\newif\ifsubmit
\submitfalse

\begin{document}
%
\title{$\mathcal{A^{3}D}$: A Platform of Searching for Robust Neural Architectures and  Efficient Adversarial Attacks}
%
%
%
%

\author{%

  Jialiang Sun, Wen Yao\textsuperscript{\Letter}, Tingsong Jiang \textsuperscript{\Letter},
Chao Li,  Xiaoqian Chen\\
  

\thanks{J. Sun, W. Y, T. Jiang, C. L, and X. C are with Defense Innovation Institute, Chinese Academy of Military Science, Beijing 100071, China.}

\thanks{C. L is also with School of Artificial Intelligence, Xidian
	University, Xian 710071, China.}

\thanks{* indicates equal contributions. \Letter\,indicates corresponding author.}

}
\IEEEtitleabstractindextext{%
\begin{abstract}
  The robustness of deep neural networks (DNN) models has attracted increasing attention due to the urgent need for security in many applications. 
  Numerous existing open-sourced tools or platforms are developed to evaluate the robustness of DNN models by ensembling the majority of adversarial attack or defense algorithms. Unfortunately, current platforms do not possess the ability to optimize the architectures of DNN models or the configuration of adversarial attacks to further enhance the robustness of models or the performance of adversarial attacks. To alleviate these problems, in this paper, we first propose a novel platform called auto adversarial attack and defense (\text{$A^{3}D$}), which can help search for robust neural network architectures and efficient adversarial attacks. In \text{$A^{3}D$}, we employ multiple neural architecture search methods, which consider different robustness evaluation metrics, including four types of noises: adversarial noise, natural noise, system noise, and quantified metrics, resulting in finding robust architectures. Besides, we propose a mathematical model for auto adversarial attack, and provide multiple optimization algorithms to search for efficient adversarial attacks. In addition, we combine auto adversarial attack and defense together to form a unified framework. Among auto adversarial defense, the searched efficient attack can be used as the new robustness evaluation to further enhance the robustness. In auto adversarial attack, the searched robust architectures can be utilized as the threat model to help find stronger adversarial attacks. Experiments on CIFAR10, CIFAR100, and ImageNet datasets demonstrate the feasibility and effectiveness of the proposed platform, which can also provide a benchmark and toolkit for researchers in the application of automated machine learning in evaluating and improving the DNN model robustnesses. 
\end{abstract}

\begin{IEEEkeywords}
Auto Machine Learning, Adversarial Defense, Adversarial Attack, Evolutionary Algorithm.
\end{IEEEkeywords}}

\maketitle

\IEEEdisplaynontitleabstractindextext

%
\IEEEpeerreviewmaketitle

\IEEEraisesectionheading{\section{Introduction}}

\IEEEPARstart{C}{omputer} vision tasks such as the image classification \cite{Remote2022,wang2017residual,lu2007survey}, object detection \cite{QiangInstant,zhao2019object,szegedy2013deep}, semantic segmentation, and object tracking have been applied widely in the real world. However, as the core component, deep neural network (DNN) models have proved to be vulnerable against adversarial examples (AEs), which are obtained by adding specific perturbations to the original images. The AEs can make the trained DNN models output wrong results. With the expanding discovery of AEs, it brings huge potential threats to safety-senstive areas such as self-driving or facial recognition systems. Hence, exploring the robustness of DNN models against AEs is becoming a pressing issue.

Focusing on studying the robustness of DNN models, the related works can be categorized into the adversarial defense and adversarial attack. Adversarial defense aims to make DNN models more robust towards AEs. One of the effective defense methods is designing inherently more robust network architectures. In contrast, the goal of adversarial attacks is to generate specific adversarial noises to fool DNN models, which is an optimization problem. In this way, researchers can obtain the lower bound of prediction to evaluate the robustness of DNN models. Over the past years, the majority of works have been devoted to generating the stronger noises, such as fast gradient sign method (FGSM) \cite{fgsm}, projection gradient descent (PGD) \cite{pgd2018}, and Carlini $\&$ Wagner (CW) attack \cite{cw2017}. Various existing works explore the performance of the adversarial attack from the aspects of changing the attack operation (the updating way of gradient), attack magnitude, iteration numbers, and the loss functions.

Due to the extensive contributions to the adversarial community, it is increasingly difficult to identify the merits of the methods. Thus, to provide a fair comparison environment for various adversarial attack and defense methods, many open-sourced platforms are developed. Papernot et al. \cite{papernot2016technical} proposed CleverHans, a library that provides adversarial training techniques using Fast Gradient Sign Method (FGSM) or Jacobian-based Saliency Map Attacks (JSMA). Rauber et al. \cite{rauber2017foolbox} deveolped the foolbox. Compared with CleverHans, they provide more adversarial attack algorithms to evaluate the robustness of DNN models. Nicolae et al.~\cite{nicolae2018adversarial} proposed adversarial robustness toolbox (ART). Compared with the previous works that only focus on image classification, they also provide the adversarial attack algorithms for object detection and speech recognition. Dong et al. \cite{dong2020benchmarking} provided the benchmark of adversarial robustness on image classification. There are also some adversarial robustness platforms in other areas, such as the natural language processing \cite{zeng2020openattack} and graph data \cite{li2020deeprobust}.

Despite the fact that various platforms mentioned above have been developed to evaluate the performance of existing adversarial attack or defense methods, they can not possess the ability to optimize the manually designed adversarial attack or defense method to further improve their own performance. To alleviate this problem, the application of the auto machine learning (AutoML)  technique has been becoming one of the effective approaches, which can realize the purpose of automatical hyperparameters optimization, model selection, or algorithms design.


The process of AutoML includes defining the search space, conducting the search strategy, and determining the evaluation metric. Given the evaluation metric determined according to different tasks, different search strategies are performed to search for the best configuration in the predefined search space. In AutoML for robust architectures, the popularly adopted technique is called neural architecture search (NAS). The model search space mainly follows that of differentiable architecture search (DARTS) \cite{liu2018darts}, a kind of cell-based neural architecture. The search strategy follows the variants of DARTS, such as PC-DARTS \cite{xu2019pc}. By defining the robustness evaluation metrics such as Jacobian matrix \cite{Differentiable2021} or Hessian matrix \cite{AdvRush2021}, the robust architecture could be sought. Their methods are denoted as DSRNA and AdvRush, respectively. In AutoML for the efficient adversarial attack, the search space in composite adversarial attack (CAA) \cite{Composite2021} is the attacker operation, attack magnitude, and optimization step. Yao et al. \cite{Automated2021} further adds the transformation of data and the target of the attacker into the search space. Their search strategies are evolutionary optimization or random search algorithms. 

\begin{table*}[t]
	\renewcommand\arraystretch{1.5}
	\scriptsize
	
	\centering
	
	\caption{The search space, search strategy and performance evaluation in our proposed \text{$A^{3}D$} framework.}
	
	\label{Tab4}
	\setlength{\tabcolsep}{4mm}{
		\begin{tabular}{ccccccccccc}
			
			\toprule
			
			Type & Search Space	&  Search Strategy 	&  Performance Evaluation  \\		
			\midrule		
			\multirow{5}{*}{Adversarial Attack}& Attacker Operation   &    \textbf{Single-objective}:   &               \\				
			& Attacker Loss Function   & PSO, DE, Local search        & Robustness Accuracy           \\		
			& Attacker Magnitude  & \textbf{Multi-objective}:        &   Time Cost          \\		
			
			&\text { Attacker Iteration Number}   &NSGA-II,       &  \\
			&\text { Attacker Restart}   &Random Search      &  \\
			
			\midrule
			\multirow{8}{*}{Adversarial Defense}& None   & \textbf{Differentiable}:         &\textbf{Adversarial Noise}:            \\	
			& max$\_$pool$\_$3x3  & DARTS, PC-DARTS,        & FGSM, PGD, Auto adversarial attack             \\		
			& 	avg$\_$pool$\_$3x3  & NASP, FairDARTS,        & \textbf{System Noise}:            \\
			& skip$\_$connect  & SmoothDARTS         & ImageNet-S          \\
			& sep$\_$conv$\_$3x3   & \textbf{Non-differentiable}:        & \textbf{Natural Noise}:            \\			
			& sep$\_$conv$\_$5x5   &  Random Search,        & ImageNet-C, ImageNet-P          \\
			& dil$\_$conv$\_$3x3   & DE,         & \textbf{Quantific Metric}:            \\
			&dil$\_$conv$\_$5x5  &    Weight Sharing based Random Search      & Hessian matrix, Jacobian matrix            \\

			\bottomrule		
	\end{tabular}}	
\end{table*}

Though some works about AutoML for adversarial attack and defense have been developed, there are still two main problems released openly. On the one hand, the settings of the search space, search strategy, and evaluation metrics have a great influence on the final performance of the searched adversarial attack or architectures. However, these three core parts in current methods differ greatly, which brings huge difficulty to the fair evaluation of the search algorithms. On the other hand, current researches about AutoML for adversarial attack and adversarial defense are separate, resulting in the problem that the searched attack does not perform well in more robust architectures, or the searched robust architectures do not perform well in stronger attacks. Hence, it is necessary to find more robust architectures under the searched stronger attacks or find more efficient attacks under the searched architectures.

To alleviate the problems mentioned above, we propose a novel framework called auto adversarial attack and defense (\text{$A^{3}D$}). It enables combining NAS for robust network architectures and evolutionary search for efficient attack schemes together to form a unified framework, as shown in Fig~\ref{add}. The introduction of the detailed functions of our proposed framework can be seen in Table \ref{Tab4}. In $A^{3}D$, our contributions can be summarized as follows:

1) In auto adversarial defense, we benchmark eight representative NAS methods, including the differentiable and non-differentiable based, for robust model architectures under comprehensive robustness evaluations. Specially, we include four types of evaluations: adversarial noise, natural noise, system noise, and quantified metrics as the robustness evaluations. In addition, we design the corresponding loss functions according to the characteristic of robustness evaluations and NAS methods. 

2) In auto adversarial attack, we benchmark four evolutionary algorithms to search for efficient adversarial attacks, including single-objective and multi-objective optimization, which can find the near-optimal configuration of adversarial attacks that possess the lower robust acuuracy and less evaluation cost.  

3) We make the first step towards combining the auto adversarial attack and NAS methods for robust architectures together, forming a unified framework. The searched adversarial attack can help search more robust architecture, while the searched model helps generate adversarial examples with better transferability.

The remainder of this paper is organized as follows. Section \ref{sec2} introduces the preliminary knowledge about adversarial attack and defense, auto machine learning, and AutoML for adversarial attack and defense. Section \ref{sec3} introduces our proposed framework of auto adversarial attack and defense. Section \ref{sec4} elaborates on the experimental settings and results. Section \ref{sec5} draws the conclusions.

\section{Background and Related Work}\label{sec2}

In this section, the preliminary about adversarial attack and defense, auto machine learning, and AutoML for adversarial attack and defense are introduced, respectively.
\subsection{Adversarial Attack and Defense}
The aim of adversarial attacks is to deceive the DNN models by adding the specific perturbation to the original image (also called the clean image). Denote the original image $x$, the perturbation $\Delta x$,  the magnitude of the added perturbation$\epsilon$, the DNN model $\mathcal{F}$, the classification label $y$, the final adversarial example $x_{adv}$. The whole formula of adversarial attacks can be expressed as Eq. \ref{attack}, which aims to find the near-optimal $\Delta x$ to maximize the loss values of the prediction of DNN models.
\begin{equation}
\begin{array}{l}
\underset{\Delta x}{\arg \max }  \mathcal{L}\left(x_{adv}, y ; \mathcal{F}\right)\\
x_{a d v}=x+\Delta x \\
\text { s.t. }\|\Delta x\| \leq \varepsilon.
\end{array}
\label{attack}
\end{equation}


To address above optimization problem in adversarial attacks, many works have been developed over the past years. At the earliest time, Goodfellow et al. \cite{goodfellow2014explaining} first proposed fast gradient sign method (FGSM), which is a single-step updating method, as shown in Eq. \ref{fgsm}.

\begin{equation}
x_{a d v}=\operatorname{clip}_{[0,1]}\left\{x+\epsilon \cdot \operatorname{sign}\left(\nabla_{x} \mathcal{L}(x, y ; \mathcal{F})\right)\right\}.
\label{fgsm}
\end{equation}

Madry et al. \cite{pgd2018} further proposed projected gradient descent (PGD), which obtains the near-optimal perturbation by multi-step iterations, as illustrated in Eq. \ref{pgd}.
\begin{equation}
x_{l+1}=\operatorname{project}\left\{x_{l}+\epsilon_{step} \cdot \operatorname{sign}\left(\nabla_{x} \mathcal{L}\left(x_{l}, y ; \mathcal{F}\right)\right)\right\}
\label{pgd}
\end{equation}
where the initial image can be original image or the perturbed image by random initialization.$\epsilon_{step}$ stands for the perturbation magnitude at each iteration.

To improve the performance of adversarial attacks, Carlini and Wagner \cite{carlini2017towards} modified the loss function of the adversarial attack, which adopts the distance of logit values between the label $y$ and the second most-likely class, as presented in Eq. \ref{cw}.

\begin{equation}
\mathcal{L}_{C W}(x, y ; \mathcal{F})=\max \left(\max _{i \neq y}\left(\mathcal{F}(x)_{(i)}\right)-\mathcal{F}(x)_{(y)},-\kappa\right)
\label{cw}
\end{equation}
where $\kappa$ establishes the margin of the logit of the adversarial class being larger than the logit of runner-up class \cite{Composite2021}.

Apart from these adversarial attacks, other work such as MultiTargeted (MT) attack \cite{gowal2019alternative}, Momentum Iterative
(MI) attack \cite{dong2018boosting}, and Decoupled Direction and Norm (DDN) attack \cite{rony2019decoupling} are also further developed. To this end, it can be concluded that existing manually-designed adversarial attack methods needs to preset the magnitude of perturbation, attack iteration number, and the loss functions.

To improve the robustness of DNN models on the adversarial examples generated by the adversarial attack, various defense methods are also proposed over the past years. The adversarial training technique is one of the most notable defense methods, which was first introduced by Goodfellow et al. \cite{goodfellow2014explaining}. It can be modeled into a dual optimization problem consisting of the generation of adversarial examples and the training of the parameters of DNN models based on the adversarial examples. Besides, much effort has been also devoted to developing other defense methods, such as the defensive distillation \cite{papernot2016distillation}, dimensionality reduction \cite{bhagoji2017dimensionality}, input transformations \cite{guo2017countering}, and activation transformations \cite{dhillon2018stochastic}.

With the rapid development of the defense community, it tends to be more difficult to evaluate the robustness of DNN models through adversarial attacks. For one thing, the time cost of attacking the models is relatively too long. For another thing, the manually-designed attack algorithms are easily trapped in the local optimum. For example, one manually-designed adversarial attack algorithm is effective in one defensed model, while ineffective in another defensed model. Therefore, given the defensed model, it is necessary to realize the automatical search for the near-optimal adversarial attack that can obtain lower robust accuracy with less time cost, which can not only release the burden of the person but also improve the performance of adversarial attack algorithms.

\subsection{Auto Machine Learning}

As a branch of AutoML, the purpose of neural architecture search is to automatically search for better neural architectures for the specific task. It could be modeled into a bi-level optimization problem. The inner problem is a gradient-based weight optimization problem, while the outer one is the model architecture parameters optimization problem. The mathematical model of NAS could be modeled as Eq. \ref{art}. 

\begin{equation}
\begin{array}{ll}
\min _{\alpha} & \mathcal{L}_{v a l}\left(w^{*}(\alpha), \alpha\right) \\
\text { s.t. } & w^{*}(\alpha)=\operatorname{argmin}_{w} \mathcal{L}_{\text {train }}(w, \alpha).
\end{array}
\label{art}
\end{equation}

where $\alpha$ denotes the searched architecture, $w^{*}$ stands for the optimal weights for the selected architecture. To solve the above optimization problem, if each model architecture is to be trained from scratch in the process of searching, the computational burden could not be born. Then the approximate one-shot search algorithm DARTS is proposed, which is presented in Algorithm \ref{alg1}. Based on this, a series of improved one-shot NAS algorithms are developed. For instance, PC-DARTS \cite{xu2019pc} and NASP \cite{yao2020efficient} are proposed to reduce the calculating memory, enhance the searching performance and reduce the model parameters.

Apart from NAS, AutoML also include the hyper-parameters optimization.

%


\begin{algorithm}[!ht]
	\caption{Differentiable Architecture Search (DARTS)}
		\label{alg1}
	\LinesNumbered
	\KwIn{ 
		The initial weight ${w}$, initial architecture $\alpha$, search space,  maximum epoch of searching $T$, the number of warm$\_$epoch $T_{warm}$.
	}
	\KwOut{The searched near optimal network architecture  $\alpha$.}
	
	 Create the mixed operation according to the defined search space; \\ 
	  Training epoch $t=1$; \\ 
	\While{t<T}{
		\If{$t$< $T_{warm}$}{
		Update ${w}$ by calculating $\nabla_{w} \mathcal{L}_{\text {train }}(w, \alpha)$;\\
	}
		\Else{
		Update the architecture  $\alpha$ by calculating gradient $\nabla_{\alpha} \mathcal{L}_{v a l}\left(w-\xi \nabla_{w} \mathcal{L}_{\text {train }}(w, \alpha), \alpha\right)$;\\
		 Update ${w}$ by calculating $\nabla_{w} \mathcal{L}_{\text {train }}(w, \alpha)$;\\
	}
     $t=t+1$;\\
	}
\textbf{return} $\alpha$.
	
\end{algorithm}

\subsection{AutoML for Adversarial Attack and Defense}

\begin{figure*}[t]
\centering
\includegraphics[scale=0.87]{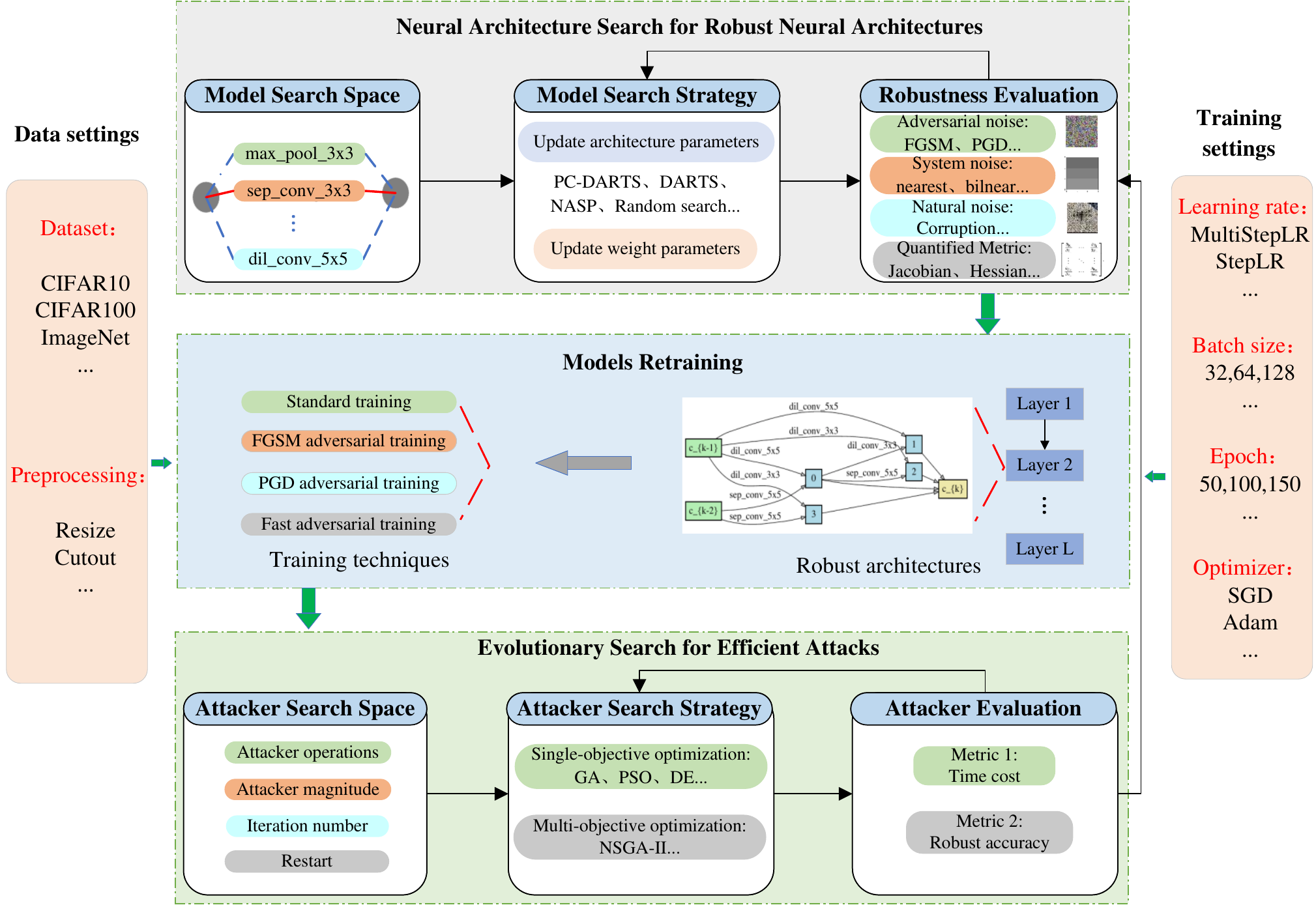}
\caption{The illustration of $A^{3}D$ framework.The top module is neural architecture search for robust neural architectures. When the robust evaluation metric is fixed, this module can automatically search for robust model architectures. The middle module is to retrain the searched architectures using different training techniques, such as PGD adversarial training. The bottom module is evolutionary search for efficient attack schemes. When the searched model is trained, the near-optimal adversarial attack scheme can be found by evolutionary algorithms. The searched attack schemes can also be used as new robustness evaluations in adversarial noise.}
\label{add}
\end{figure*}

In AutoML for the adversarial attack, Fu et al. \cite{automated2021Qi} tried to find the decision-based adversarial attack strategy based on the AutoML technique. Mao et al. \cite{Composite2021} proposed the composite adversarial attack (CAA). They searched for the near-optimal adversarial attack scheme by NSGA-II, which can possess a higher attack success rate compared with the single attack or ensemble attack.  Francesco et al. \cite{ReliableFrancesco}  proposed the ensemble attack of four types of attacks, which can also generate superior attack performance. Yao et al. \cite{Automated2021} further proposed the method for jointly searching for the combination of attack loss functions, attack operations, and hyperparameters.

In AutoML for the adversarial defense, existing methods can be categorized into the differentiable and non-differentiable based. The non-differentiable based methods evaluate the robustness of each architecture using adversarial attacks like PGD during the search process. The processes of searching and evaluating the architectures are separate, which is relatively time-consuming. Guo et al. \cite{whencvpr} applied the NAS method in searching for robust network architectures towards PGD. Other similar works are as \cite{liu2021multi,vargas2019evolving}, which utilize the evolutionary algorithms to find more robust architectures. In contrast, the differentiable methods adopt the differentiable metric into differentiable architecture search methods, which greatly accelerates the search process. Hosseini et al. \cite{Differentiable2021} proposed DSRNA, which combines the quantified metrics into the differential architecture search to efficiently find robust model architectures. Mok et al. \cite{AdvRush2021} proposed AdvRush, which finds more robust architectures based on the eigenspectrum of the Hessian matrix.
\section{The Framework of Auto Adversarial Attack and Defense}\label{sec3}

\begin{figure*}[h]
	\centering
	\subfloat[Clean]{\includegraphics[width=1.3in]{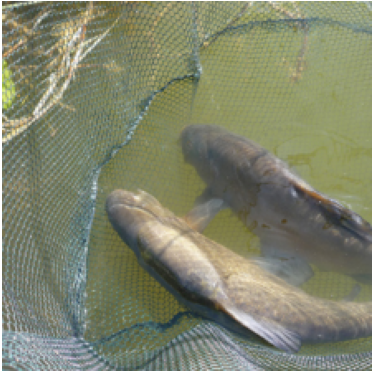}%
	}
	\hfil
	\subfloat[Frost]{\includegraphics[width=1.3in]{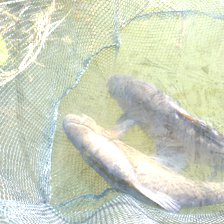}%
	}
	\hfil
	\subfloat[Snow]{\includegraphics[width=1.3in]{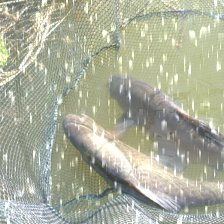}%
	}
	\hfil
	\subfloat[Fog]{\includegraphics[width=1.3in]{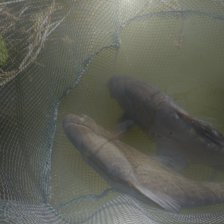}%
	}
	\hfil
	\subfloat[Contrast]{\includegraphics[width=1.3in]{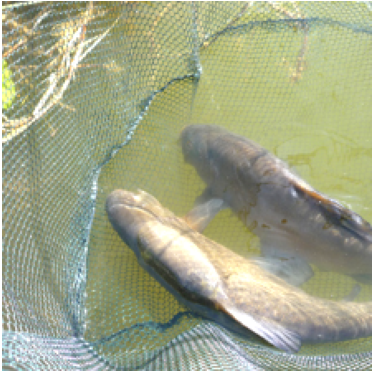}%
	}
	
	\subfloat[FGSM-LinfAttack]{\includegraphics[width=1.3in]{figure/contrast.png}%
	}
	\hfil
	\subfloat[PGD-LinfAttack]{\includegraphics[width=1.3in]{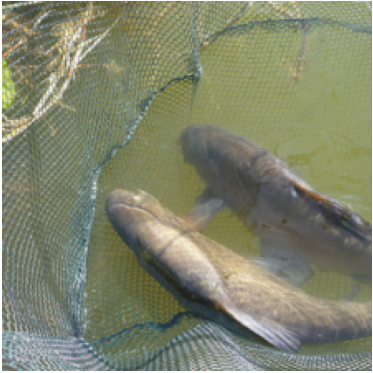}%
	}
	\hfil
	\subfloat[pil-opencv$\_$bilinear]{\includegraphics[width=1.3in]{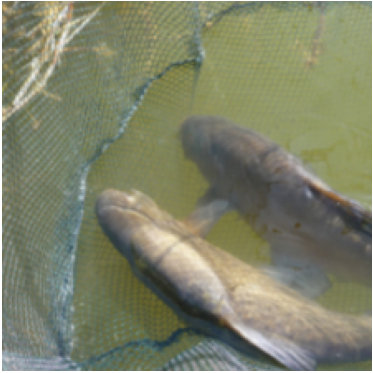}%
	}
	\hfil
	\subfloat[opencv-pil$\_$hanmming]{\includegraphics[width=1.3in]{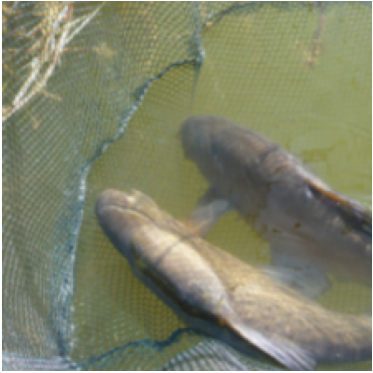}%
	}
	\hfil
	\subfloat[Jacobian]{\includegraphics[width=1.3in]{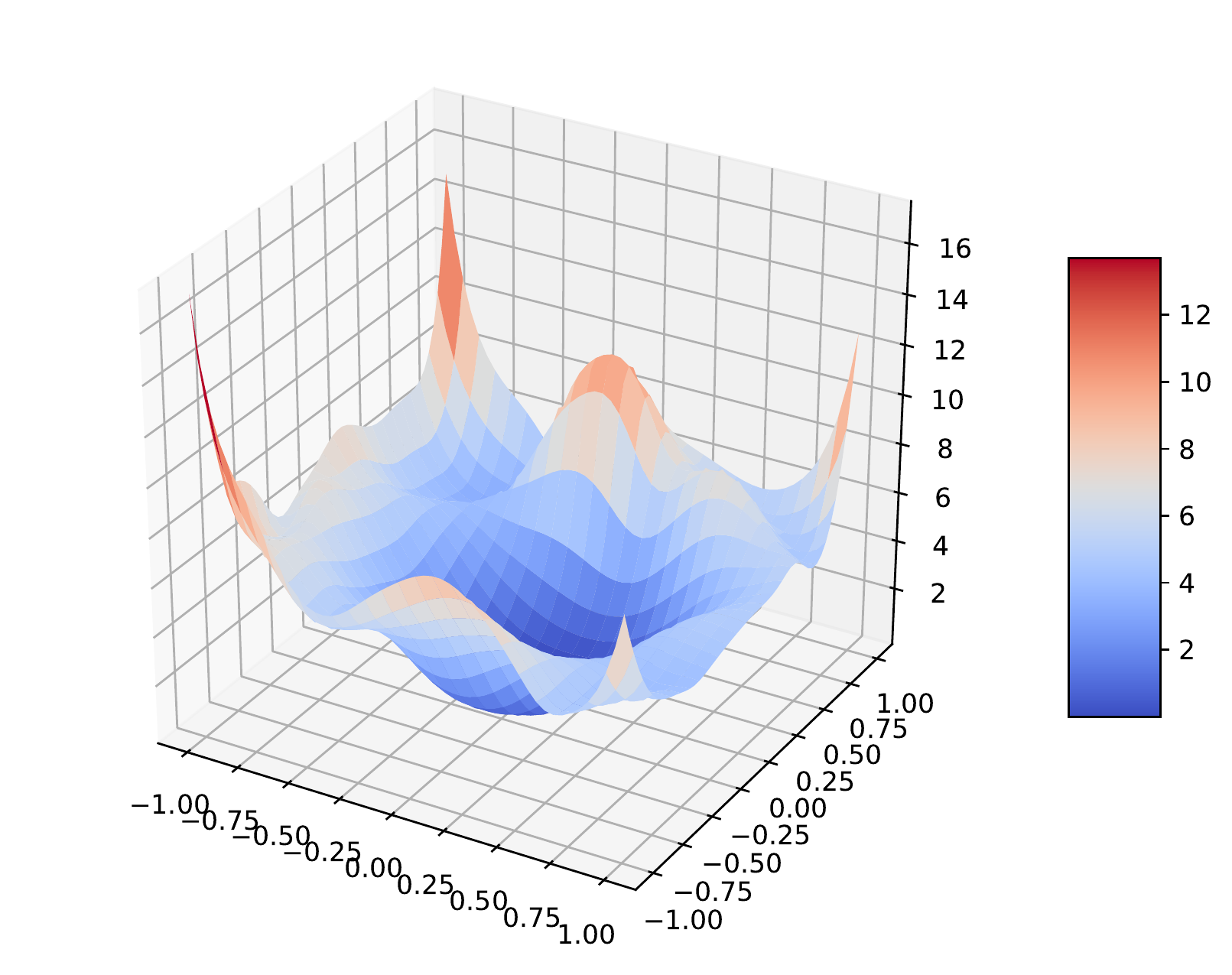}}
	\caption{The visualization of different robustness evaluations. (a) is the original image, (b)-(e) are the images with natural noises, (f)-(g) are the images with system noises, (h)-(i) are the images with adversarial noises, and (j) is the calculation of quantified metrics, such as Jacobian. }
	\label{noise}
\end{figure*}

The overview of our proposed framework is illustrated in Fig~\ref{add}. The top module is to automatically search for robust network architectures when different robustness evaluation metrics are considered. The bottom module is evolutionary algorithms for searching of efficient adversarial attack schemes. The adversarial noise that is generated by the searched auto adversarial attack can also be combined into the robustness evaluation in NAS for robust architectures as the more reliable evaluation. In the following, the proposed \text{$A^{3}D$} framework is elaborated in detail.

\subsection{AutoML for Adversarial Defense}
In this section, various NAS methods to search for robust architectures using different evaluations are introduced in detail.

The mathematical model of searching for robust architectures can be described as Eq.~\ref{robust}.
\begin{equation}
\begin{array}{ll}
\min\limits_{\alpha} & \mathcal{L}_{val}\left(w^{*}(\alpha), \alpha\right) + R\left(w^{*}(\alpha), \alpha\right) \\
\text { s.t. } & w^{*}(\alpha)=\operatorname{argmin}_{w} \mathcal{L}_{\text {train }}^{\text {adv }}(w, \alpha),
\end{array}
\label{robust}
\end{equation}
where $\alpha$ denotes the searched architecture, $w^{*}$ stands for the optimal weights for the selected architecture, $R$ represents the robustness evaluation, which has multiple forms. For instance, if we take the robust accuracy under adversarial attacks as the robustness evaluation, $R$ can be defined as the loss of the images produced by adversarial attacks and the labels. If we take the gradient-based estimation as the robustness evaluation such as the Jacobian metric, $R$ can be defined as the value of Jacobian metric. In this way, we can find more robust architectures by minimizing the $R$ value.
\subsubsection{Robustness Evaluation}

In this module, we introduce four types of robustness evaluations, including adversarial noises, natural noises, system noises and quantified metrics. These robustness evaluations can be easily combined into existing NAS frameworks to help search for robust architectures. 
\begin{itemize} 
\item Adversarial Noises
\end{itemize} 
In adversarial noises, we ensemble the majority of classical $l_{p}$ norm based adversarial attack algorithms to generate adversarial noises. The magnitude of the $l_{p}$ norm of these noises is bounded with the certain threshold. The accuracy on the generated adversarial noise is utilized to evaluate the robustness of models , also called robust accuracy, which is expressed as Eq.~\ref{eq1}.

\begin{equation}
R_{A}\left(w^{*}(\alpha), \alpha, x_{i}^{(\mathrm{adv})}\right ),
\label{eq1}
\end{equation}%
where $x_{i}^{(\mathrm{adv})}$ is the adversarial noise generated by adversarial attack algorithms such as FGSM and PGD. For the given DNN model, the higher accuracy under adversarial noise means the more robustness. More details about $l_{p}$ norm based adversarial attacks provided by us can be seen in the Section \ref{autoattack}.

\begin{itemize} 
\item Natural Noises
\end{itemize}

Different from adversarial noises generated by gradient-based attack algorithms, natural noises do not limit the magnitude of perturbations. However, the requirement of adding the natural noises is that the semantic of original images can not be changed. Hence, natural noises can also be called semantic perturbations, which aim to make the final image look natural but can fool the DNN models. Over the past yeas, much effort has been devoted to natural noises in the image classification task. In this paper, we provide the common corruptions provided by Hendrycks et al. \cite{chen2020anti} to evaluate the robustness of models on natural noises, which includes the datasets called CIFAR10-C and ImageNet-C. Common corruptions provide the benchmark datasets of examples that can simulate the natural scenes, including the brightness, contrast, defocus blur, fog, frost, gaussian blur.


\begin{equation}
R_{N}\left(w^{*}(\alpha), \alpha, x_{i}^{(\mathrm{nat})}\right ),
\label{eq2}
\end{equation}%
where $x_{i}^{(\mathrm{nat})}$ is the natural noise sample.

\begin{itemize} 
\item System Noises
\end{itemize}
System noises refers to the inaccuracy inherent to a system due to the inconsistent implementations of decoding and resize, such as ImageNet-S \cite{wang2021real}. Taking the opencv-pil$\_$bilinear as an example, the original images are decoded by opencv tools and resized by pil$\_$bilinear, then the generated images can also cause the dropping of the accuracy of models. Apart from the opencv-pil$\_$bilinear, ImageNet-S also provides the opencv-pil$\_$bilinear, opencv-pil$\_$hanmming, pil-opencv$\_$bilinear, pil-opencv$\_$cubic, pil-opencv$\_$nearest, pil-pil$\_$bilinear, pil-pil$\_$cubic, and pil-pil$\_$nearest.

\begin{equation}
R_{S}\left(w^{*}(\alpha), \alpha, x_{i}^{(\mathrm{sys})}\right ),
\label{eq3}
\end{equation}%
where $x_{i}^{(\mathrm{sys})}$ is the system noise sample.

\begin{itemize} 
\item Quantified  metrics
\end{itemize}
Apart from the three above game-based robustness evaluation metrics, we also include quantified metrics such as the $F$ norm of the Jacobian and Hessian matrix of the input data. These quantified metrics can evaluate the smoothness of the landscape of DNN models. The smaller quantified values mean the more smooth landscapes, resulting in the predictions of DNN models can not be easily changed by adding the perturbations. These metrics do not rely on generating noises on original test data, which can be easily combined into differentiable neural architecture search methods \cite{Differentiable2021,AdvRush2021}.

The Frobenius ($F$) norms of Jacobian matrix and Hessian matrix are calculated as Eq. \ref{eq5d5} and Eq. \ref{eq5}.


\begin{equation}
R_{J}\left(w^{*}(\alpha), \alpha, x_{i}\right ) = \|J({x_{i}})\|_{\mathrm{F}}^{2},
\label{eq5d5}
\end{equation}%
where $J$ denotes the Jacobian matrix of the output vector of 10 classifier with respect to input data, $i$ represents the number of input sample.

\begin{equation}
R_{H}\left(w^{*}(\alpha), \alpha, x_{i}\right ) = \left\|H_{\mathrm{std}}({x_{i}})\right\|_{\mathrm{F}}^{2},
\label{eq5}
\end{equation}%
where $H_{\mathrm{std}}$ denotes the Hessian matrix of loss with respect to input data.

In general, much effort has been devoted to evaluate the robustness of DNN models. In this robustness evaluation module, we provide four popular types of evaluations. Some of them are visualized in Fig~.\ref{noise}.

\subsubsection{NAS for Robust Neural Architectures}

In this module, we provide variety of NAS methods, including differentiable and non-differentiable strategires to search for robust architectures. In detail, these methods includes DARTS, PC-DARTS, NASP, SmoothDARTS, FairDARTS, Random Search, DE, and Weight Sharing based Random Search. Due to the comprehensive and varied nature of robustness evaluations, these evaluations can not be combined directly into NAS according to the same formula. 

To ensure the fair comparison environment, as illustrated in Table \ref{NASSP}, the search space of all NAS algorithms is the same, including none,  max$\_$pool$\_$3x3, avg$\_$pool$\_$3x3, skip$\_$connect, sep$\_$conv$\_$3x3, sep$\_$conv$\_$5x5, dil$\_$conv$\_$3x3, and dil$\_$conv$\_$5x5. The aim of NAS is to find the near-optimal combination of these eight operations to form a whole neural network. 
\begin{table}[h]
	\centering
	\caption{The search space of NAS.}
	\renewcommand{\arraystretch}{1} 
	\begin{tabular}{ccc}
		\midrule
		none &  max$\_$pool$\_$3x3\\
		avg$\_$pool$\_$3x3  & skip$\_$connect\\
		sep$\_$conv$\_$3x3 & sep$\_$conv$\_$5x5 \\
		dil$\_$conv$\_$3x3    & dil$\_$conv$\_$5x5   \\
		\midrule
	\end{tabular}
	\label{NASSP}
\end{table}


As for the differentiable search methods, including DARTS, PC-DARTS, NASP, FairDARTS, and SmoothDARTS, to realize the process of differentiable search under specific adversarial noise, we need to design the specific loss functions.  

For the adversarial noises, we take the adversarial loss into the total loss functions in the search process, which is as Eq.~\ref{eee}.

\begin{equation}
\mathcal{L}\left(w^{*}(\alpha), \alpha ; \mathcal{D}_{\mathrm{val}}\right)+\gamma \mathcal{L}\left(w^{*}(\alpha), \alpha ; \mathcal{D}_{\mathrm{adv}}\right),
\label{eee}
\end{equation}
where $\gamma$ is the coefficient to control the balance of clean accuracy and robust accuracy. $\mathcal{D}_{\mathrm{adv}}$ is the adversarial example of $\mathcal{D}_{\mathrm{val}}$ generated by specific adversarial attack.

For the fixed noise data such as the system noise and natural noise, we directly add them into the clean dataset. The mixture dataset is denoted as $\mathcal{D}_{\mathrm{mix} \_ \mathrm{val}}$.  The total loss function is as Eq.~\ref{mix}.

\begin{equation}
\mathcal{L}\left(w^{*}(\alpha), \alpha ; \mathcal{D}_{\mathrm{mix} \_ \mathrm{val}}\right).
\label{mix}
\end{equation}

For the quantified metrics, we just directly consider them as the regularizer of the original loss. Taking the Jacobian matrix as an example, the total loss is as Eq.~\ref{eq111}. 

\begin{equation}
\mathcal{L}\left(w^{*}(\alpha), \alpha ; \mathcal{D}_{\mathrm{val}}\right)+\gamma R_{J}\left(w^{*}(\alpha), \alpha, x_{i}\right ) .
\label{eq111}
\end{equation}

These total losses are combined into Eq.~\ref{art} to realize the purpose of automatically search for robust architectures.

As for the non-differentiable NAS methods, including Random Search, Weight Sharing based Random Search, and DE, each sampled architecture is trained by 7-step PGD adversarial training and evaluated using above mentioned robustness evaluation metrics in the search stage.

We take three NAS methods, including DARTS, DE, and Weight Sharing based Random Search as the examples to show the process of searching for robust neural architectures, which is presented in Algorithm \ref{alg2}, Algorithm \ref{alg3}, and Algorithm \ref{alg4}, respectively. Other algorithms about NAS for robust architectures can be seen in the Appendix.

\begin{algorithm}[!ht]	
	\caption{Differentiable Architecture Search for Robust Architectures}
		\label{alg2}
	\LinesNumbered
	\KwIn{ 
		The initial weight ${w}$, initial architecture $\alpha$, search space,  maximum epoch of searching $T$, the number of warm$\_$epoch $T_{warm}$, robustness  evaluation metric.
	}
	\KwOut{The searched near optimal network architecture  $\alpha$.}
	
	Create the mixed operation according to the defined search space; \\ 
	Determine the $\mathcal{L}_{\text {train }}$ according to the robustness metric by Eq.~\ref{eee}, Eq.~\ref{mix} or Eq.~\ref{eq111};\\
	Training epoch $t=1$; \\ 
	\While{t<T}{
		\If{$t$< $T_{warm}$}{
			Update ${w}$ by calculating $\nabla_{w} \mathcal{L}_{\text {train }}(w, \alpha)$;\\
		}
		\Else{
			Update the architecture  $\alpha$ by calculating gradient $\nabla_{\alpha} \mathcal{L}_{v a l}\left(w-\xi \nabla_{w} \mathcal{L}_{\text {train }}(w, \alpha), \alpha\right)$;\\
			Update ${w}$ by calculating $\nabla_{w} \mathcal{L}_{\text {train }}(w, \alpha)$;\\
		}
		$t=t+1$;\\
	}
	\textbf{return} $\alpha$.
	
\end{algorithm}

\begin{algorithm}[!ht]
	\caption{DE based Search for Robust Architectures}
	\label{alg3}
	\LinesNumbered
	\KwIn{ 
		The initial weight ${w}$, initial architecture $\alpha$,search space, training epoch $t$.
	}
	\KwOut{The searched near optimal network architecture  $\alpha$.}
	
	Encode the individual according to the defined search space; \\ 
Determine the $\mathcal{L}_{\text {train }}$ according to the robustness metric by Eq.~\ref{eq1}, Eq.~\ref{eq2}, Eq.~\ref{eq3}, Eq.~\ref{eq5d5} and Eq.~\ref{eq5} ;\\
$\textbf{P}$ $\gets$ generate the initial population according to the encoding;\\
Searching epoch $t=1$; \\ 
\While{t<T}{
	
			$\textbf{P}_{m}$$\leftarrow$ Population mutation;\\
	$\textbf{P}_{c}$  $\leftarrow$ Population crossover based on $\textbf{P}_{m}$;\\
	//  \textbf{Population selection}\\
	Individual counter number $j = 1$;\\
	\For{$j <= n$}{
		\If {fitness $(\textbf{P}_{c})[j]$  $>$ fitness$(\textbf{P})[j]$}
		{$\textbf{P}[j]$ = $\textbf{P}_{c}[j]$;}
		$j=j+1$;
	}
	
	Obtain the individual $\alpha$ with the best robustness $\leftarrow$ $argmax$($fitness$(\textbf{P}))\\
	$t=t+1$;\\
}
	\textbf{return} $\alpha$.
	
\end{algorithm}

\begin{algorithm}[!ht]	
	\caption{Weight Sharing based Random Search for Robust Architectures}
	\label{alg4}
	\LinesNumbered
	\KwIn{ 
		The initial weight ${w}$, initial architecture $\alpha$, search space,  maximum epoch of searching $T$, the number of warm$\_$epoch $T_{warm}$, robustness  evaluation metric.
	}
	\KwOut{The searched near optimal network architecture  $\alpha$.}
	
	$S$ $\gets$ Create the mixed operation according to the defined search space; \\ 
	Determine the $\mathcal{L}_{\text {train }}$ according to the robustness metric by Eq.~\ref{eq1}, Eq.~\ref{eq2}, Eq.~\ref{eq3}, Eq.~\ref{eq5d5} and Eq.~\ref{eq5} ;\\
	Searching epoch $t=1$; \\ 
	\While{t<T}{
		
	 \text{arch $\gets$ sample\_uniformly\_at\_random($S$)};\\
		{$S$.update\_weights\_of\_single\_architecture(arch, $w$, $\nabla_w \mathcal{L}_{advtrain}$)} by the training technique;\\
		$t=t+1$;\\
	}
	\While{i $<$ 1000}{
		 \text{arch\_samples $\gets$ sample\_uniformly\_at\_random(	$S$)};
}
 arch $\in$ arch\_samples with the best robustness evaluation;\\
	\textbf{return} $\alpha$.
	
\end{algorithm}

\subsection{AutoML for Adversarial Attack}\label{autoattack}

In this section, given the trained models, the auto adversarial attack aims to generate stronger noises to evaluate the robustness of the models. Various search stratgies are developed to search for near-optimal aduto adversarial attack, including single-objective optimization and multi-objective optimization.

\subsubsection{Attacker Evaluation}
The performance of the adversarial attack is evaluated using the robust accuracy of the model, which is calculated as Eq. \ref{eq1}. Besides, we also include the time cost of the evaluation of an attack sequence on total test samples as another metric.

\subsubsection{EA for attacks}
The single-objective optimization mathematical model of auto adversrial attack:
\begin{equation}
\begin{array}{ll}
\min\limits_{\textbf{x}}  & R(\textbf{x}) \\
\text { s.t. } & 1 \leq \textbf{x}(4k-3)) \leq n_{\text {attacker }} \\
& 1 \leq \textbf{x}(4k-2)) \leq n_{\text {loss }} \\
& 1 \leq \textbf{x}(4k-1) \leq n_{e p s} \\
& 1 \leq \textbf{x}(4k) \leq n_{\text {step }} \\
\end{array}
\label{eqsingle}
\end{equation}
where $k$ denotes the maximum number of the attack cell in one attack sequence,  $n_{\text {attacker }}$ stands for the number of the attacker operation in the search space,  $n_{\text{eps}}$ denotes the number of the attacker magnitudes in the search space, $n_{\text{step}}$ denotes the number of the attacker iteration number in the search space. $R(\textbf{x})$ is the robust accuracy of the trained model.

The multi-objective optimization mathematical model of auto adversrial attack:
\begin{equation}
\begin{array}{ll}
\min\limits_{\textbf{x}} & F(x)=\left[R(\textbf{x}), T(\textbf{x})\right]^{T} \\
\text { s.t. } & 1 \leq \textbf{x}(4k-3)) \leq n_{\text {attacker }} \\
& 1 \leq \textbf{x}(4k-2)) \leq n_{\text {loss }} \\
& 1 \leq \textbf{x}(4k-1) \leq n_{e p s} \\
& 1 \leq \textbf{x}(4k) \leq n_{\text {step }} \\
\end{array}
\end{equation}
where $T(\textbf{x})$ stands for the time cost of evlauting the performance of the attack sequence.

Our definition of the search space includes the attacker operation, attacker loss functions, attacker magnitude, attacker iteration number, and attacker restart. The search space of the attacker operations is illustrated in Table~\ref{attackerop}. Two types of adversarial attacks, including the $l_{2}$ and $l_{\infty}$ norm are provided. The 
$l_{2}$ norm attacks include five attacker operations, namely FGSM-L2Attack,  PGD-L2Attack, DDNL2Attack, MT-L2Attack, and MI-L2Attack. The 
$l_{\infty}$ norm attacks include six attacker operations, namely FGSM-LinfAttack, PGD-LinfAttack, CW-LinfAttack, MT-LinfAttack , and MI-LinfAttack. 

\begin{table}[h]
	\centering
	\caption{The search space of attacker operations.}
	\renewcommand{\arraystretch}{1} 
	\begin{tabular}{ccc}
		\midrule
		$l2$ & $l\infty$  \\
		\midrule
		MI-L2Attack  &  FGSM-LinfAttack\\
		PGD-L2Attack  & PGD-LinfAttack\\
		CW-L2Attack & CW-LinfAttack \\
		MT-L2Attack    & MT-LinfAttack   \\
		MomentumIterative-L2Attack     & MI-LinfAttack   \\
		&MomentumIterative-LinfAttack\\
		\midrule
	\end{tabular}
	\label{attackerop}
\end{table}

Besides, we provide different attacker loss functions, which are illustrated in Table~\ref{caa}. It has proved that the loss functions also have a great influence on the performance of adversarial attacks. We include four types of loss functions: $l_{\text{CE}}$, $l_{\text{Hinge}}$, $l_{\text{L1}}$ and $l_{\text{DLR}}$. Further, calculating the output of logit or probability is also one key factor for the performance of adversarial attacks. Logit stands for the output of the final full-connected layer of DNNs, while probability means the output handled by the softmax operation. Due to the calculation of cross entropy needs to ensure the positive input, the logit output is not considered. Thus, the search space of provided loss functions can form into seven loss functions totally. In practical, due to the specific design of the loss function of CW-Attack itself, we do not take the loss functions of that into consideration.  

\begin{table}[htbp]
	\centering
	\caption{The search space of attacker loss functions.}
	\renewcommand{\arraystretch}{1} 
	\begin{tabular}{cccc}
		\midrule
		\multicolumn{1}{c}{Name} &Logit/Prob & Loss  \\
		\midrule
		\multicolumn{1}{c}{$l_{\text{CE}}$}    & P&  $
		-\sum_{i=1}^{K} y_{i} \log \left(Z(x)_{i}\right)
		$\\
		\multicolumn{1}{c}{$l_{\text{Hinge}}$}     &L/P & $
		\max \left(-Z(x)_{y}+\max _{i \neq y} Z(x)_{i},-\kappa\right)
		$ \\
		\multicolumn{1}{c}{$l_{\text{L1}}$}     &L/P & $
		-Z(x)_{y}
		$ \\
		\multicolumn{1}{c}{$l_{\text{DLR}}$}     &L/P & $
		-\frac{Z(x)_{y}-\max _{i \neq y} Z(x)_{i}}{Z(x)_{\pi_{1}}-Z(x)_{\pi_{3}}}
		$   \\
		\midrule
	\end{tabular}
	\label{caa}
\end{table}

Further, in our defined search space, the iteration number and magnitude are divided uniformly into 8 parts. Thus, each attacker sequence consists of multiple attack cells. Given the predefined maximum magnitude, the magnitude of the generated adversarial example by the attacker sequence is required to not exceed it. 

\begin{algorithm}[!ht]	\label{alg5}
	\caption{DE based Search for Efficient Attacks}
	\LinesNumbered
	\KwIn{ 
		The population size ${n}$, search space, training epoch $t$.
	}
	\KwOut{The searched near-optimal attack  $\textbf{x}$.}
	
	Encode the individual according to the defined search space; \\ 
	$\textbf{P}$ $\gets$ generate the initial population according to the encoding;\\
	Searching epoch $t=1$; \\ 
	\While{t<T}{
		
		$\textbf{P}_{m}$$\leftarrow$ Population mutation;\\
		$\textbf{P}_{c}$  $\leftarrow$ Population crossover based on $\textbf{P}_{m}$;\\
		$\textbf{P}_{f}$  $\leftarrow$ Population fituning based on $\textbf{P}_{c}$;\\
		//  \textbf{Population selection}\\
		Individual counter number $j = 1$;\\
		\For{$j <= n$}{
			\If {fitness $(\textbf{P}_{f})[j]$  $<$ fitness$(\textbf{P})[j]$}
			{$\textbf{P}[j]$ = $\textbf{P}_{f}[j]$;}
			$j=j+1$;
		}
		
		Obtain the individual  $\textbf{x}$ with the lowest robustness $\leftarrow$ $argmin$($fitness$(\textbf{P}))\\
		$t=t+1$;\\
	}
	\textbf{return}  $\textbf{x}$.
	
\end{algorithm}

%
%

\begin{algorithm}[h]
	\caption{NSGA-II based Search for Efficient Attacks}
	\label{alg6}
	\LinesNumbered
	\KwIn{The maximum number of iteration ${T}$ in NSGA-II, population size $n$, training set $D_{train}$, validation set $D_{test}$, the crossover rate $P_{c}$, the mutation rate $P_{m}$, the length of the searched adversarial attack $c$;}
	\KwOut{ The near optimal adversarial attack.}
	Generate the initial population $\textbf{P}$  with $n$ candidate adversairial attack randomly;
	\\
	Evaluate the robustness accuracy and time cost of each individual in  $\textbf{P}$;\\
	
	\While{$t \leftarrow 1:T$}{
		Generate $n$ offspring individuals using crossover and mutation operation;\\
		Merge the offspring individuals with $\textbf{P}$ and conduct the evaluation;\\
		Select $n$ individuals to form the new $\textbf{P}$ using non-dominated-sorting and crowding distance;\\ 	
	}
	
	Return The near-optimal adversarial attack
\end{algorithm}

\section{Experiments}\label{sec4}

To verify the effectiveness of our proposed $A^{3}D$ framework, we conduct a series of experiments for the image classification task on CIFAR10, CIFAR100, and ImageNet datasets. In section \ref{sec41}, the experimental protocol is described. In section \ref{sec42}, the experiment results of AutoML for adversarial attack are introduced. In section \ref{sec43}, the experimental results of AutoML for adversarial defense are introduced. In section \ref{sec44}, the circuit of auto adversarial attack and defense is elaborated.

\subsection{Experimental Protocol}\label{sec41}

\textbf{In the phase of searching for architectures}: 
In the following sections, we select eight NAS methods, including DARTS, PC-DARTS, Random Search, NASP, FairDARTS, SmoothDARTS, Weight-sharing based Random Search, and DE. In the differentiable NAS methods, the supernets constructed by these methods are trained using the identical setting. The search epoch is set to 50. The batch size of data is set to 64. The initial channel is set to 16. The layer of the total network is set to 8. The original samples of 50,000 in the training data are split in half for updating the weight parameters and architecture parameters, respectively. For updating the weight parameters, the SGD optimizer with the initial learning rate of 0.1 and the weight decay factor of $3 \times 10^{-3}$ is adopted, which is annealed to zero through cosine scheduling. For updating the architecture parameters, the Adam optimizer with the learning rate of $6 \times 10^{-4}$ and the weight decay factor of $1 \times 10^{-3}$. As for the coefficients $\gamma$ of all experiments in the total loss are all set to 1. In addition, both the percentages of natural noise and system noise are set to 0.5.

\textbf{In the phase of retraining the searched architectures}: The searched cell would be repeated 20 times to form a whole network. The initial channel is set to 36, and the total layer is set to 20. For the fair
evaluation of each architecture’s intrinsic robustness, all the
tested architectures are trained for 50 epochs under identical training settings. 
The SGD optimizer with the initial learning rate of 0.025 and the weight decay factor of $3 \times 10^{-4}$ is adopted. 

\textbf{Robustness evaluation}:  The robustness of the trained model is evaluated using four types of metrics. In the adversarial noise, the attack magnitude of all attack algorithms is set to 0.01. The robust accuracy of the trained model under adversarial attack is calculated as Eq. \ref{eq1}.

\begin{table*}[htbp]
	\renewcommand\arraystretch{1.5}
	\scriptsize
	\centering
	\caption{Comparison of the robust accuracy ($\%$) and time cost (s) under the $l_{\infty}$ attack of the manually-designed attacks, the searched attacks by local search, PSO, DE, NSGA-II across various defense strategies with the magnitude set as 8/255 using 500 images on CIFAR10 dataset.}
	
	\label{CIFAR10Linf}
	\setlength{\tabcolsep}{4mm}{
		\begin{tabular}{c|c|ccccccccccc}			
			\toprule			
			\multirow{2}{*}{Defense Method}&	\multirow{2}{*}{Model} &\multicolumn{2}{c}{Clean}&  \multicolumn{2}{c}{FGSM-LinfAttack} &  \multicolumn{2}{c}{PGD-LinfAttack-7} &\multicolumn{2}{c}{CW-LinfAttack-50} &\multicolumn{2}{c}{MI-LinfAttack-50}\\
			
			\cmidrule(r){3-4}   \cmidrule(r){5-6}  \cmidrule(r){7-8}
			\cmidrule(r){9-10} 	\cmidrule(r){11-12}
			&	& acc & time	 &   acc		
			&  time      &  acc   &   time  &  acc   &   time  & acc   &   time  \\		
			\midrule		
			{PGDAT \cite{pgd2018}}&	{ResNet-50}   &  89.6&0.89&  61.2&3.05&54.6&12.81 &53.4&73.58&53.2&73.44\\	
			{AWP}\cite{wu2020adversarial}	&	\text {WRN-28-10 }     & 89.8 &1.12&66.8&4.10&63.0&17.25&59.8&97.96&61.8&97.99\\
			{AWP}\cite{wu2020adversarial}	&	\text {WRN-34-10 }  &85.8&1.26  &62.6&4.93&57.8&21.00 &55.4&120.68&57.0&120.73\\
			MART \cite{wang2019improving} & WRN-28-10 &88.4&1.08&69.2&4.15&63.4&16.95&58.0&100.04&62.0&100.29\\
			
			TRADES \cite{zhang2019theoretically} & WRN-34-10&86.8&1.26&60.6&4.15&55.2&20.56&52.4&121.44&53.8&122.0\\
			Feature Scatter \cite{zhang2019defense}  & WRN-28-10 & 92.0 &1.10&78.2&4.20&68.6&17.33&50.6&100.76&65.6&101.27\\
			Interpolation \cite{zhang2019adversarial} & WRN-28-10 &90.6&1.10 &78.4&4.08&70.6&16.76&62.0&100.39&71.4&100.64\\
			Regularization \cite{jin2020manifold} & ResNet-18 & 92.2&0.61&80.2&1.35&58.8&5.33&35.6&29.72&57.4&29.66\\
			\midrule
			
			\multirow{2}{*}{Defense Method}&	\multirow{2}{*}{Model} &\multicolumn{2}{c}{Local search}&  \multicolumn{2}{c}{PSO} &  \multicolumn{2}{c}{DE} &\multicolumn{2}{c}{Random Search} &\multicolumn{2}{c}{NSGA-II}\\

			\cmidrule(r){3-4}  \cmidrule(r){5-6}  \cmidrule(r){7-8}
			\cmidrule(r){9-10} 	\cmidrule(r){11-12}
			&	& acc & time	 &   acc		
			&  time      &  acc   &   time  &  acc   &   time  & acc   &   time  \\	
			\midrule		
			{PGDAT \cite{pgd2018}}&	{ResNet-50}   & 53.4&81.9&52.2 & 359.3&\textbf{50.8}&458.9 &53.0&204.14& 52.0 &145.88\\	
			{AWP}\cite{wu2020adversarial}	&	\text {WRN-28-10 }     &59.8  &134.2&59.8&540.1&\textbf{59.2}&788.2&59.8&303.55&59.6&204.03\\
			{AWP}\cite{wu2020adversarial}	&	\text {WRN-34-10 }  &55.6&161.2& 55.2&654.4&\textbf{54.8} &946.0&55.2&360.54& 55.6&241.26\\
			MART \cite{wang2019improving} & WRN-28-10 &58.0&132.8&57.4&538.7&\textbf{56.6}&772.5&58.2&301.65&57.4&205.96\\
			
			TRADES \cite{zhang2019theoretically} & WRN-34-10&52.6&158.7&52.6&657.9&\textbf{51.6}&912.7&52.4&360.92&52.2 &240.49\\
			Feature Scatter \cite{zhang2019defense}  & WRN-28-10&53.4 &129.4&\textbf{49.6}&533.9&38.4&717.0&54.4&299.85&56.2&205.18\\
			Interpolation \cite{zhang2019adversarial} & WRN-28-10 &64.8&137.6&63.0&548.4&\textbf{36.6}&810.11&68.4&308.66&65.8&208.44\\
			Regularization \cite{jin2020manifold} & ResNet-18 &41.0&29.7&45.6&129.9&\textbf{5.00}&189.45&53.6&96.90&42.2&66.16 \\
			
			\bottomrule		
	\end{tabular}}
\end{table*}

\begin{table*}[htbp]
	\renewcommand\arraystretch{1.5}
	\scriptsize
	\centering
	\caption{Comparison of the robust accuracy ($\%$) and time cost (min) under the $l_{2}$ attack of the manually designed attacks, the searched attacks by local search, PSO, DE, NSGA-II across various defense strategies with the magnitude set as 0.5 using 500 images on CIFAR10 dataset.}
	
	\label{CIFAR10L2}
	\setlength{\tabcolsep}{3mm}{
		\begin{tabular}{c|c|ccccccccccc}			
			\toprule			
			\multirow{2}{*}{Defense Method}&	\multirow{2}{*}{Model} &\multicolumn{2}{c}{Clean}&  \multicolumn{2}{c}{MomentumIterative-L2Attack-50} &  \multicolumn{2}{c}{PGD-L2Attack-7} &\multicolumn{2}{c}{CW-L2Attack-50} &\multicolumn{2}{c}{MI-L2Attack-50}\\
			
			\cmidrule(r){3-4}   \cmidrule(r){5-6}  \cmidrule(r){7-8}
			\cmidrule(r){9-10} 	\cmidrule(r){11-12}
			&	& acc & time	 &   acc		
			&  time      &  acc   &   time  &  acc   &   time  & acc   &   time  \\		
			\midrule		
			{PGDAT \cite{pgd2018}}&	{ResNet-50}   &  89.6&0.89& 82.4 &35.84&63.2&13.19 &69.2&72.19&61.6&73.68\\	
			{AWP}\cite{wu2020adversarial}	&	\text {WRN-28-10 }     & 89.8 &1.12&84.6&47.15&67.8&17.13&68.2&95.68&66.4&98.11\\
			{AWP}\cite{wu2020adversarial}	&	\text {WRN-34-10 }  &85.8&1.26  &80.4&60.75&63.6&20.83 &63.8&117.5&62.4&119.9\\
			MART \cite{wang2019improving} & WRN-28-10 &88.4&1.08&83.8&53.01&69.6&17.34&74.2&97.05&68.0&99.82\\
			
			TRADES \cite{zhang2019theoretically} & WRN-34-10&86.8&1.26&79.0&66.44&61.8&21.15&62.8&119.44&59.8&121.94\\
			Feature Scatter \cite{zhang2019defense}  & WRN-28-10 & 92.0 &1.10&80.8&55.27&74.2&17.84&76.8&99.31&73.6&101.66\\
			Interpolation \cite{zhang2019adversarial} & WRN-28-10 &90.6&1.10 &80.6&54.97&74.2&17.82&76.6&98.5&74.2&100.84\\
			Regularization \cite{jin2020manifold} & ResNet-18 & 92.2&0.61&81.4&14.66&80.0&5.43&75.8&29.24&79.2&30.02\\
			\midrule
			
			\multirow{2}{*}{Defense Method}&	\multirow{2}{*}{Model} &\multicolumn{2}{c}{Local search}&  \multicolumn{2}{c}{PSO} &  \multicolumn{2}{c}{DE} &\multicolumn{2}{c}{Random Search} &\multicolumn{2}{c}{NSGA-II}\\

			\cmidrule(r){3-4}  \cmidrule(r){5-6}  \cmidrule(r){7-8}
			\cmidrule(r){9-10} 	\cmidrule(r){11-12}
			&	& acc & time	 &   acc		
			&  time      &  acc   &   time  &  acc   &   time  & acc   &   time  \\	
			\midrule		
			{PGDAT \cite{pgd2018}}&	{ResNet-50}   & 58.4&205.02&57.6&775.67&58.2&256.32&58.2&229.20&\textbf{57.8}&771.10\\	
			{AWP}\cite{wu2020adversarial}	&\text {WRN-28-10 } &65.4&278.84&\textbf{65.0}&1035.9&\textbf{65.0}&348.82&65.2&270.09&\textbf{65.0}&1033.58	\\
			{AWP}\cite{wu2020adversarial}	&\text {WRN-34-10 } &60.4&336.07&60.2&1245.62&60.2&422.32&60.4&322.49&\textbf{60.0}&1237.34	\\
			MART \cite{wang2019improving} & WRN-28-10 &62.8&284.32&62.6&1057.3&62.8&357.8&62.8&276.26&\textbf{62.4}&1047.67\\
			
			TRADES \cite{zhang2019theoretically} & WRN-34-10&58.8&338.15&\textbf{58.6}&1243.5&58.8&427.86&\textbf{58.6}&324.65&\textbf{58.6}&1238.33\\
			Feature Scatter \cite{zhang2019defense}  & WRN-28-10&55.4&282.63&\textbf{54.0}&1004&57.2&349.51&57.6&270.55&\textbf{54.2}&994.20\\
			Interpolation \cite{zhang2019adversarial} & WRN-28-10 &52.2&286.91&\textbf{51.2}&1016&66.6&361.09&66.8&282.34&\textbf{51.2}&1003.92\\
			Regularization \cite{jin2020manifold} & ResNet-18 &33.4&92.02&27.0&309.36&70.8&111.33&69.6&98.25&\textbf{26.2}&311.83\\
			
			\bottomrule		
	\end{tabular}}
\end{table*}

\textbf{In the phase of searching for attacks}: 
In the process of searching for efficient adversarial attacks using evolutionary algorithms, we limit the length of the attacker sequence to 3. The maximum iteration number of each attack operation is set to 50. The population size and iteration number of PSO, DE, and NSGA-II are all set to 20 and 5, respectively. For local search, we allow the algorithm to explore the variable length of the auto adversarial attack, while the maximum length is still set to 3. The number of train samples to evaluate the models by the adversarial attack is set to 500 to accelerate the evaluation of adversarial attacks. 

\textbf{Attacker evaluation}: The performance of the searched attack is evaluated under multiple defensed models. The number of test images utilized in CIFAR10, CIFAR100, and ImageNet is 500, 500, and 1280, respectively. The batch sizes are all set to 16.

The above experiments are conducted under the same environment using a single NVIDIA GTX Titan GPU. The visualizations of the searched model architectures and composite adversarial attack are presented in Appendix.

\subsection{AutoML for Adversarial Attack}\label{sec42}

\subsubsection{The performance of the searched attacks using single-objective optimization}

To illustrate the effectiveness of AAA, we first make a comparison between AAA and existing manually-designed adversarial attacks. The threat models are defensed by various techniques, including PGD adversarial training (PGDAT) \cite{pgd2018}, AWP \cite{wu2020adversarial}, MART \cite{wang2019improving}, TRADES \cite{zhang2019theoretically}, Feature Scatter \cite{zhang2019defense}, Interpolation \cite{zhang2019adversarial}, Regularization \cite{jin2020manifold}, Fixing data \cite{rebuffi2021fixing}, OAAT \cite{addepalli2021towards}, LBGAT \cite{cui2021learnable}, Overfit \cite{rice2020overfitting}, IAR \cite{bernhard2020luring}, fast adversarial training (FastAT) \cite{wong2020fast}, label smoothing \cite{shafahi2019label}, and ensemble adversarial training (EnsembleAT) \cite{tramer2017ensemble}. The selected model architectures include WideResNet (WRN), InceptionV3, and Inception$\_$resnetV2. To decrease the computational burden, the robustness of all defensed models is evaluated by the attack that is searched on one source model. In this way, we do not need to conduct the search algorithms for each denfesed model repeatedly. The ResNet-50 model defensed by PGDAT is selected as the source model to conduct the search.

In this section, we benchmark several single-objective optimization methods for solving Eq.~\ref{eqsingle} to obtain stronger adversarial attacks. Some manually-designed attack algorithms are taken as the comparison, including FGSM-LinfAttack, PGD-LinfAttack, CW-LinfAttack, and MI-LinfAttack, where the iteration number is set to 1, 7, 50, and 50, respectively. The experimental results about $l_{\infty}$ attack on the CIFAR10 dataset are presented in Table \ref{CIFAR10Linf}. The visualization of the searched attacks by different algorithms is shown in Table \ref{visualization}. Other results on different datasets, $l_{p}$, $l_{2}$ norm based attacks are listed in Table \ref{CIFAR10L2},  Table \ref{CIFAR100Linf}, Table \ref{CIFAR100L2}, Table \ref{ImagenetLinf}, and Table \ref{ImagenetL2} in the Appendix.

From Table \ref{CIFAR10Linf}, we can see that in most cases, the stronger the adversarial attack is, the time cost is longer. Hence, it is necessary to find efficient attacks with lower robust accuracy and less time cost, which can be formulated as a multi-objective optimization problem. In three single-objective optimization algorithms, experimental results show that DE performs the best, which can find the strongest adversarial attack. Particularly in WRN-28-10 defensed by Interpolation and Regularization, we can see that the robust accuracies of models by the attack searched by DE are farther lower than other attacks, which are only 36.6$\%$ and 5.00$\%$, respectively. In general, we can see that the combination of multiple types of adversarial attacks can help improve the effectiveness of the algorithms. Almost all the searched attacks by the optimization algorithms are better than several existing manually-designed attacks. From the search results, we can see that auto adversarial attack prefers the stronger attack, such as MT-Attack, while some other relatively stronger attacks are also important, such as MomentumIterative-Attack.

\begin{table*}
	\scriptsize
	\centering
	\caption{The visualization of the searched $l_{\infty}$ attack by AAA across various defense strategies on CIFAR10 dataset. 'A' stands for the attacker operation,'L'is the loss function, 'M' denotes the attack magnitude, 'I' represents the iteration number.}
	
	\label{visualization}
	\setlength{\tabcolsep}{4mm}{
		\begin{tabular}{cccc}		
			\toprule
		{Type}&	{Search Method}&	{The searched attack} &Search time (GPU/Days)\\
			\midrule		
		$l_{\infty}$&	{Local search}  &\makecell[c]{'A': CW-LinfAttack, 'M': 8/255,'I': 25; 'A': FGSM-LinfAttack,'L':CE, \\'M': 6/255, 'I': 1;'A': CW-LinfAttack, 'M': 1/255, 'I': 31}  &0.30\\	
			\midrule
	
			$l_{\infty}$&	{PSO}	 & \makecell[c]{'A': MT-LinfAttack, 'L':Hinge, 'M': 7/255,'I': 19; 'A': MomentumIterative-LinfAttack, \\'L':Hinge, 'M': 8/255, 'I': 25;'A': CW-LinfAttack, 'M': 8/255,'I': 13} &0.24 \\
			\midrule	     
			$l_{\infty}$&	{DE}	   & \makecell[c]{'A': MT-LinfAttack, 'L':Hinge, 'M': 8/255,'I': 6; 'A': MT-LinfAttack,\\'L':CE, 'M': 8/255, 'I': 31; 'A': MomentumIterative-LinfAttack,'L':L1$\_$P, 'M': 8/255,'I': 19}   &0.15  \\
			\midrule	
			$l_{\infty}$&	Random search   &   \makecell[c]{'A': MI-LinfAttack, 'L':Hinge, 'M': 8/255,'I': 43; 'A': FGSM-LinfAttack,\\'L':Hinge, 'M': 8/255, 'I': 1; 'A': CW-LinfAttack,'M': 8/255,'I': 12}  &0.06  \\
			\midrule
			$l_{\infty}$&	NSGA-II  &      \makecell[c]{'A': PGD-LinfAttack, 'L':Hinge$\_$P, 'M': 8/255,'I': 25; 'A': CW-LinfAttack,\\ 'M': 7/255, 'I': 12; 'A': FGSM-LinfAttack,'L':L1, 'M': 1/255,'I': 1}&0.97\\
			\midrule
		$l_{2}$&	{Local search}  & \makecell[c]{'A': MT-L2Attack, 'L':DLR, 'M': 0.5,'I': 25; 'A': MI-L2Attack,'L':CE, 'M': 0.375,'I': 6}  &0.60 \\	
			\midrule
			
		$l_{2}$&	{PSO}	 & \makecell[c]{'A': MT-L2Attack, 'L':DLR, 'M': 0.5,'I': 50;'A': MT-L2Attack, \\'L':DLR, 'M': 0.5,'I': 50; 'A': CW-L2Attack, 'M': 0.375,'I': 18} &0.62 \\
			\midrule	     
		$l_{2}$&	{DE}	    &  \makecell[c]{'A': MT-L2Attack, 'L':Hinge, 'M': 0.5,'I': 25; 'A': MI-L2Attack,\\'L':Hinge, 'M': 0.25,'I': 25;'A': MI-L2Attack,'L':L1$\_$P, 'M': 0.5,'I': 43} &0.78  \\	
			\midrule	
		$l_{2}$&	Random search   &   \makecell[c]{'A': MT-L2Attack, 'L':Hinge, 'M': 0.5,'I': 31; 'A': CW-L2Attack,\\ 'M': 0.5,'I': 12;'A': MI-L2Attack,'L':L1, 'M': 0.4375,'I': 18}  &0.15  \\
			\midrule
		$l_{2}$&	NSGA-II  &    \makecell[c]{'A': MT-L2Attack, 'L':CE, 'M': 0.5,'I': 50; 'A': PGD-L2Attack,\\'L':L1$\_$P, 'M': 0.0625,'I': 18;'A': MI-L2Attack,'L':CE, 'M': 0.5,'I': 50}  &0.58 \\
			\bottomrule		
	\end{tabular}}
\end{table*}

\begin{figure*}[htbp]
	\centering
	\subfloat[Random Search]{\includegraphics[width=3in]{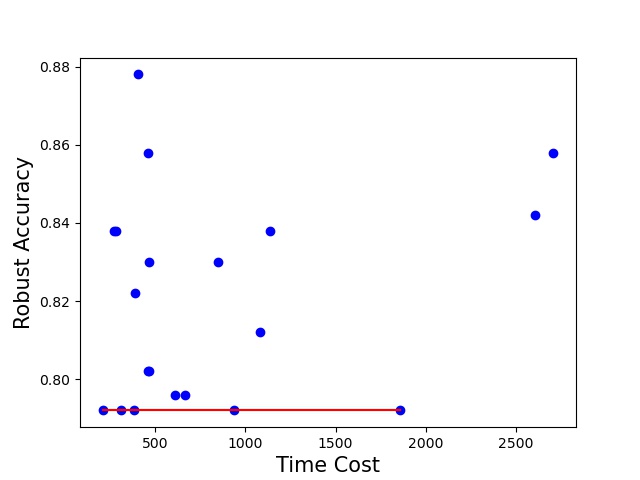}%
	}
	\hfil
	\subfloat[NSGA-II]{\includegraphics[width=3in]{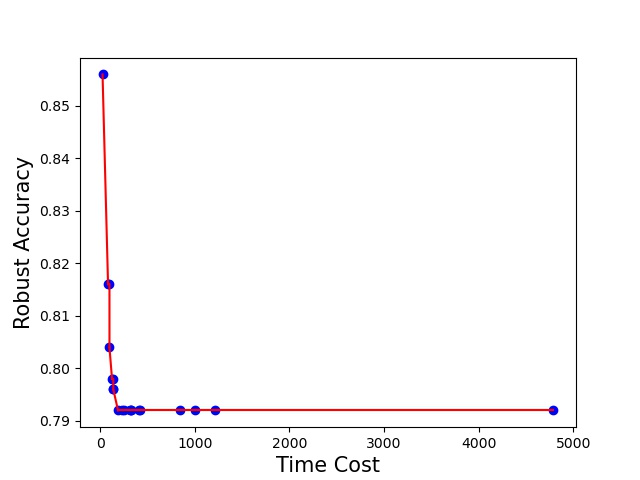}%
	}
	
	\caption{The visualization of the searched attacks using random search and NSGA-II.}
	\label{aaa}
\end{figure*}

\subsubsection{The performance of the searched attacks under multi-objective optimization}
In multi-objective optimization for efficient attacks, the attack with the lowest robust accuracy and time cost is taken as the final searched attack. If there exist multiple attacks with the same lowest robust accuracies, the one with the least time cost is regarded as the best. The comparison of random search and NSGA-II is presented in Fig. \ref{aaa}. From Fig. \ref{aaa}, we can see that the Pareto of NSGA-II is better than the random search.

The searched attacks are evaluated in the test data, the results of which are shown in Table \ref{CIFAR10Linf} and Table \ref{CIFAR10L2}. From Table \ref{CIFAR10Linf} and Table \ref{CIFAR10L2}, we can see that compared with the single-objective optimization, multi-objective optimization algorithms can find the attacks with less time cost and slightly higher robust accuracy. Among that, the robust accuracy of models by the searched attack by NSGA-II on the CIFAR10 dataset, including the $l_{\infty}$ and $l_{2}$ norm-based, is lower than random search, which can illustrate the effectiveness of NSGA-II than random search. Thus, more multi-objective optimization algorithms can be further developed to find more efficient adversarial attacks. We list the search cost of different methods in Table \ref{visualization}. From Table \ref{visualization}, we can see that NSGA-II is the most time-consuming algorithm. In general, for the CIFAR10 dataset, the best algorithms for $l_{\infty}$ and $l_{2}$ norm-based attacks are NSGA-II and DE, respectively. In addition, under our settings, the search time of $l_{\infty}$ attack is more than $l_{2}$ attack using the same optimization algorithm. It is because the robust accuracy of DNN models under $l_{2}$ attack is higher than $l_{\infty}$ attack under our settings, which means more computational cost to perform adversarial attacks.


\subsection{AutoML for Adversarial Defense}\label{sec43}

\subsubsection{The performance of the searched architectures by different methods under the same robustness evaluations}

In this section, we investigate the robustness performance of the searched architectures by various NAS methods under the same robustness evaluations. Specifically, in adversarial noises, we take the FGSM robust accuracy and Jacobian value as the evaluations in the NAS methods, and the searched architectures are evaluated in multiple adversarial noises, including FGSM-LinfAttack, PGD-LinfAttack, CW-LinfAttack, and MI-LinfAttack, where the iteration numbers are set to 1, 7, 7, and 7, respectively. In natural noises, we take the robust accuracy of the snow noises as the evaluation in NAS, and the searched architectures are evaluated in multiple natural noises, including brightness, fog, contrast, frost, snow, gaussian blur, and motion blur. Similar operations are also performed in system noises. Three manually-designed model architectures are selected for comparison. The experimental results of the performance of different models under multiple types of robustness evaluations are presented in Table \ref{NAS}, Table \ref{natural}, and Table \ref{system}, respectively. The search costs are listed in Table \ref{NAStime}.

From Table \ref{NAS}, Table \ref{natural}, and Table \ref{system}, we can see that NAS can help improve the robustness of DNN models by searching the near-optimal architectures, which can be verified in the comparison of random search and DE. Under three types of evaluations, including adversarial noises, natural noises, and system noise, DE can find more robust architectures than random search. From Table \ref{NAStime}, it can also be concluded that DE also needs more search cost than random search.

\begin{table*}[htbp]
	\renewcommand\arraystretch{1.5}
	\scriptsize
	\centering
	\caption{Comparison of the parameter number (M) and robust accuracy ($\%$) of adversarial noises of different models on the CIFAR10 dataset.}
	
	\label{NAS}
	\setlength{\textwidth}{24mm}{
		\begin{tabular}{c|c|c|ccccc}			
			\toprule			
			\multirow{1}{*}{ Method}&	\multirow{1}{*}{Type} &	\multirow{1}{*}{Param} &\multicolumn{1}{c}{Clean}&  \multicolumn{1}{c}{FGSM-LinfAttack}  & PGD-LinfAttack	 &   CW-LinfAttack		
			&  MI-LinfAttack      \\		
			\midrule		
			{ResNet-18}&	{Manual}   &   11.17&93.64&51.4&34.2&34.2&33.6  \\	
			{MobileNet-v2}&	{Manual}   & 2.30& 92.68&48.6&25.6&24.2&25.0   \\	
			{DenseNet-121}&	{Manual}   & 6.96&94.31&52.8&31.6&31.0&31.4   \\
			\midrule
			
			{DARTS$\_$Jacobian}	&	{Automatic}  &   3.23&89.13&58.6&51.2&52.4&50.8              \\
			{PC-DARTS$\_$$\_$Jacobian}	&	{Automatic}     &    4.63& 87.99&55.2&46.6&47.4&46.2       \\
			{NASP$\_$$\_$Jacobian}	&	{Automatic}   &3.22& 93.67&53.8&40.0&40.2&40.2      \\
			{FairDARTS$\_$$\_$Jacobian}	&	{Automatic}     &    2.76&   92.14&61.2&50.6&51.4&50.2       \\
			{SmoothDARTS$\_$$\_$Jacobian}	&	{Automatic}   &4.34&     93.96&56.6&42.2&41.4&  41.4       \\
			
			\midrule
			{DARTS$\_$FGSM}	&	{Automatic}  & 4.47  &92.86&   58.0&38.8&38.8&38.8           \\
			{PC-DARTS$\_$$\_$FGSM}	&	{Automatic}     & 3.80&91.39&57.2&43.8&44.6&43.2     \\
			{NASP$\_$$\_$FGSM}	&	{Automatic}         & 2.36&93.37&56.4&44.4&43.3&44.6  \\
			{FairDARTS$\_$$\_$FGSM}	&	{Automatic}     &     3.43&93.57&58.2&46.0&46.2&45.0          \\
			{SmoothDARTS$\_$$\_$FGSM}	&	{Automatic}   &2.35&92.66&\textbf{61.0}&\textbf{53.0}&\textbf{54.0}&\textbf{52.6}                  \\
			{Random$\_$Search$\_$FGSM}	&	{Automatic}  &   2.64&88.57&54.2&43.6&44.6&43.4                 \\
			{DE$\_$$\_$FGSM}	&	{Automatic}     &    3.07&92.26&58.4&45.2&45.0&  45.2         \\
			{Weight Sharing based Random$\_$Search$\_$FGSM}	&	{Automatic}  & 3.01&93.06&54.6&38.0&38.4&37.8 \\                  		
			\bottomrule		
	\end{tabular}}
\end{table*}

\begin{table*}[htbp]
	\renewcommand\arraystretch{1.5}
	\scriptsize
	\centering
	\caption{Comparison of the parameter number (M) and robust accuracy ($\%$) of natural noises of different models on the CIFAR10 dataset.}
	
	\label{natural}
	\setlength{\textwidth}{24mm}{
		\begin{tabular}{c|c|c|cccccccc}			
			\toprule			
			\multirow{1}{*}{ Method}&\multirow{1}{*}{ Type}&	\multirow{1}{*}{Params} &\multicolumn{1}{c}{Clean}&  \multicolumn{1}{c}{Brightness}  & Fog	 &   Contrast		
			&  Frost    &  Snow &Gauss$\_$blur&Motion$\_$blur\\		
			\midrule		
			{ResNet-18}&	{Manual}   &11.17&  93.64&  92.99&92.13&89.76&82.88&76.56&87.1&80.76\\	
			{MobileNet-v2}&	{Manual}   & 2.30&92.68&92.23&90.86&87.35&81.17&77.11&83.8&77.45\\
			{DenseNet-121}&	{Manual}   &6.96&94.31&93.76&92.61&90.48&82.69&78.06&88.35&81.35
			\\
			\midrule
			
			{DARTS$\_$Natural}	&	{Automatic}  & 2.83&93.99&93.52& 92.55&89.67&84.46&79.17&88.48&82.5                  \\
			{PC-DARTS$\_$$\_$Natural}	&	{Automatic}     &  3.46&91.73& 91.48&90.44&86.91&83.56&77.66&85.82&82.25          \\
			{NASP$\_$$\_$Natural}	&	{Automatic}  &3.95&   93.24&92.91&91.5&88.74&84.68&79.61&88.32&83.96              \\
			{FairDARTS$\_$$\_$Natural}	&	{Automatic}     &    3.60&  92.06&91.71&90.44&87.68&85.27&79.33&86.23&81.24         \\
			{SmoothDARTS$\_$$\_$Natural}	&	{Automatic}           &3.38&  89.42&88.99&87.49&82.53&79.15&71.19&83.94&79.26        \\
			{Random$\_$Search$\_$Natural}	&	{Automatic}  & 2.43&89.26&89.07&86.6&82.39&79.38&72.65&79.86&73.34                   \\
			{DE$\_$$\_$Natural}	&	{Automatic}     &    2.93&92.61&92.25&91.41&88.2&84.96&79.26&87.42&82.52           \\
			{Weight Sharing based Random$\_$Search$\_$Natural}	&	{Automatic}  &    2.50& 92.66 &92.35&90.86&88.21&85.15&79.35&87.55&82.96              \\
			
			\bottomrule		
	\end{tabular}}
\end{table*}

\begin{table*}[htbp]
	\renewcommand\arraystretch{1.5}
	\scriptsize
	\centering
	\caption{Comparison of the parameter number (M) and robust accuracy ($\%$) of system noises of different models on the CIFAR10 dataset.}
	
	\label{system}
	\setlength{\textwidth}{24mm}{
		\begin{tabular}{c|c|c|ccccc}			
			\toprule			
			\multirow{1}{*}{ Method}&	\multirow{1}{*}{Type} &\multirow{1}{*}{ Params}&\multicolumn{1}{c}{Clean}&  \multicolumn{1}{c}{opencv-pil$\_$hamming}  & opencv-pil$\_$bilinear	 &   pil-opencv$\_$nearest		
			&   \\		
			\midrule		
			{ResNet-18}&	{Manual}   &  11.17&93.64&87.60&85.79&85.70  \\	
			{MobileNet-v2}&	{Manual}   &  2.30&92.68&85.47&81.90&85.48   \\	
			{DenseNet-121}&	{Manual}   &  6.96&94.31&87.97&86.18&86.15 \\	
			\midrule	
			{DARTS$\_$System}	&	{Automatic}  &  4.45&  92.53&85.06&82.23&84.99                \\
			{PC-DARTS$\_$$\_$System}	&	{Automatic}  &2.67   &         92.75&  86.79&85.35&84.89     \\
			{NASP$\_$$\_$System}	&	{Automatic}   &4.25&91.8&84.88&83.09&85.16                 \\
			{FairDARTS$\_$$\_$System}	&	{Automatic}     &        2.75&93.12&87.12&85.56&86.88        \\
			{SmoothDARTS$\_$$\_$System}	&	{Automatic}        &    3.14&93.32&87.94&86.29&85.89         \\
			{Random$\_$Search$\_$System}	&	{Automatic}  &    2.43&90.69&84.84&81.81&85.27               \\
			{DE$\_$$\_$System}	&	{Automatic}     &  2.64&92.07&85.39&82.76&85.26          \\
			{Weight Sharing based Random$\_$Search$\_$System}	&	{Automatic}  &  3.21&90.94&84.67&81.58&85.57                \\

			\bottomrule		
	\end{tabular}}
\end{table*}

\begin{figure}[htbp]
	\centering
	\includegraphics[scale=0.28]{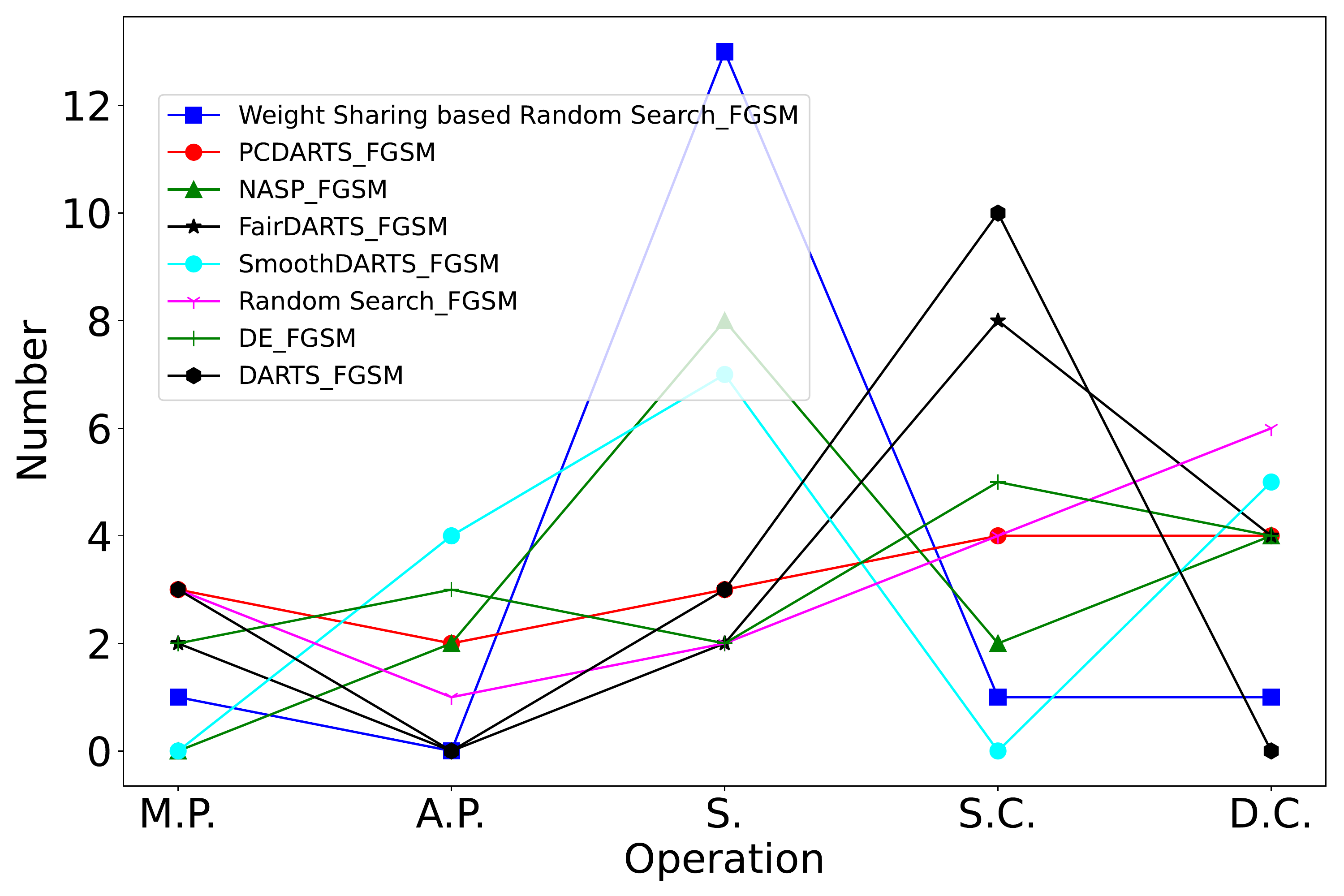}
	\caption{The number of operations of different architectures.}
	\label{op_number}
\end{figure}

If random$\_$search$\_$FGSM is taken as the baseline method in the NAS method, it can be concluded that both differentiable and non-differentiable methods can find more robust architectures. Among all NAS methods, SmoothDARTS$\_$FGSM can achieve the best performance. The robust accuracy under MI-LinfAttack can reach 52.6$\%$, farther higher than other manually-designed model architectures and NAS methods. In addition, we can see that the quantified metrics, such as the Jacobian matrix, can help find robust architectures, while it is not absolute, which may also cause the dropping of the robustness performance of the searched architectures, such as SmoothDARTS$\_$Jacobian. The robustness of DARTS$\_$FGSM under FGSM attack is relatively higher, while farther lower under other attacks than other models. These experimental results show that the appropriate robustness metric should be determined accordingly for different NAS methods.

To analyze the effect of different operations on the performance of architectures, we visualize the number of each operation of the searched architectures in Fig. \ref{op_number}. The max$\_$pool, avg$\_$pool, sep$\_$conv, dil$\_$conv, and skip$\_$connect, are denoted as M.P., A.P., and S., respectively. From Fig. \ref{op_number}, we can see that the number of skip$\_$connect of Weight Sharing based Random Search$\_$FGSM is the most, and the robustness of that is also poor. All architectures have at least two skip$\_$connect, which also verifies the importance of skip$\_$connect, such as the application of it in ResNet. Thus, we can conclude that the skip$\_$connect is necessary but can not be too more. Besides, DARTS$\_$FGSM has the largest number of sep$\_$conv, the robustness of which is also relatively poor. In contrast, the most robust architecture SmoothDARTS$\_$FGSM has operations with more diversity. In general, it can be concluded that the enrichment of a single type of operation is not conducive to improving model robustness, such as max$\_$pool or skip$\_$connect.

\begin{table*}[htbp]
	\renewcommand\arraystretch{1.5}
	\caption{The comparison of the search time (GPU/Days) of different NAS methods udner different robustness evaluations.}
	\label{NAStime}
	\centering
	{
		\begin{tabular}{ccccccccccc}
			\hline
			Method     & FGSM & PGD   & Natural  & System& Jacobian&Hessian \\ \hline
		DARTS &    0.6   &1.27&0.14&0.7&1&1\\
			PC-DARTS &    0.61&  1.32&0.06&0.19&0.375&0.75\\
			NASP &   0.17& 0.73&0.06&0.42&1&3.5 \\
			SmoothDARTS &0.83& 1.88&0.08&0.73& 2.15&5.15      \\
			FairDARTS &   0.81&1.5&0.06&0.06&1.42&4.7   \\
			Random Search &  1.15&1.27&1.10&1.10&1.10&1.20\\
			DE &     12&13&11&11&11&12  \\
			Weight Sharing-based Random Search     &0.6&0.7&0.55 &0.55 &0.55 &0.55    \\ \hline
	\end{tabular}}
\end{table*}

\subsubsection{The performance of the searched architectures using different robustness evaluations}

\begin{figure*}[h]
	\centering
	\subfloat{\includegraphics[width=2.3in]{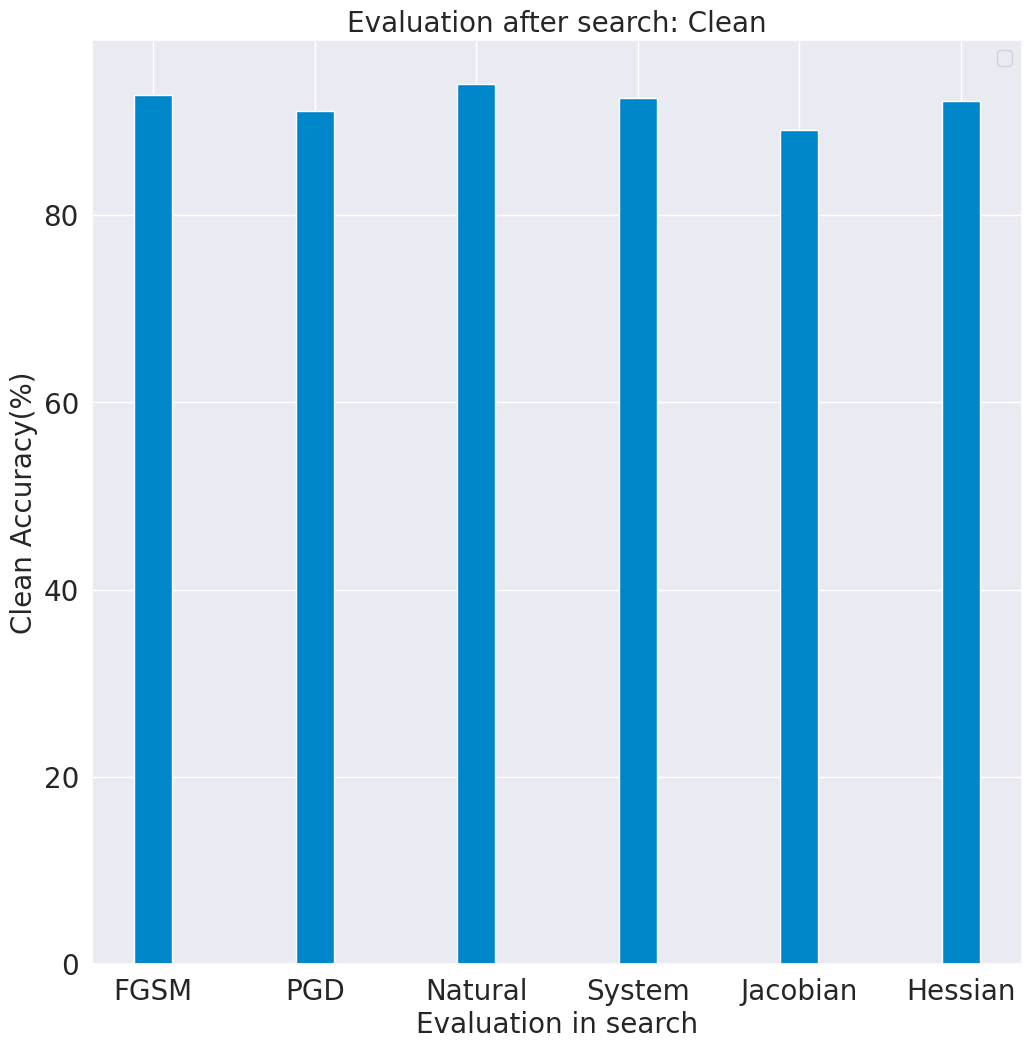}%
	}
	\hfil
	\subfloat{\includegraphics[width=2.3in]{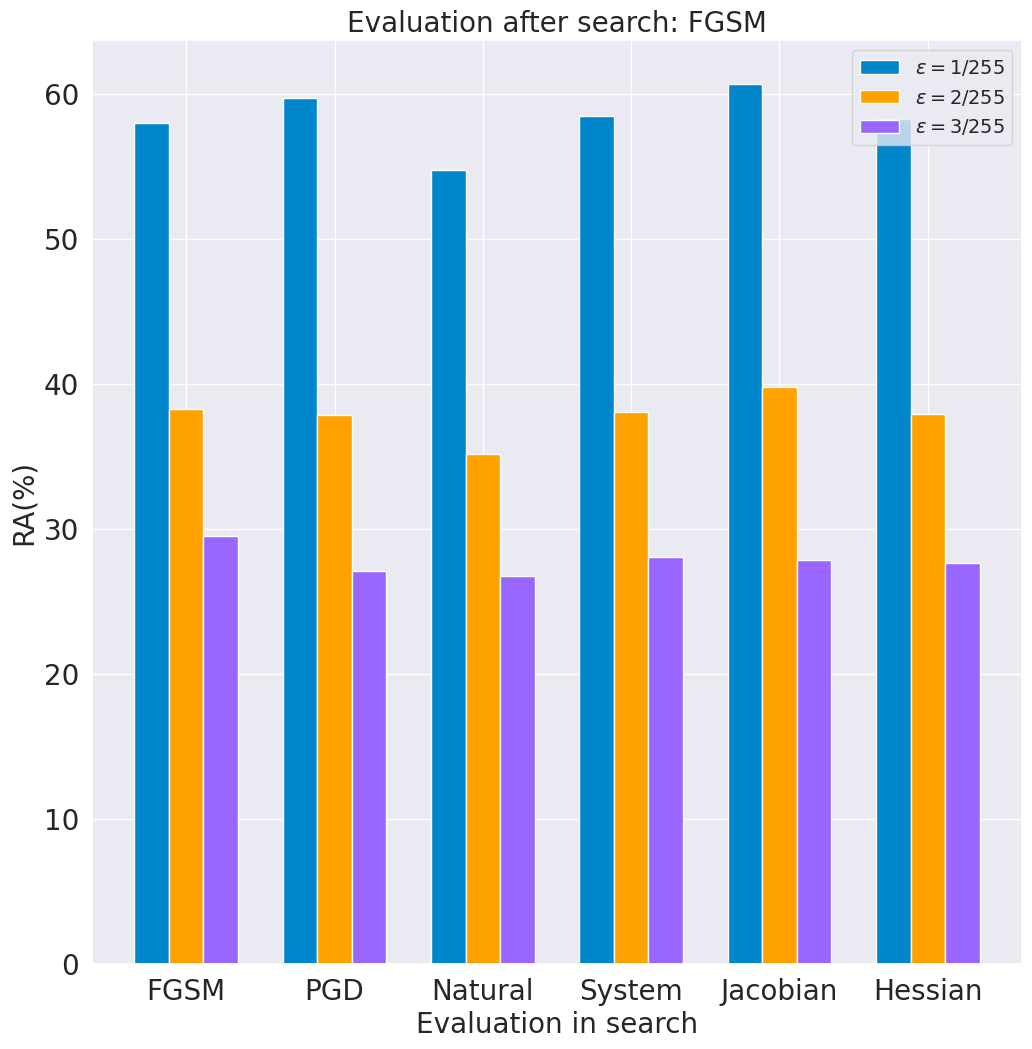}%
	}
	\subfloat{\includegraphics[width=2.3in]{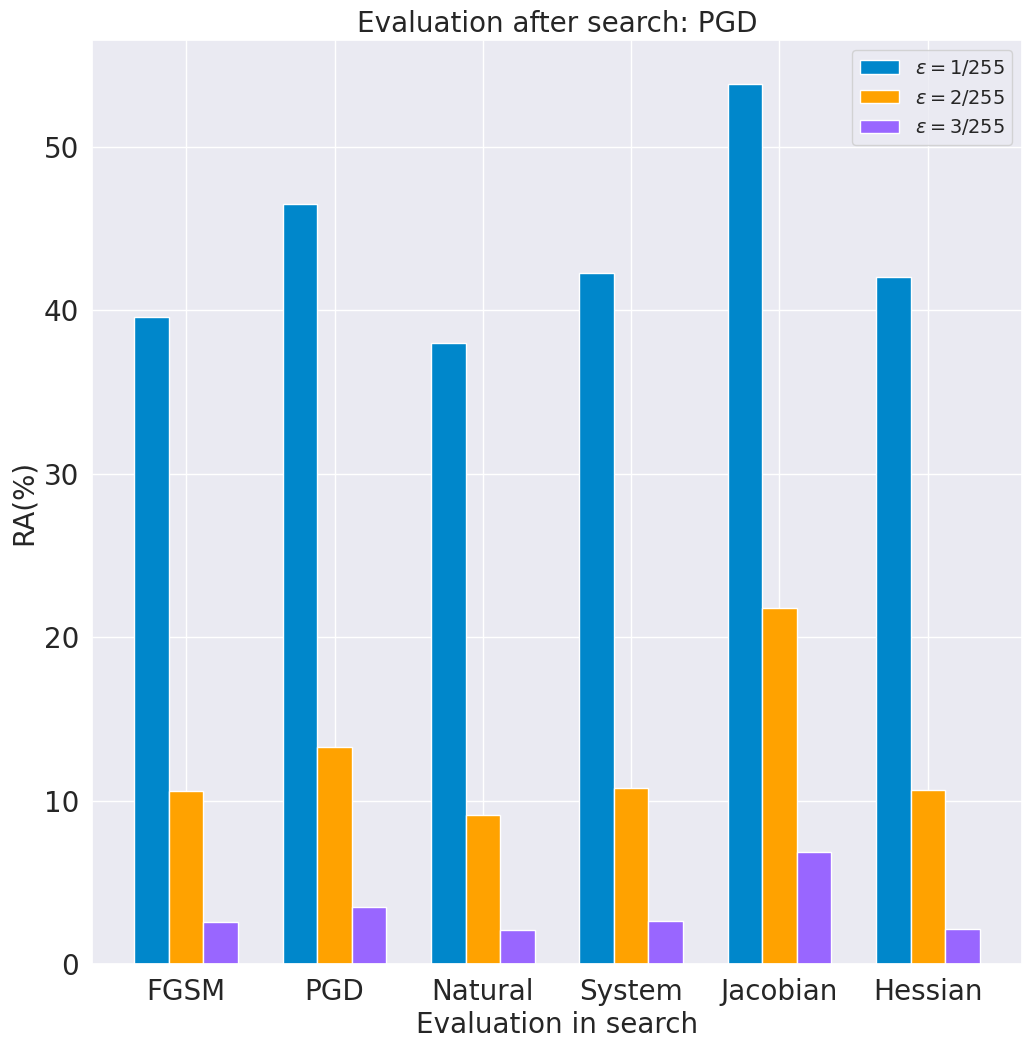}%
	}
	
	\subfloat{\includegraphics[width=2.3in]{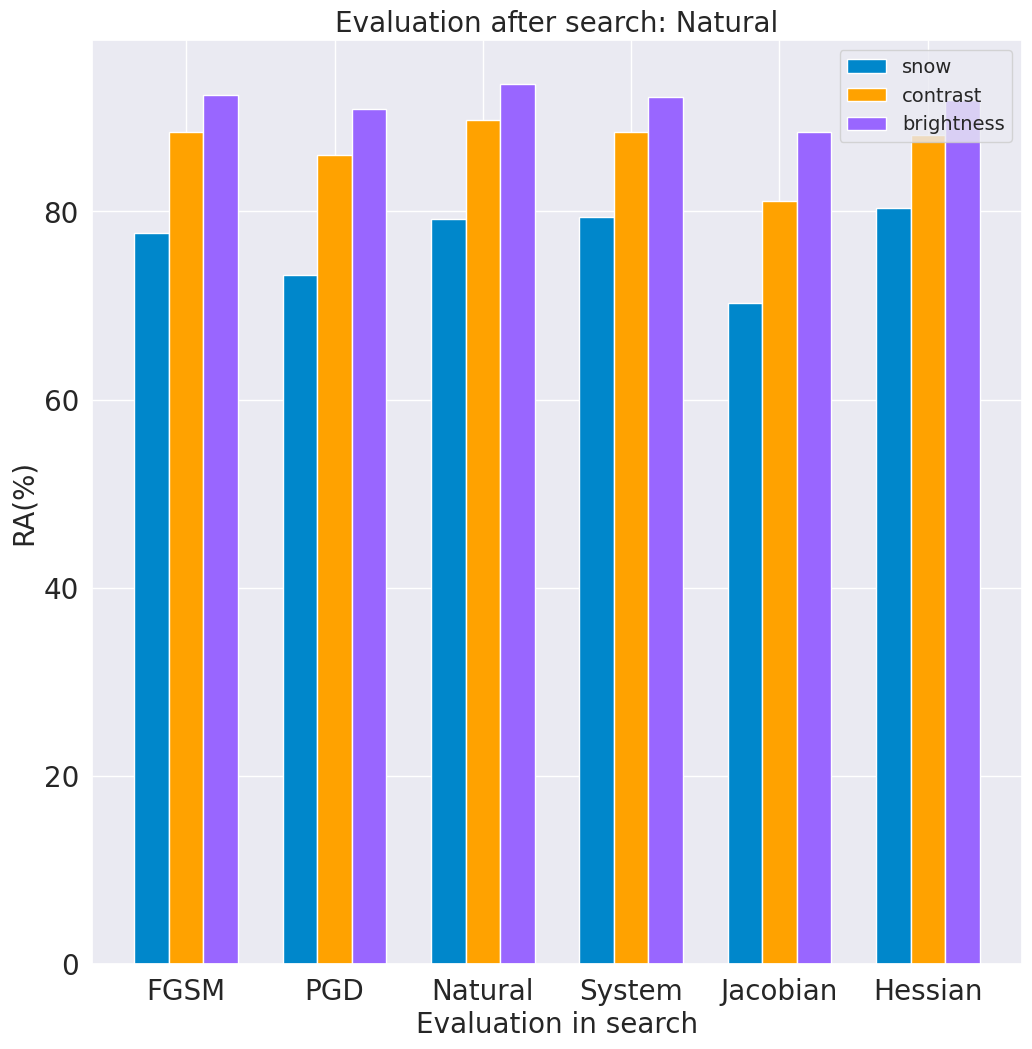}%
}
	\hfil
	\subfloat{\includegraphics[width=2.3in]{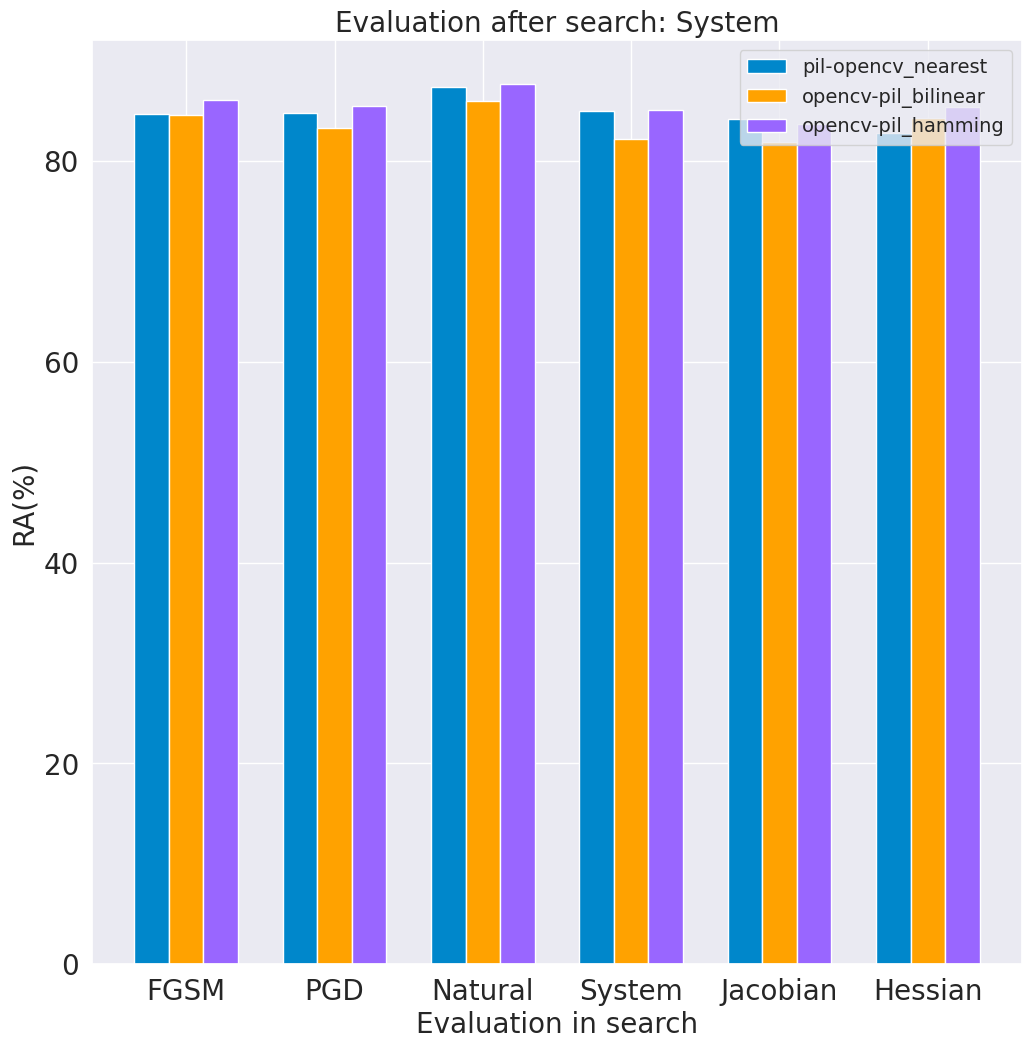}%
	}
	\subfloat{\includegraphics[width=2.3in]{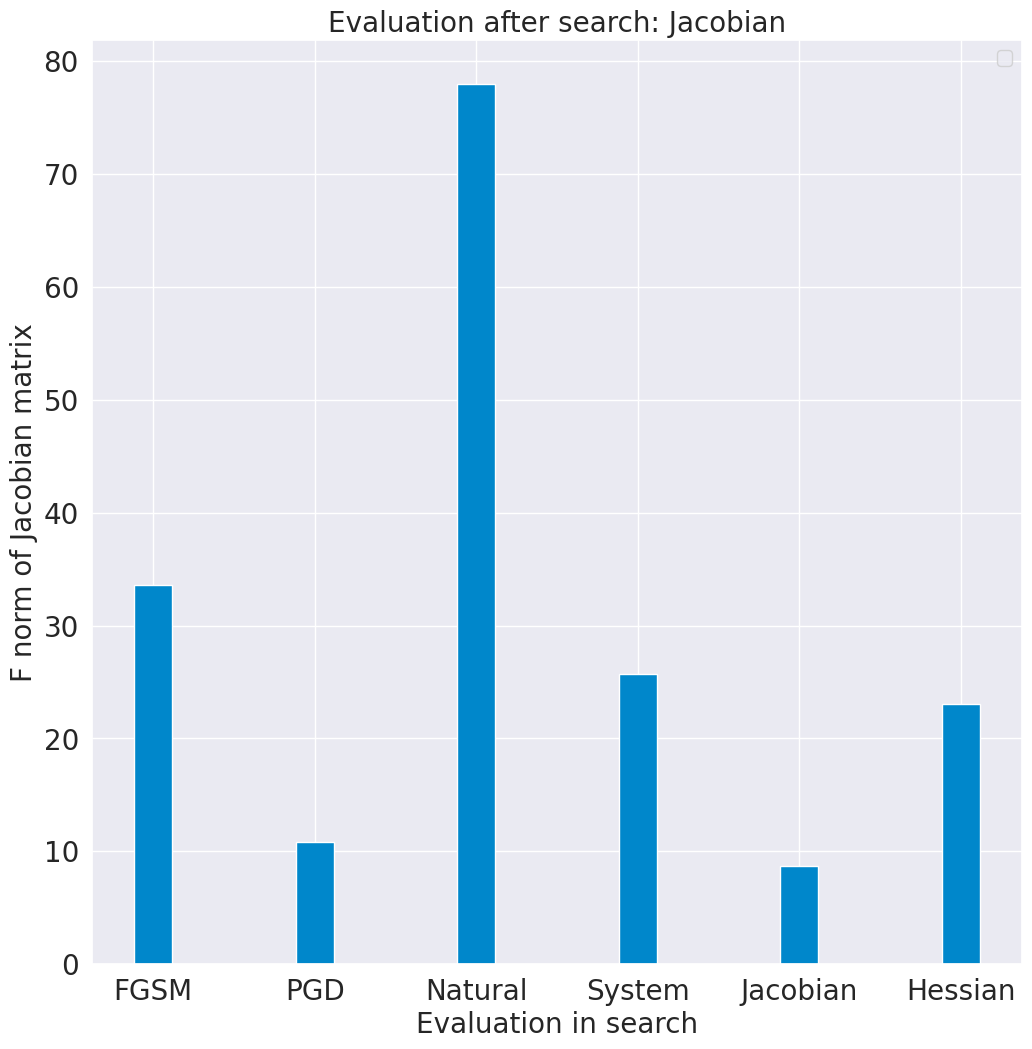}%
	}
	
	\subfloat{\includegraphics[width=2.3in]{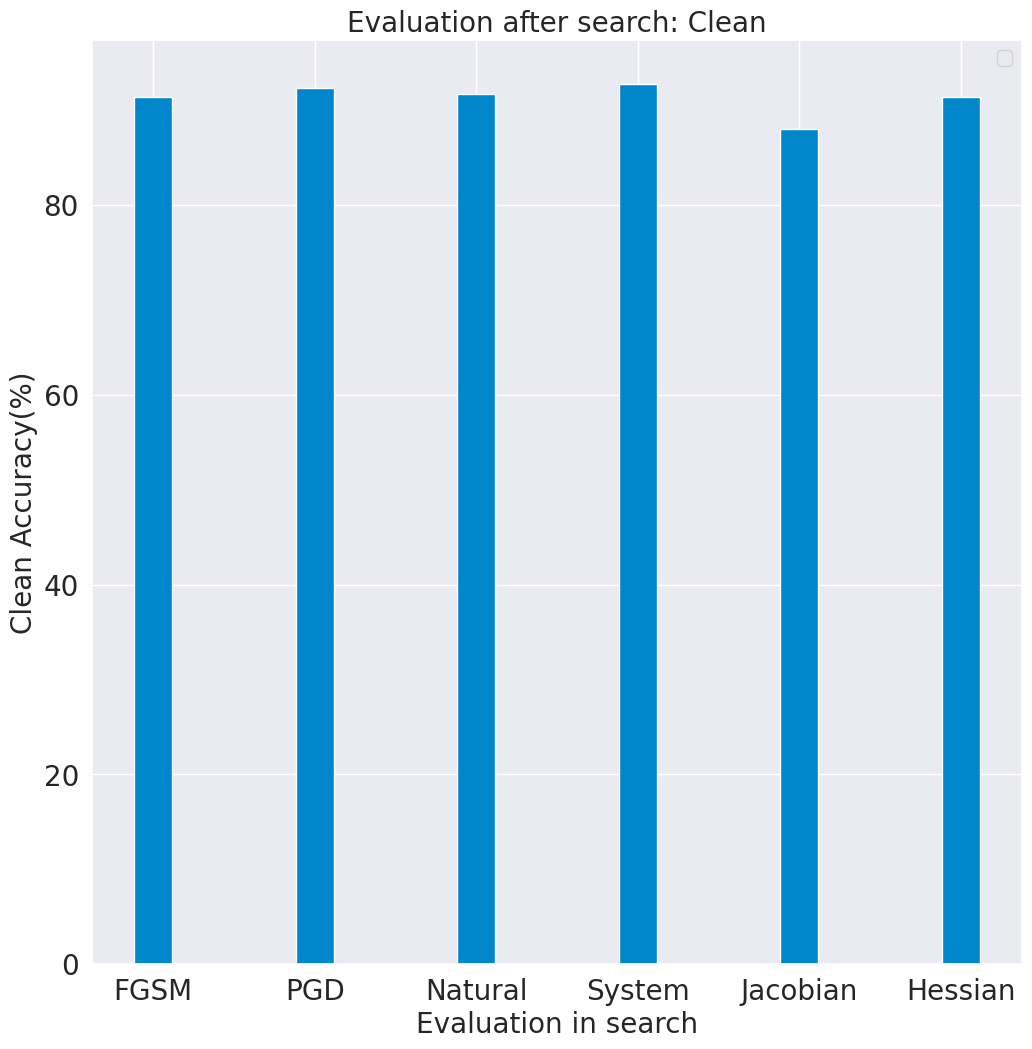}%
	}
	\hfil
	\subfloat{\includegraphics[width=2.3in]{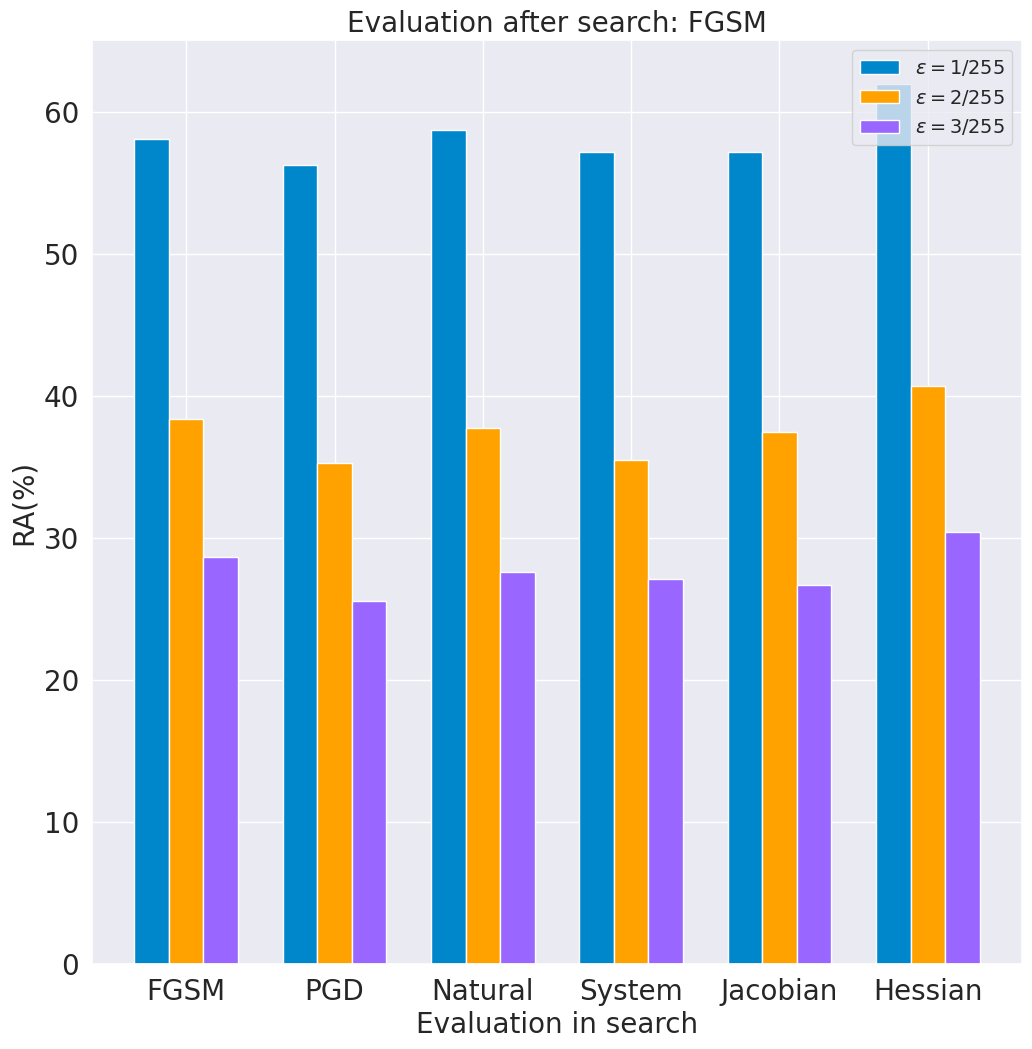}%
		}
	\subfloat{\includegraphics[width=2.3in]{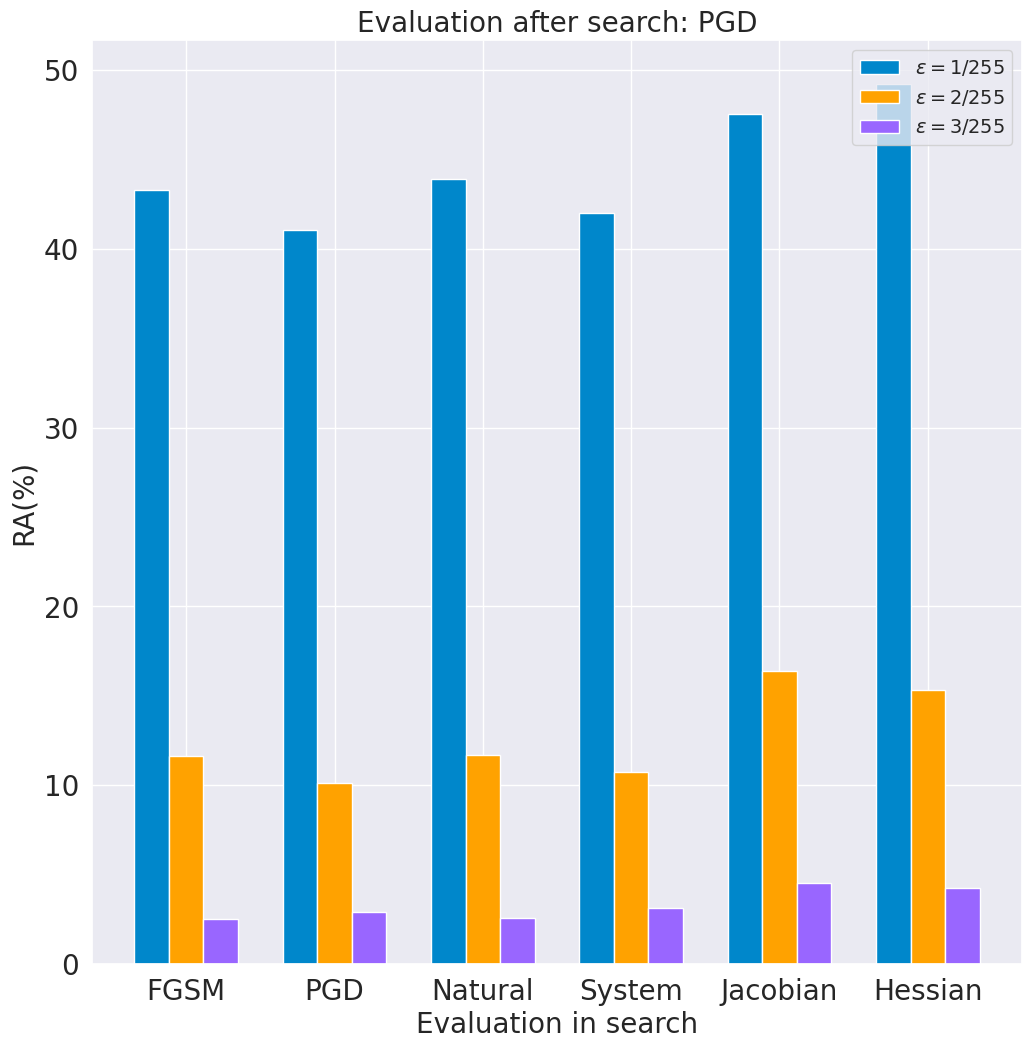}%
	}
	
	\subfloat{\includegraphics[width=2.3in]{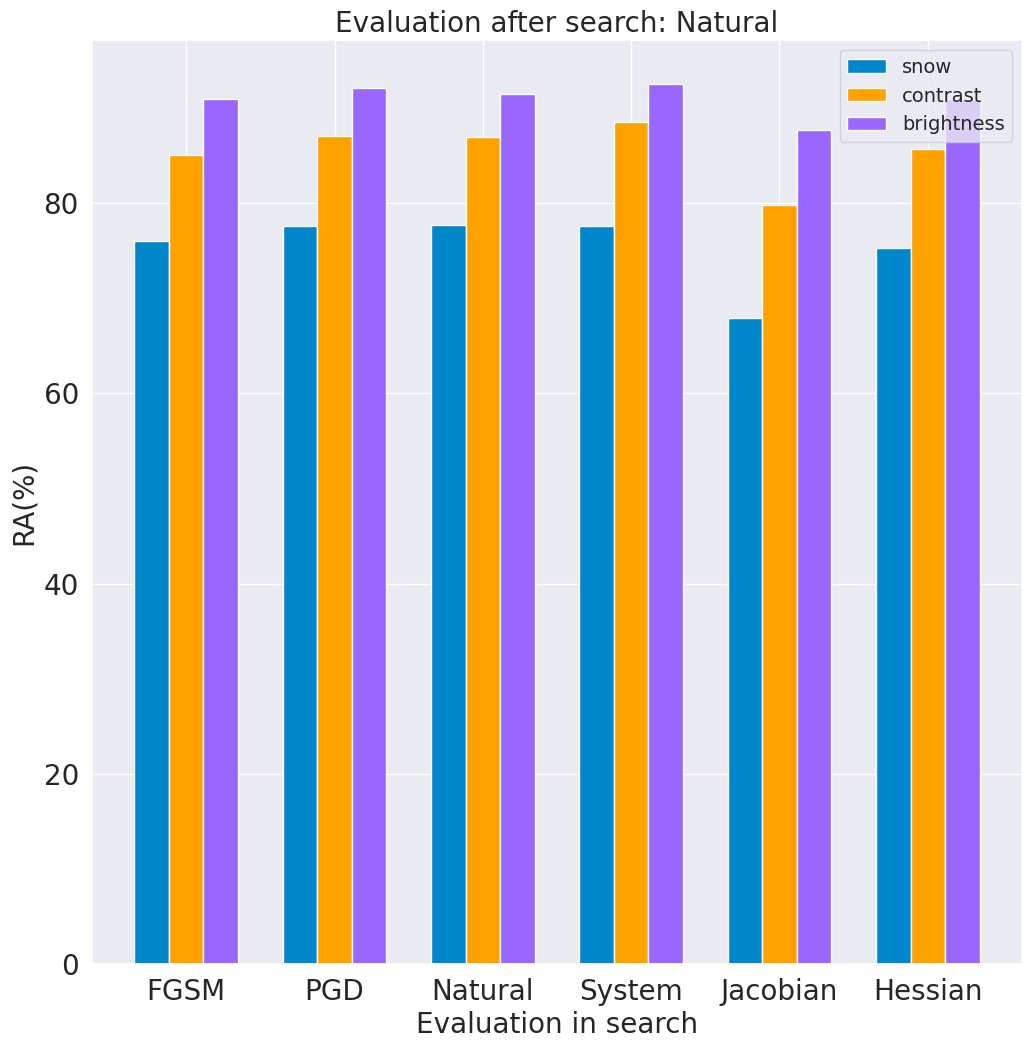}%
	}
	\hfil
	\subfloat{\includegraphics[width=2.3in]{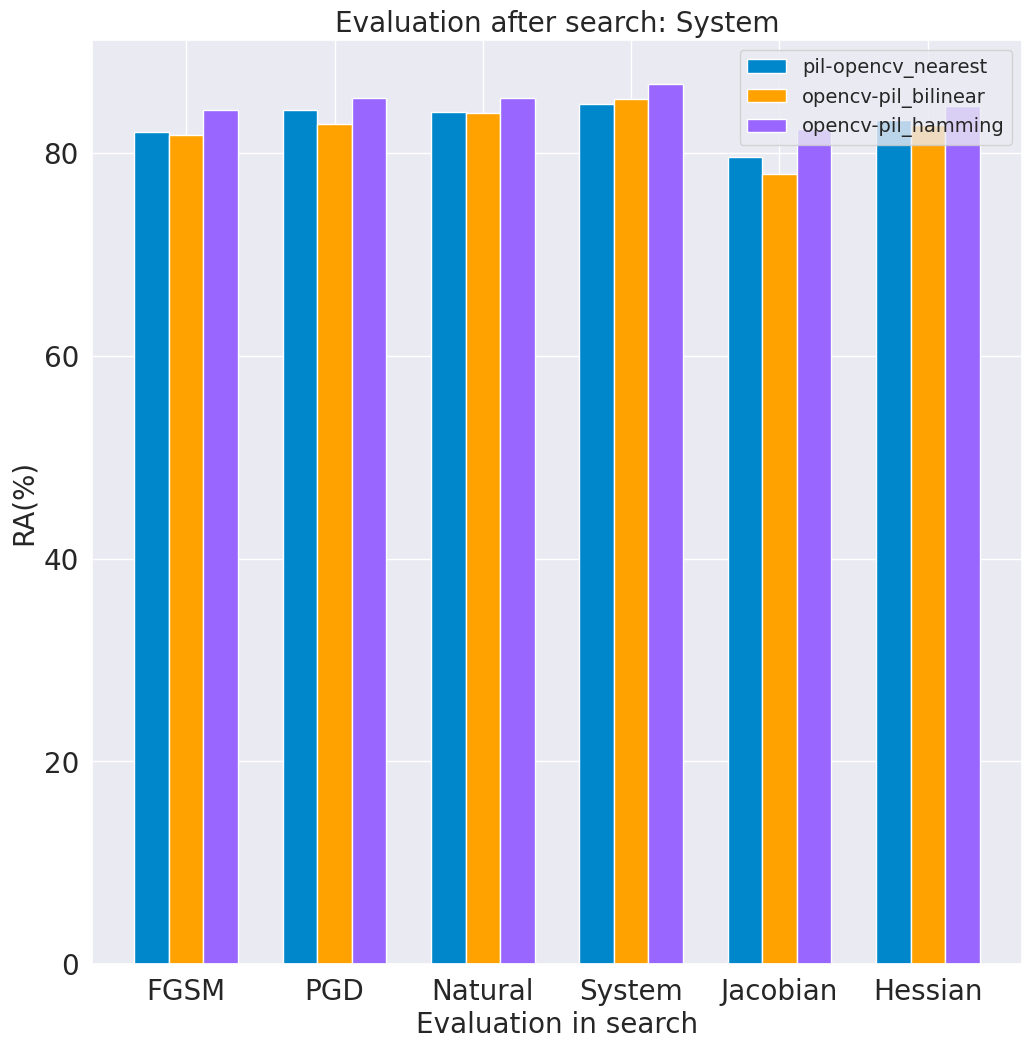}%
	}
	\subfloat{\includegraphics[width=2.3in]{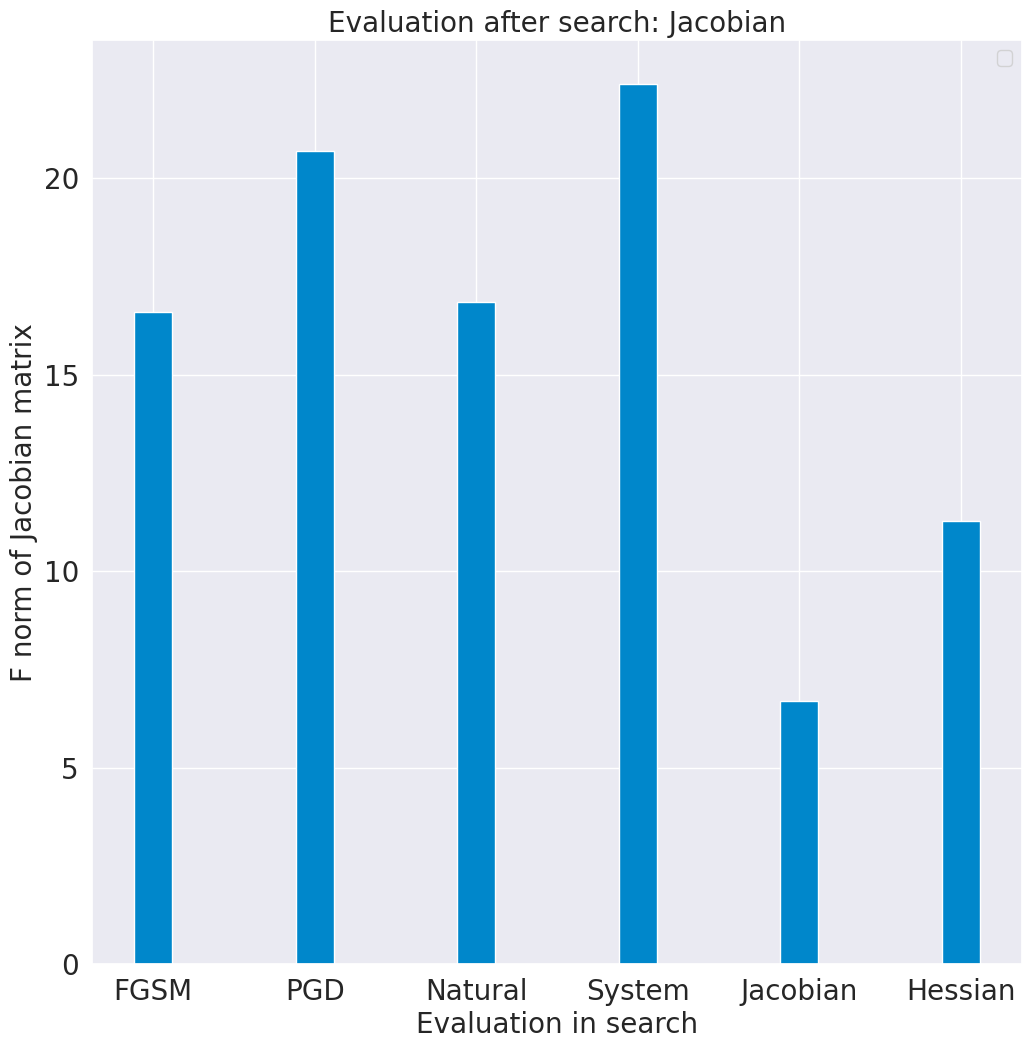}%
}
	\caption{The performance of the searched architectures by DARTS (first two rows) and PC-DARTS (last two rows) }
	\label{DARTS}
\end{figure*}

\begin{figure*}[h]
	\centering
	\subfloat{\includegraphics[width=2.3in]{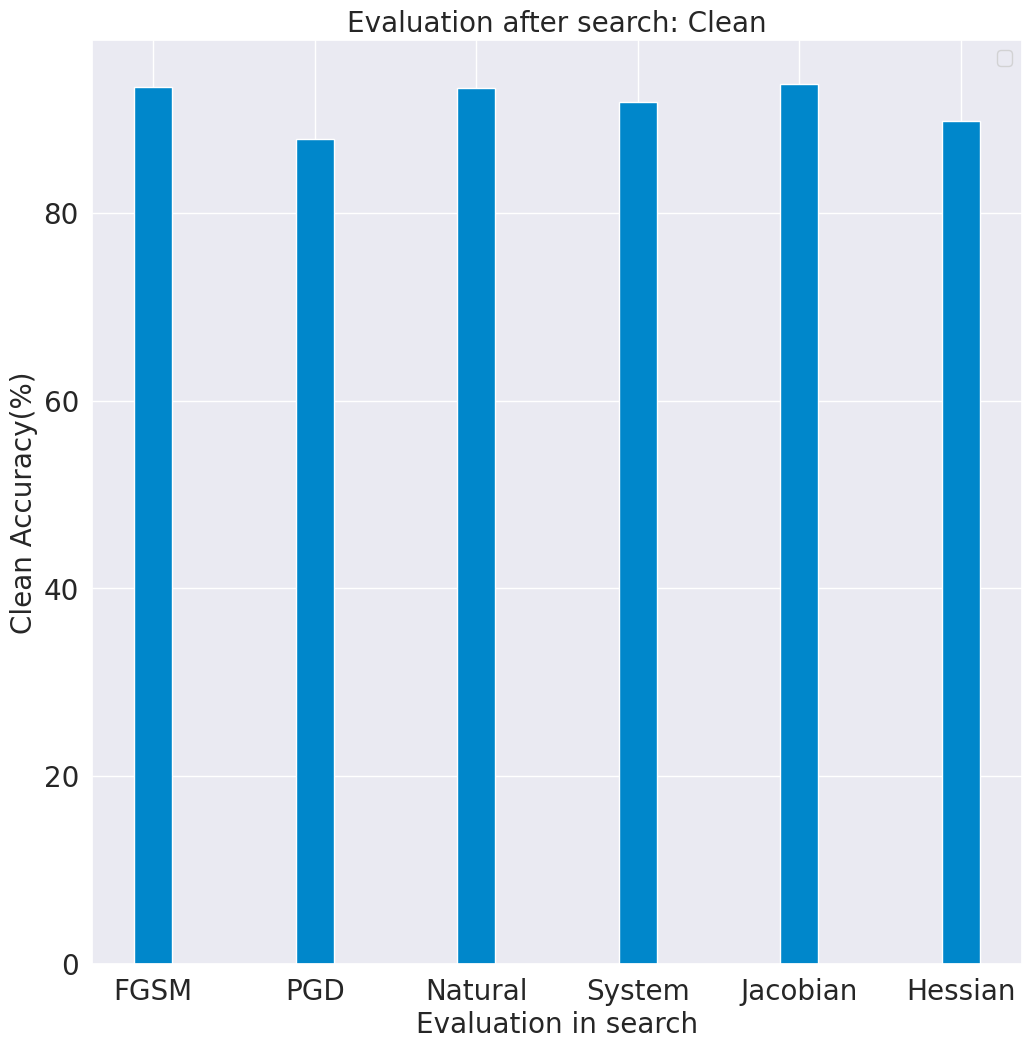}%
}
	\hfil
	\subfloat{\includegraphics[width=2.3in]{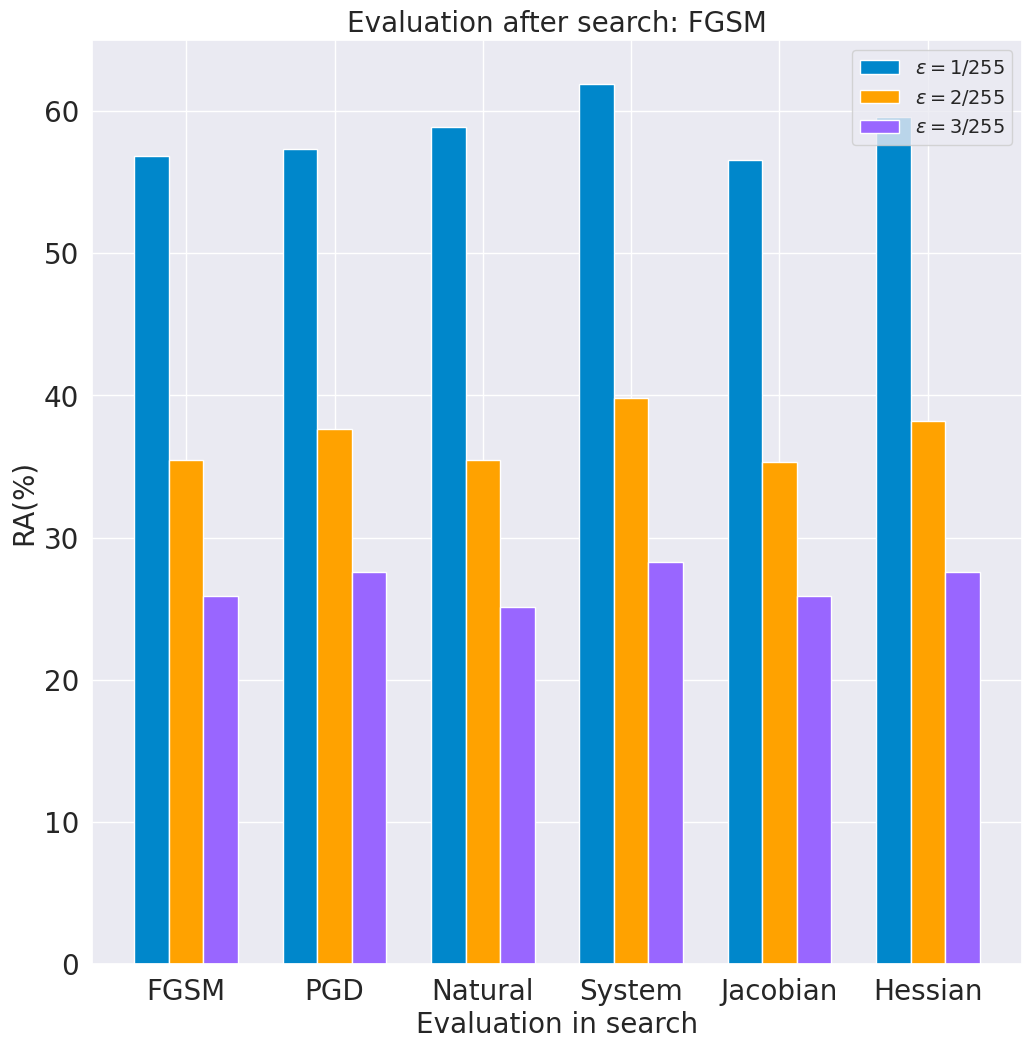}%
}
	\subfloat{\includegraphics[width=2.3in]{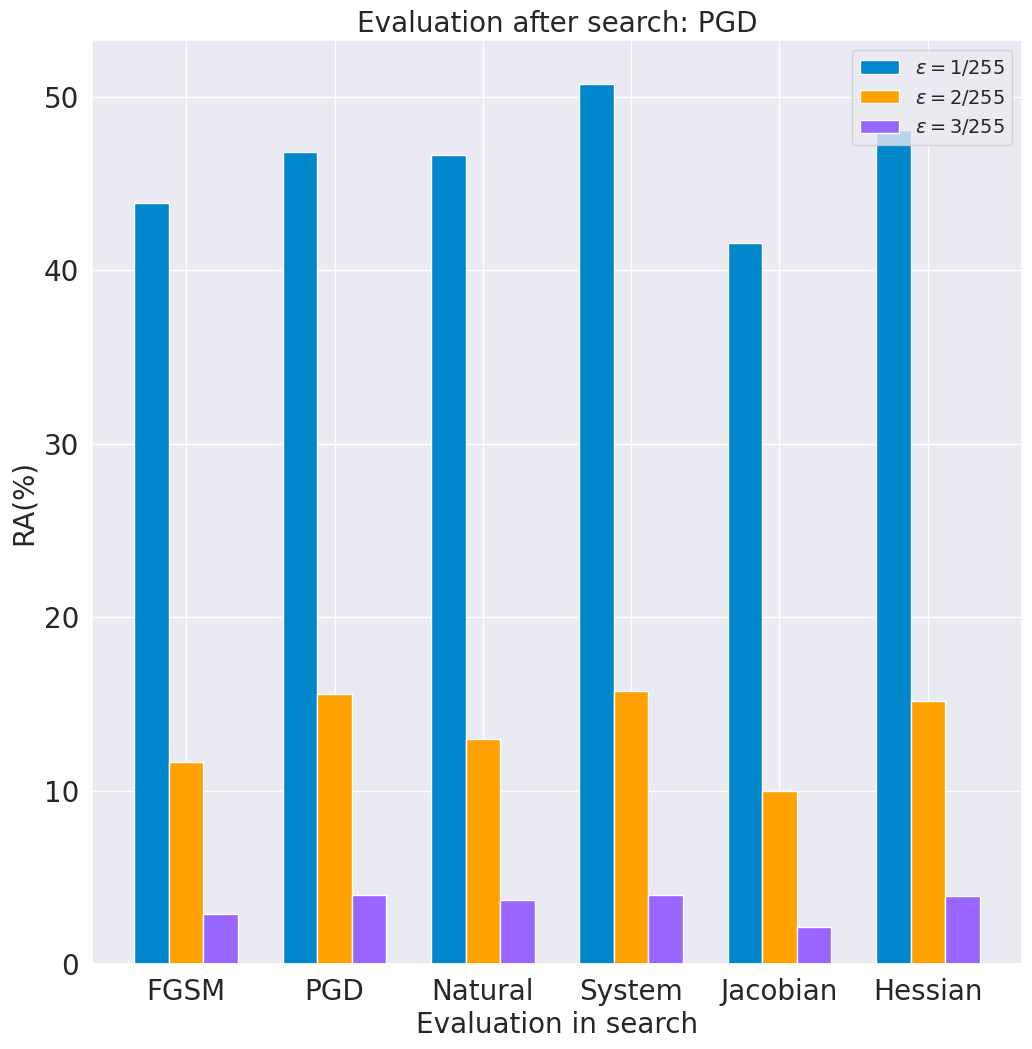}%
}
	
	\subfloat{\includegraphics[width=2.3in]{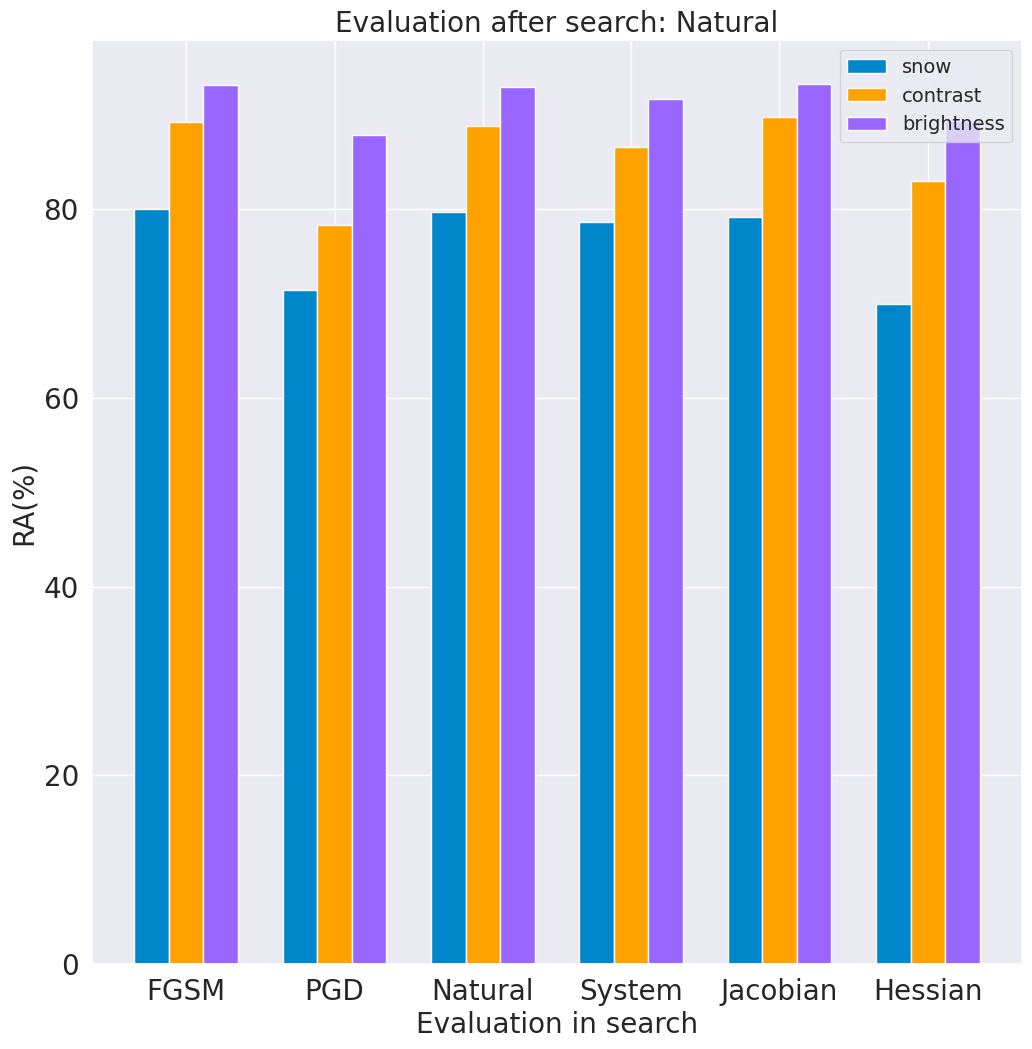}%
}
	\hfil
	\subfloat{\includegraphics[width=2.3in]{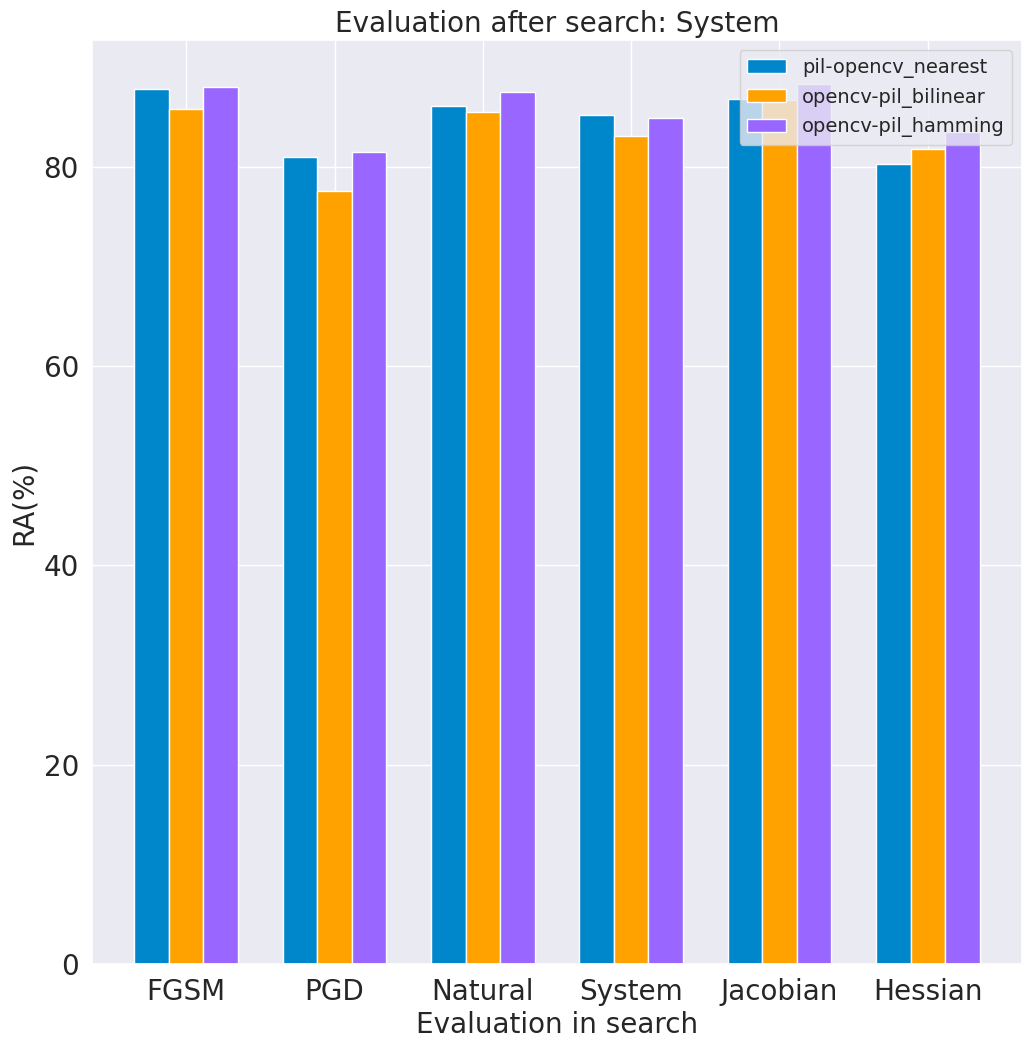}%
}
	\subfloat{\includegraphics[width=2.3in]{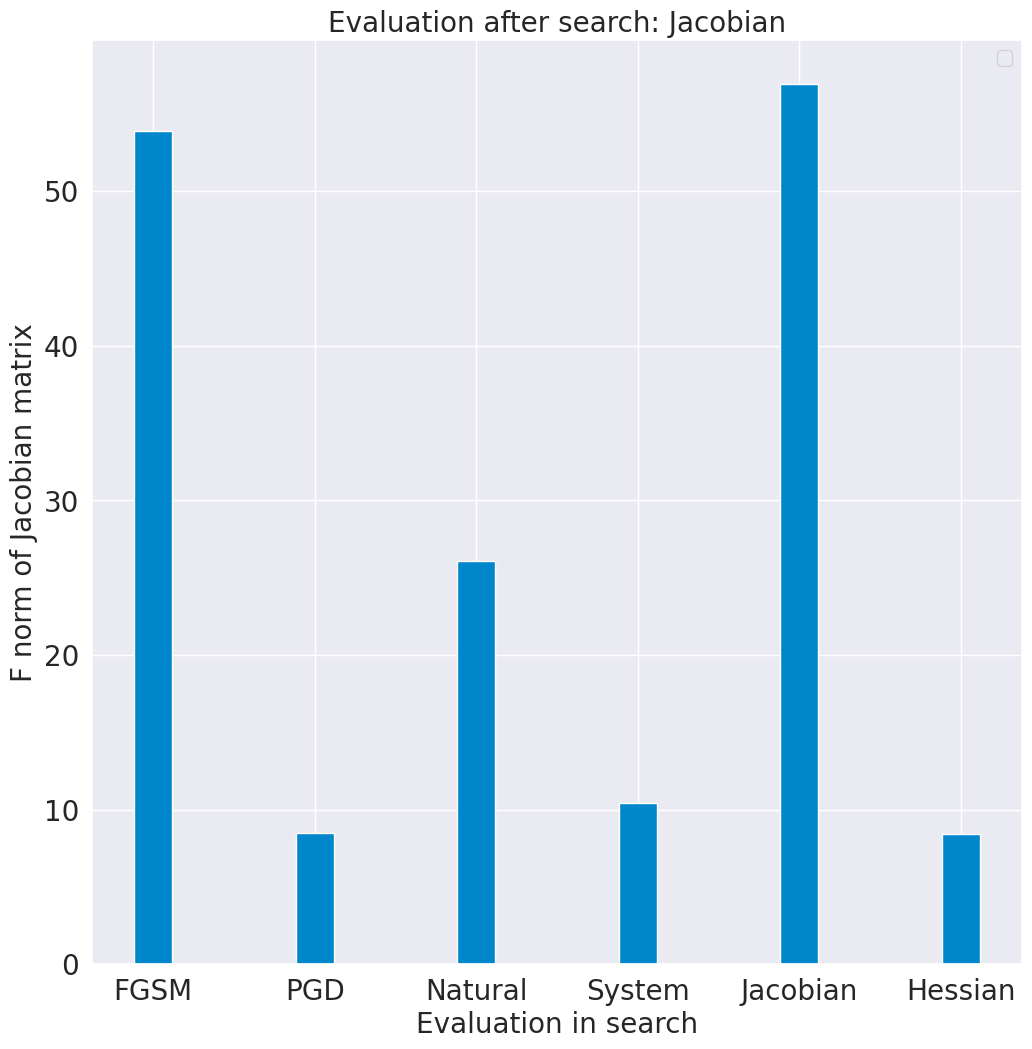}%
}

	\subfloat{\includegraphics[width=2.3in]{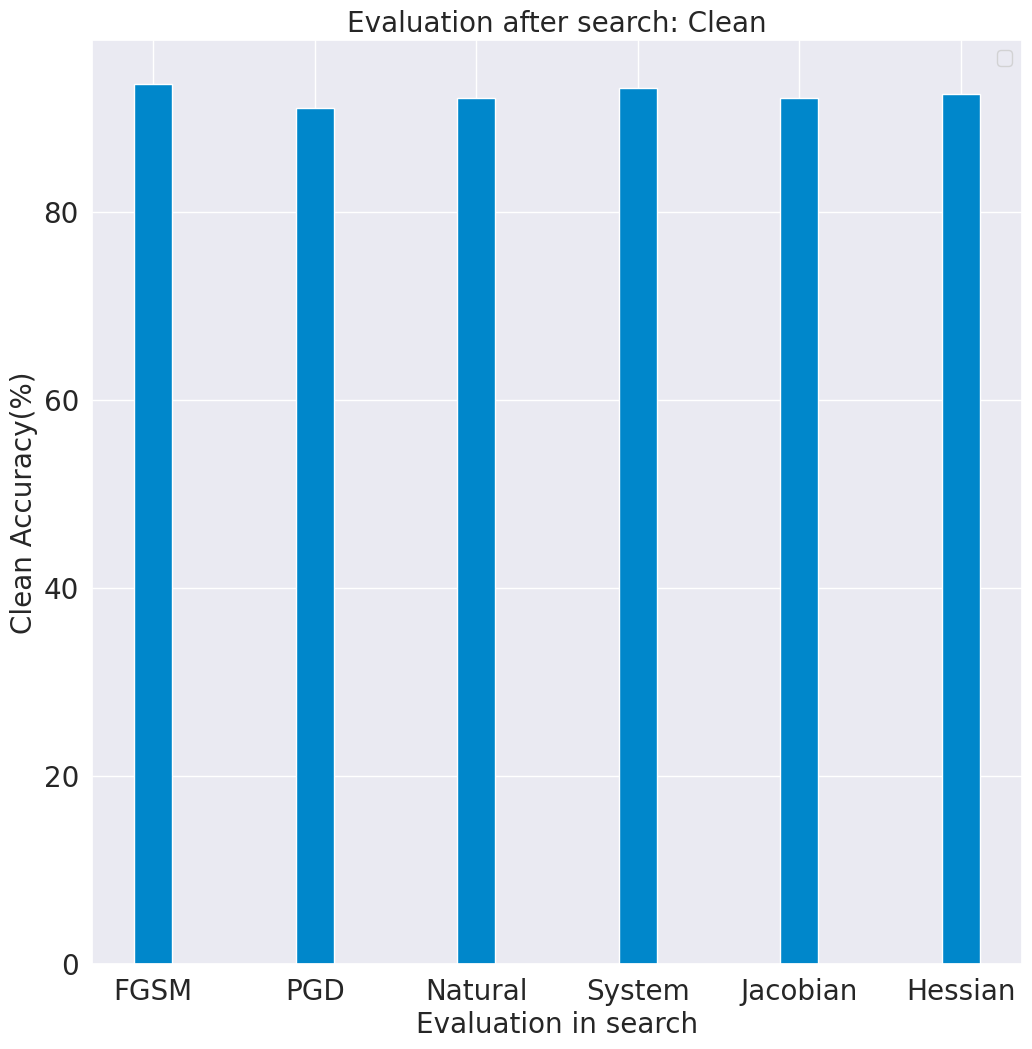}%
}
\hfil
\subfloat{\includegraphics[width=2.3in]{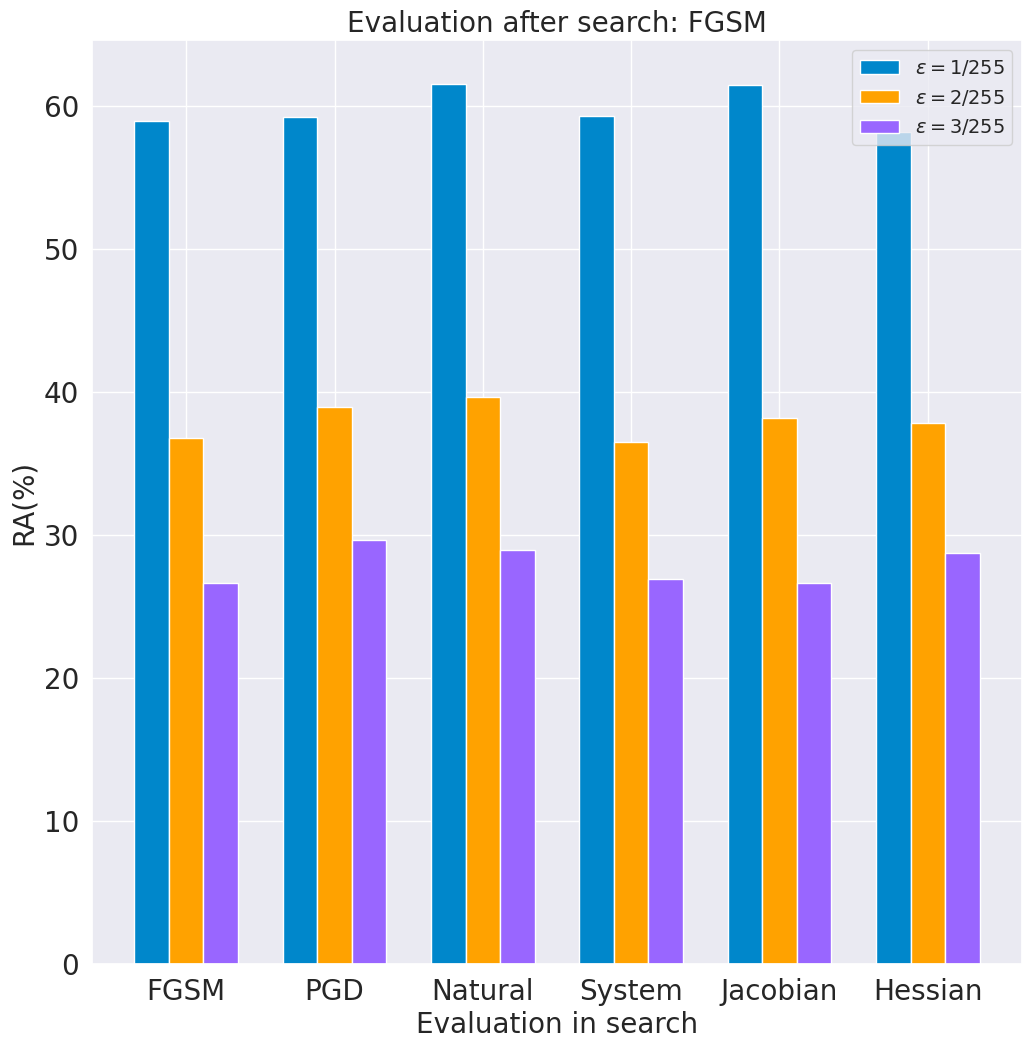}%
}
\subfloat{\includegraphics[width=2.3in]{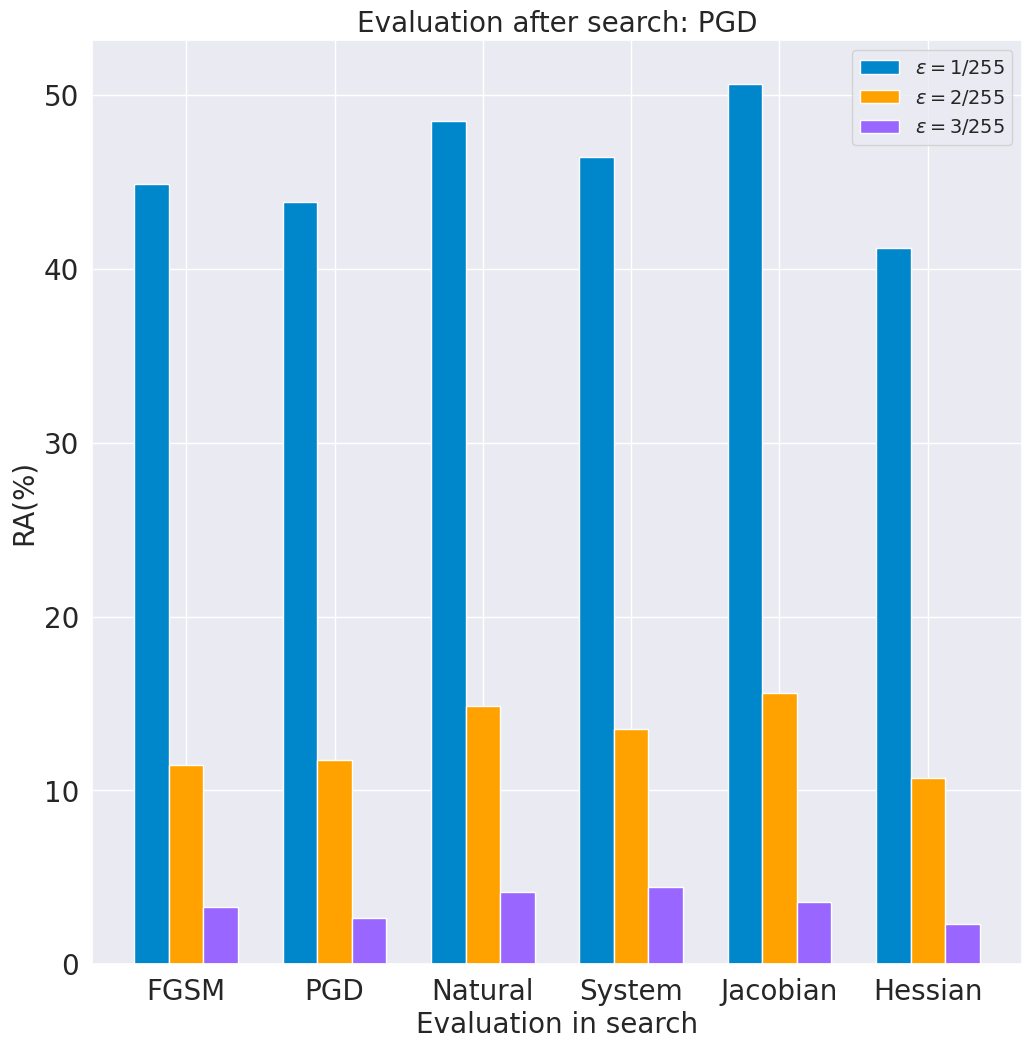}%
}

\subfloat{\includegraphics[width=2.3in]{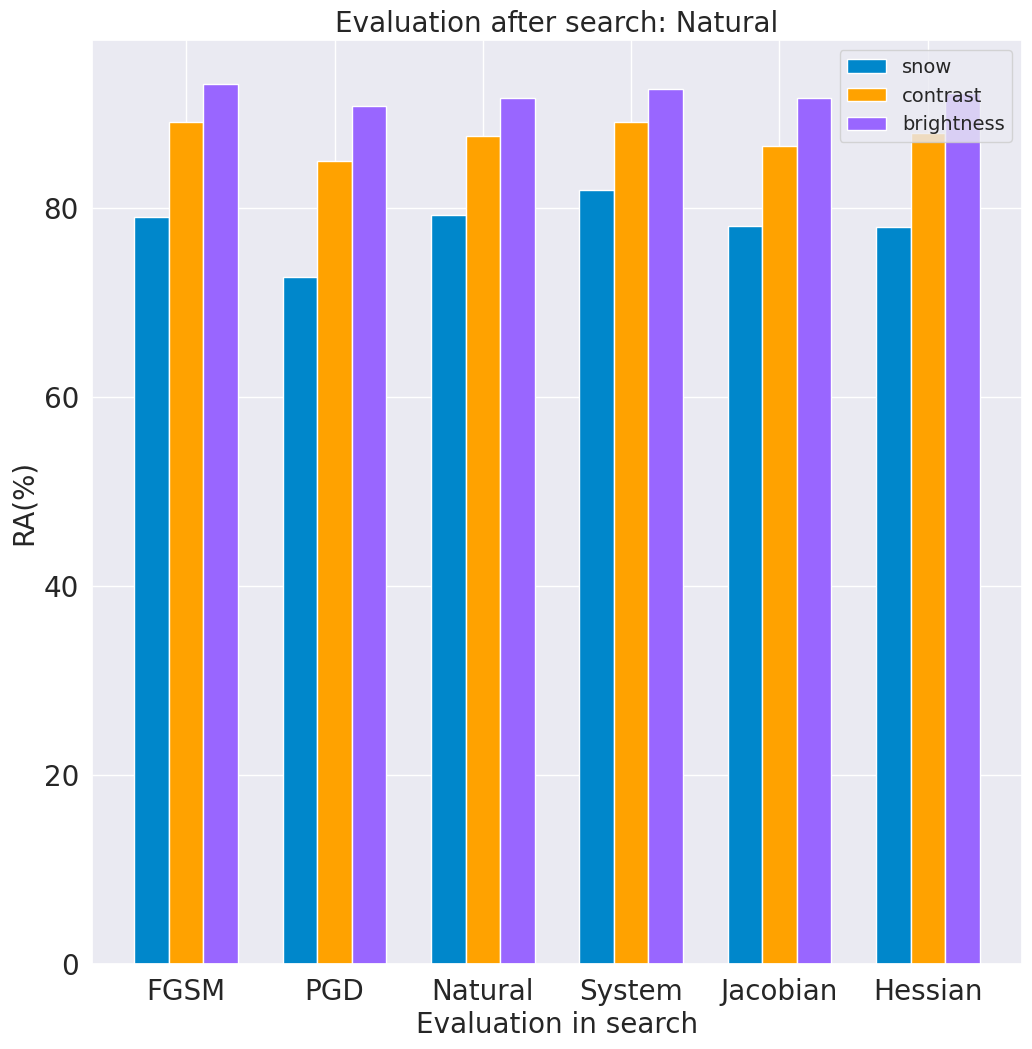}%
}
\hfil
\subfloat{\includegraphics[width=2.3in]{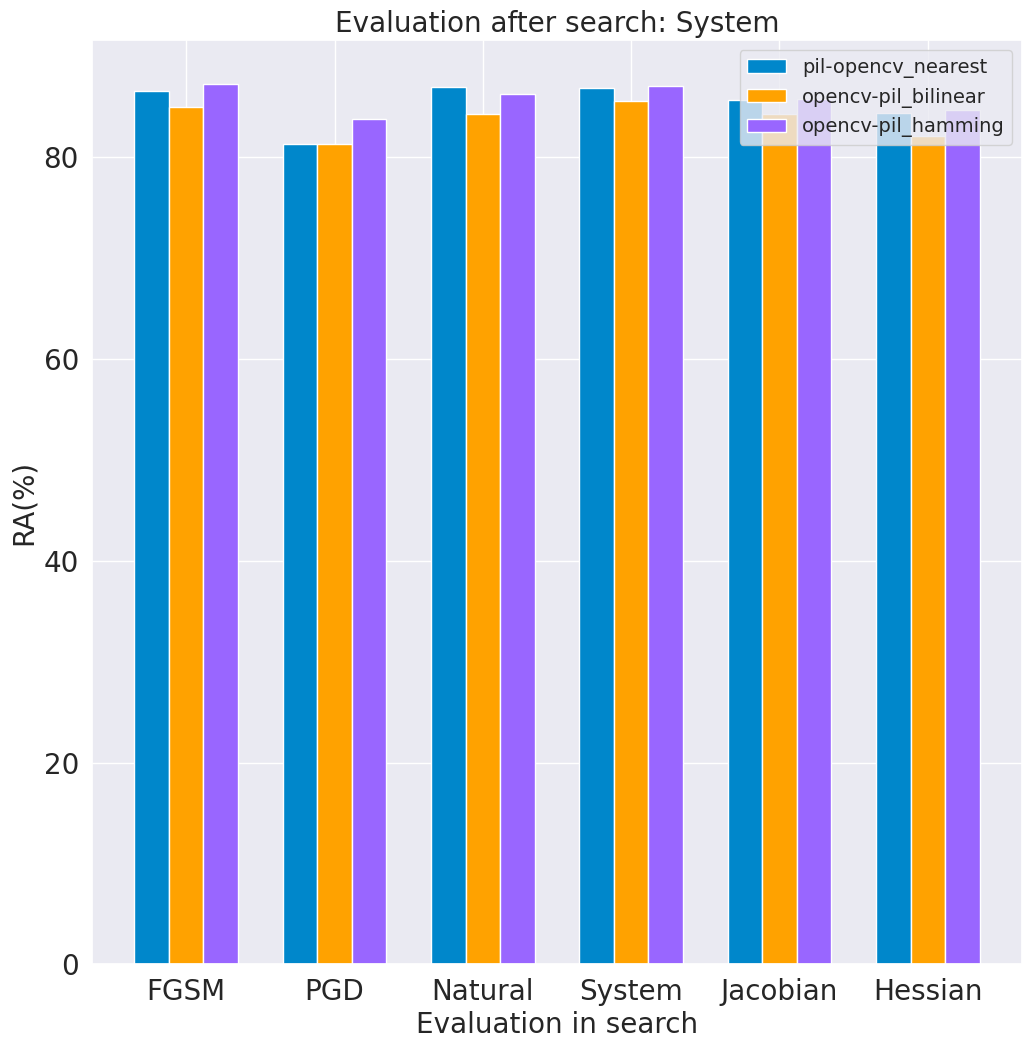}%
}
\subfloat{\includegraphics[width=2.3in]{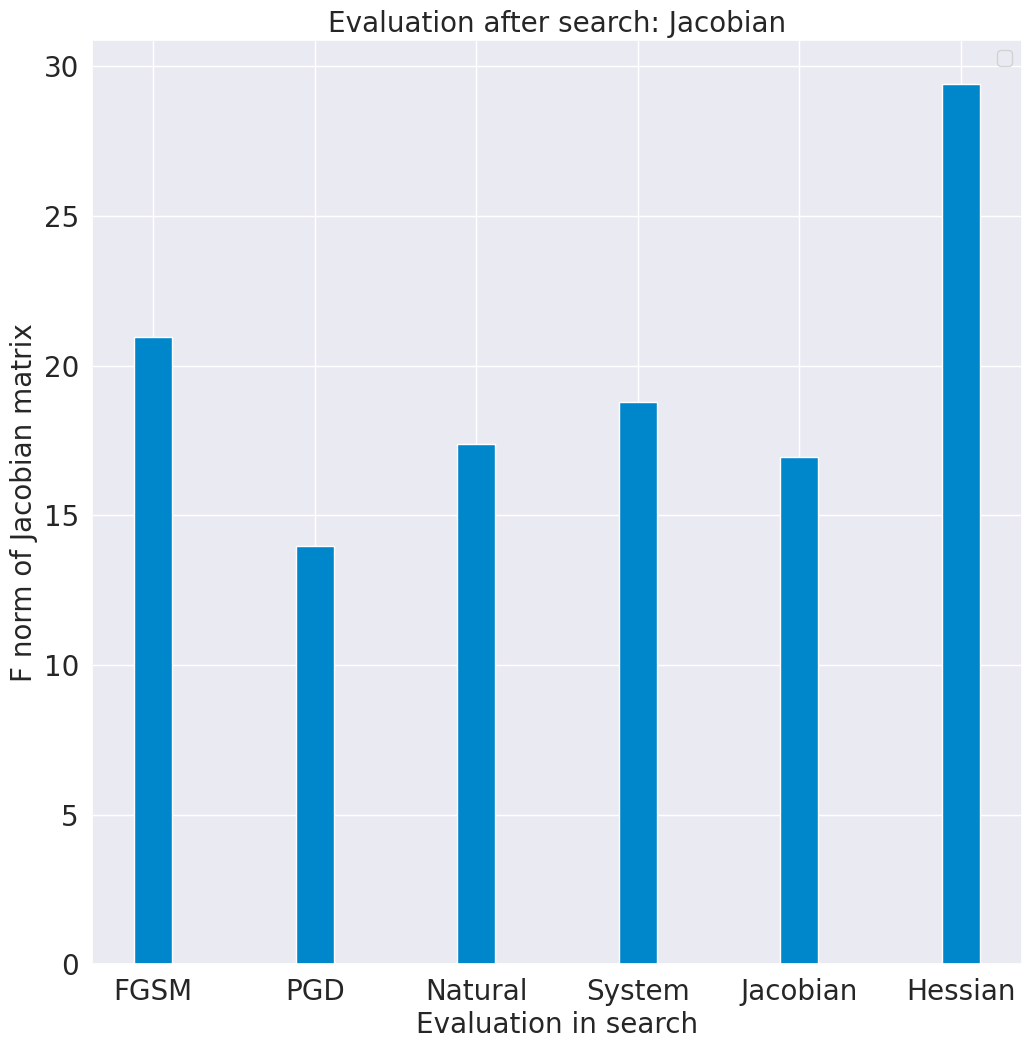}%
}

	\caption{The performance of the searched architectures by NASP (first two rows) and FairDARTS (last two rows)}
	\label{NASP}
\end{figure*}

\begin{figure*}[h]
	\centering
	\subfloat{\includegraphics[width=2.3in]{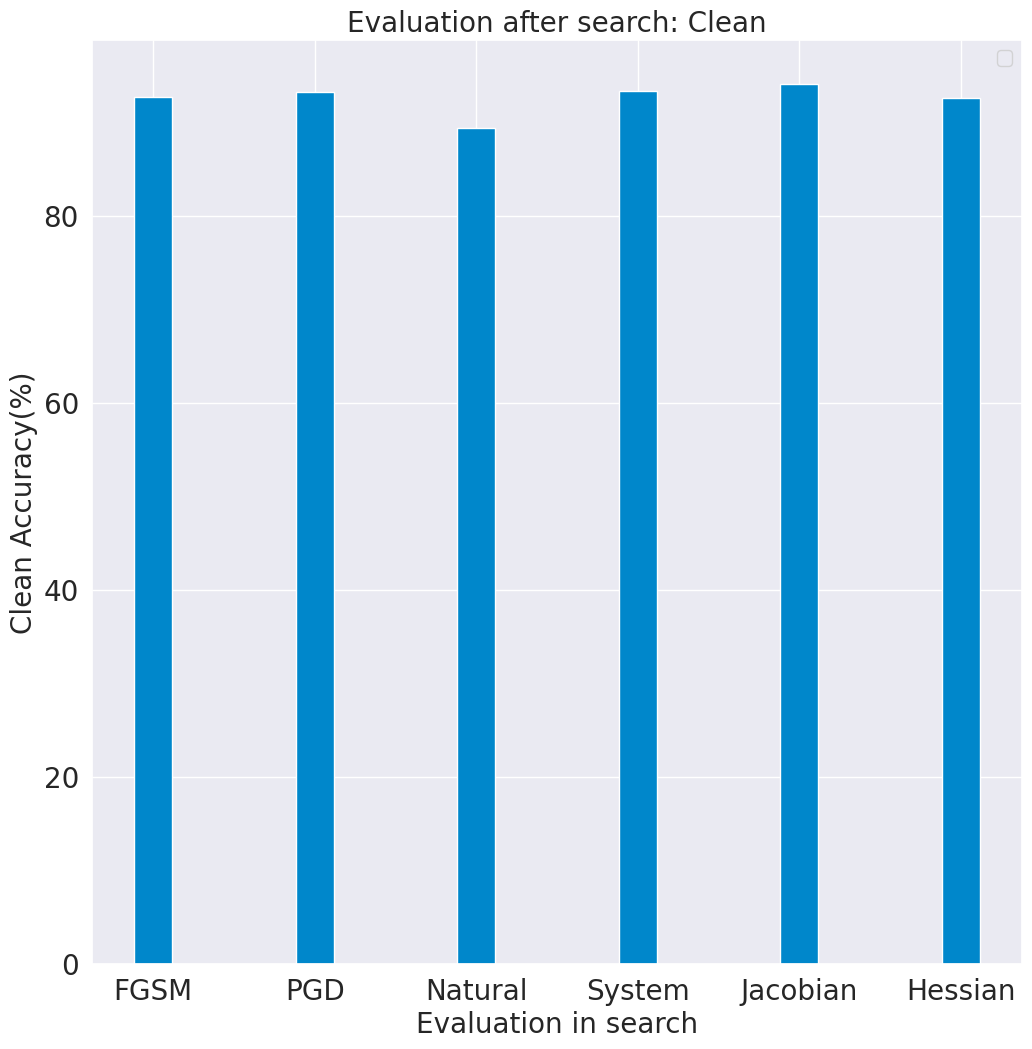}%
	}
	\hfil
	\subfloat{\includegraphics[width=2.3in]{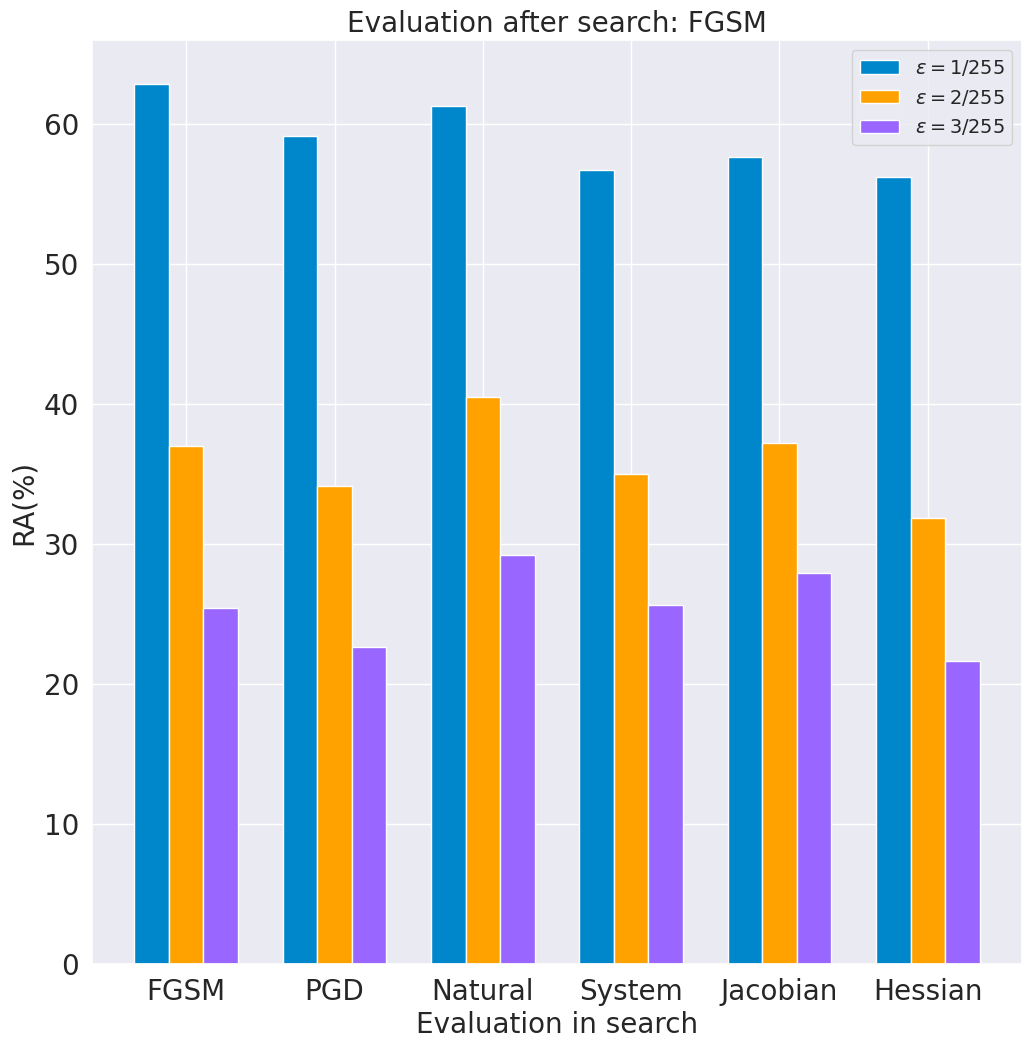}%
	}
	\subfloat{\includegraphics[width=2.3in]{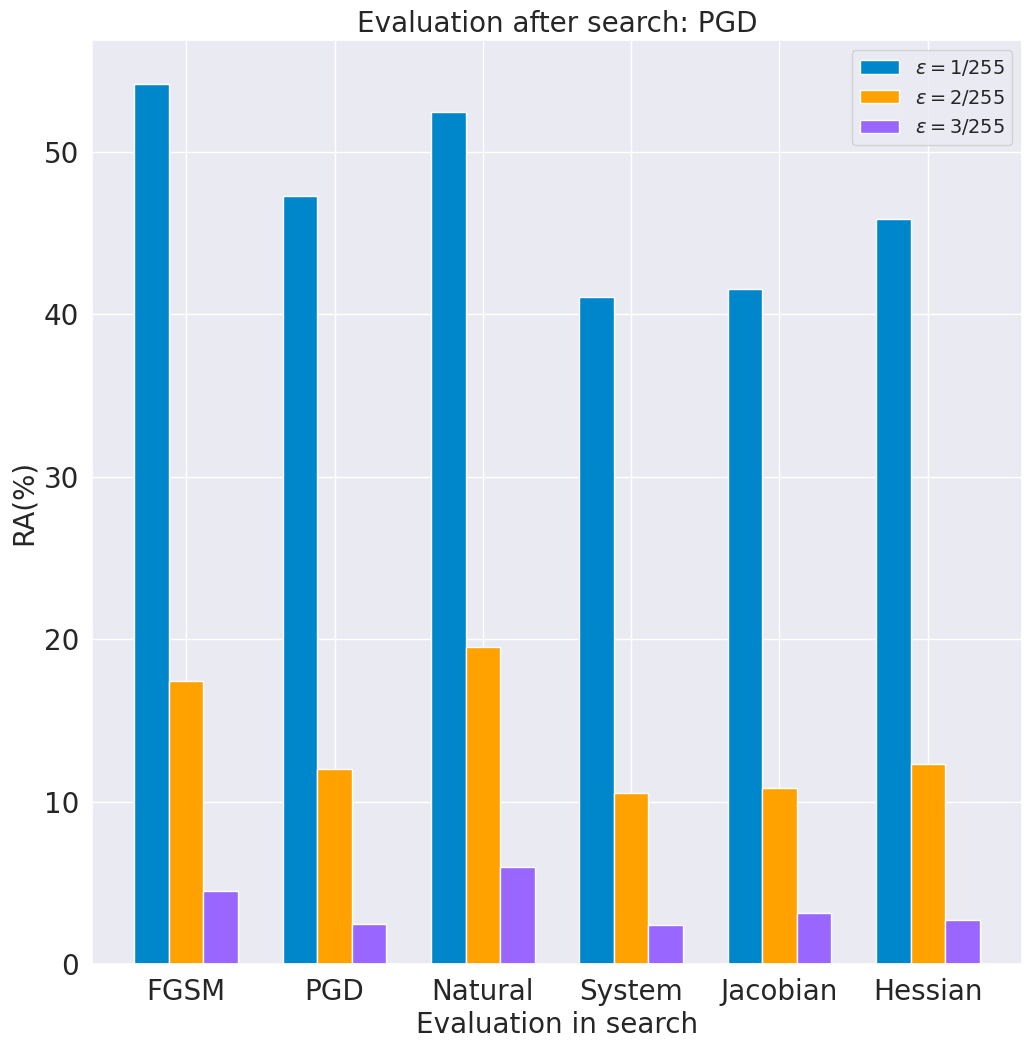}%
	}
	
	\subfloat{\includegraphics[width=2.3in]{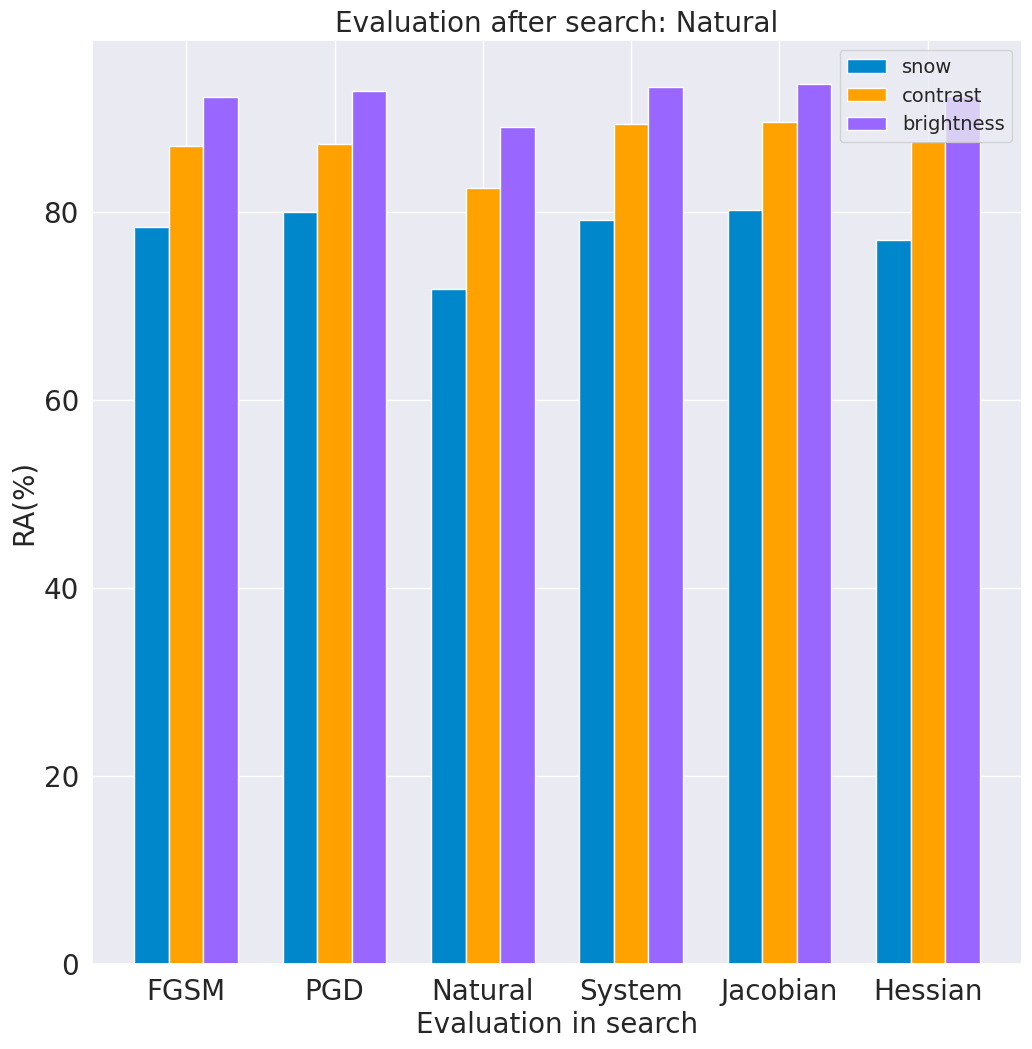}%
	}
	\hfil
	\subfloat{\includegraphics[width=2.3in]{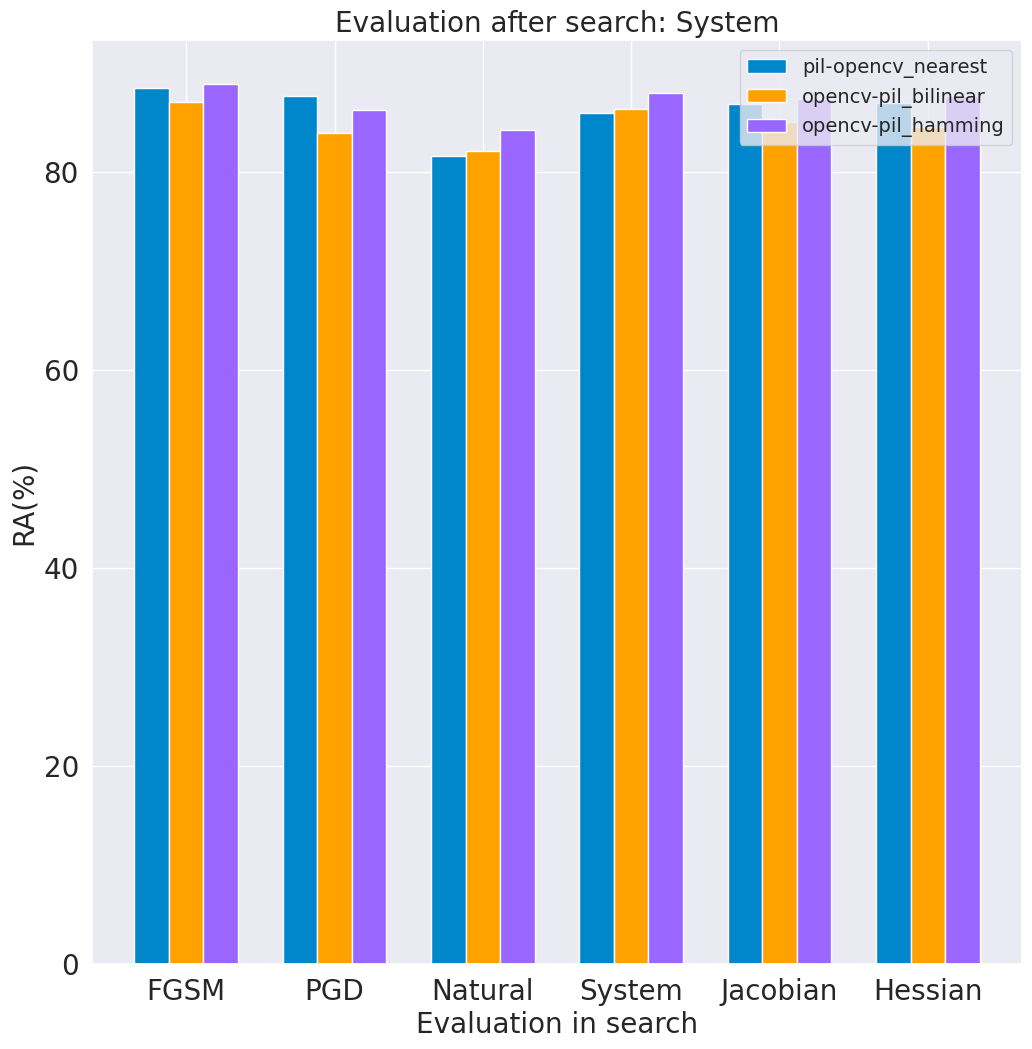}%
	}
	\subfloat{\includegraphics[width=2.3in]{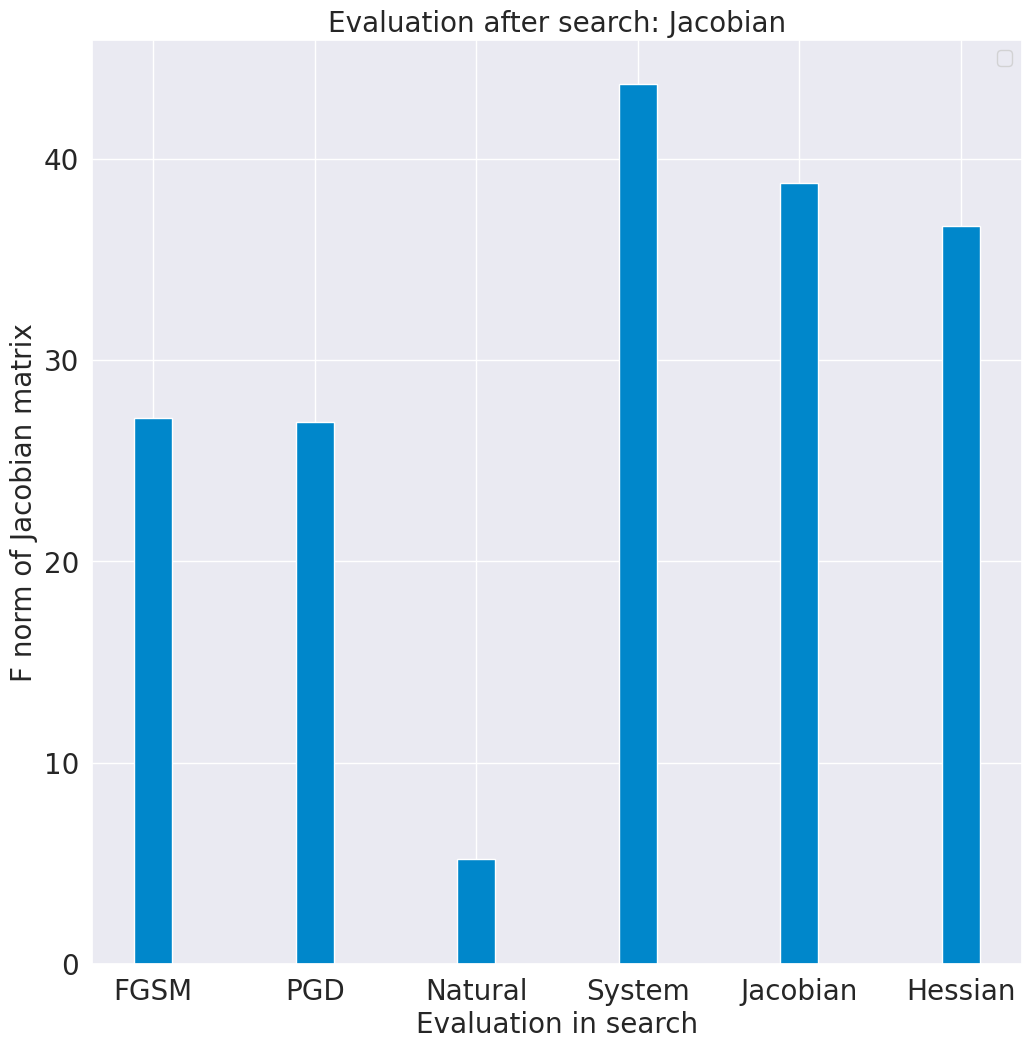}%
	}
	
	\subfloat{\includegraphics[width=2.3in]{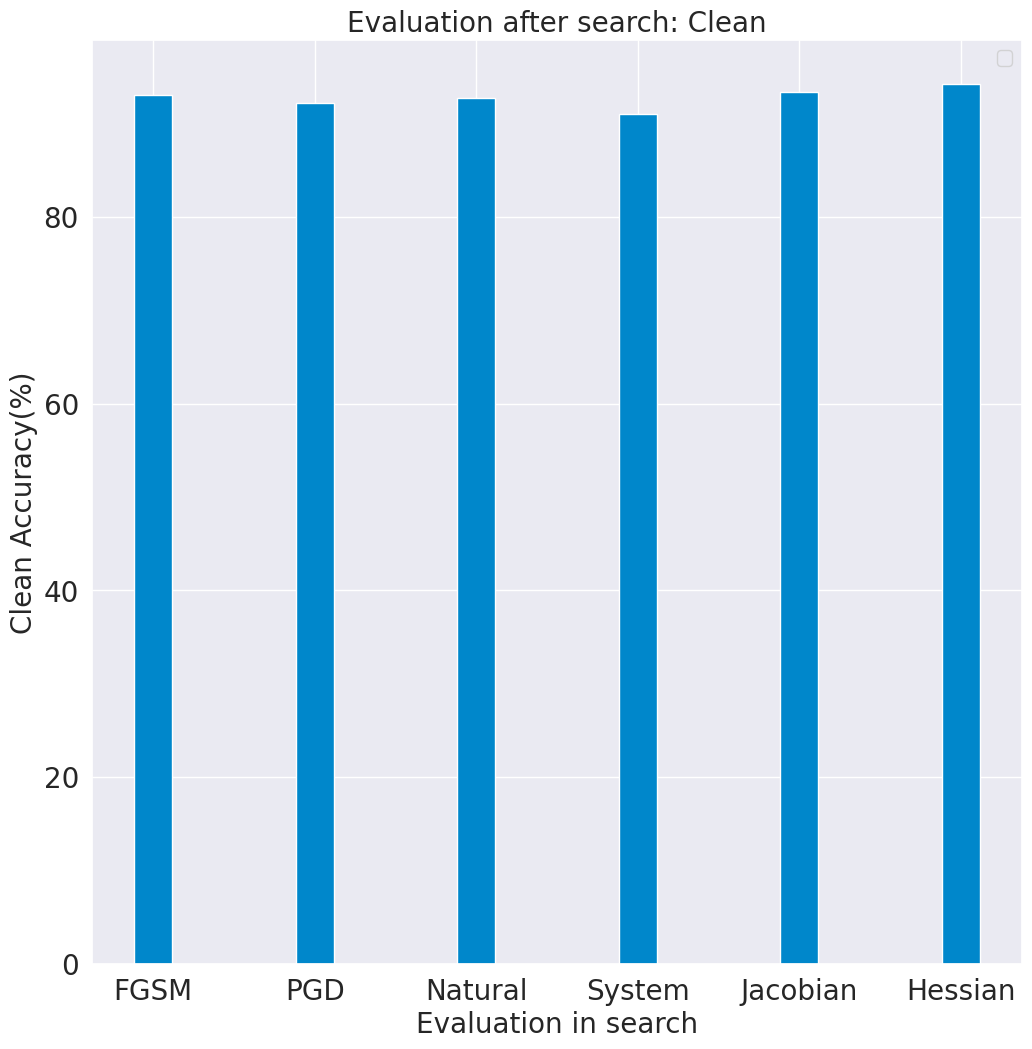}%
	}
	\hfil
	\subfloat{\includegraphics[width=2.3in]{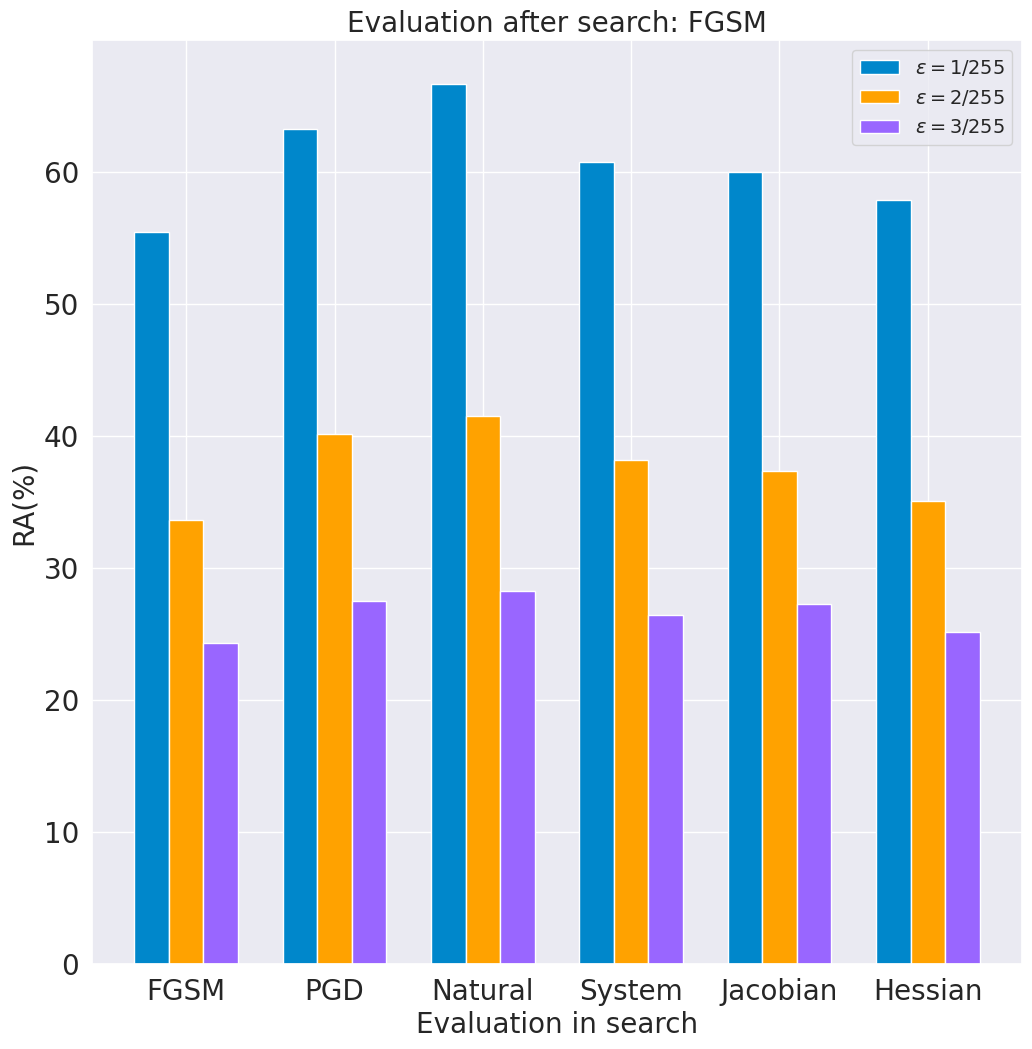}%
	}
	\subfloat{\includegraphics[width=2.3in]{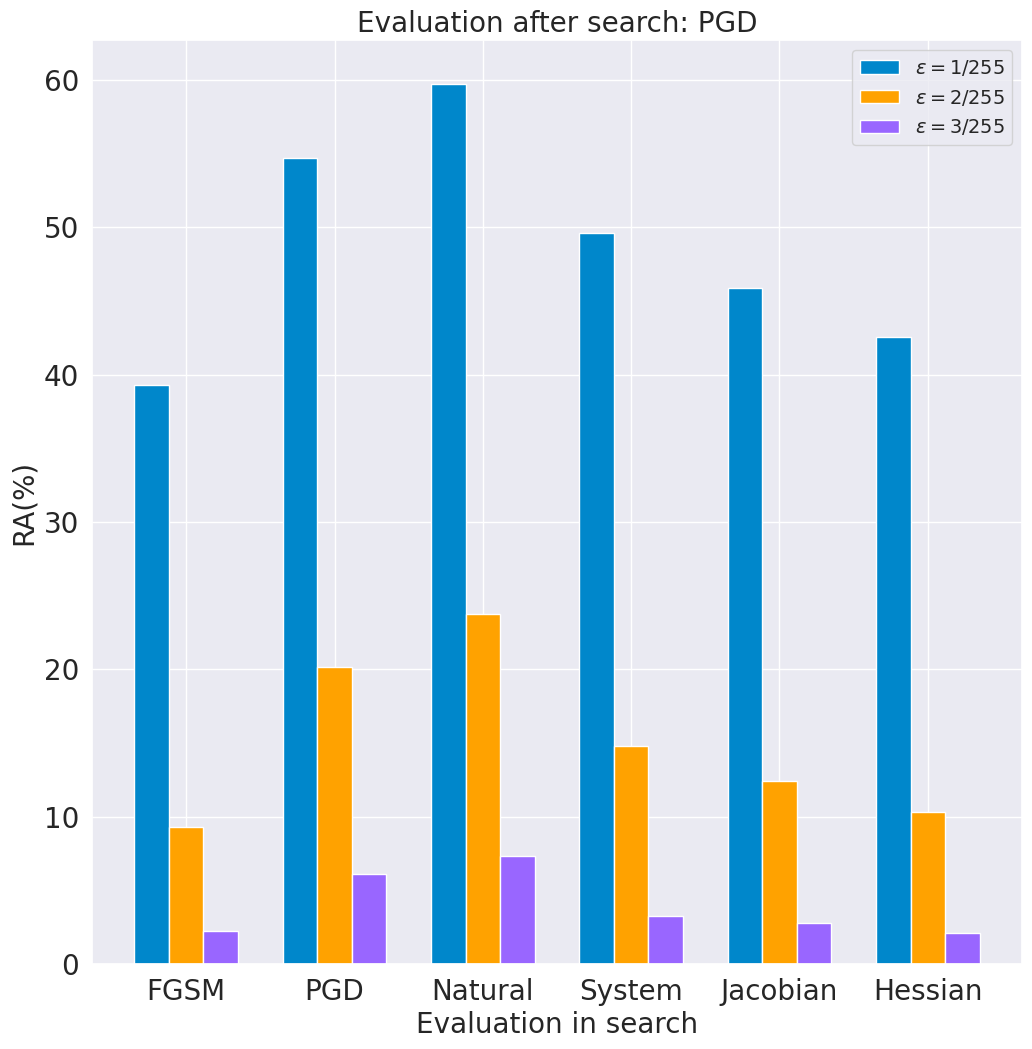}%
	}
	
	\subfloat{\includegraphics[width=2.3in]{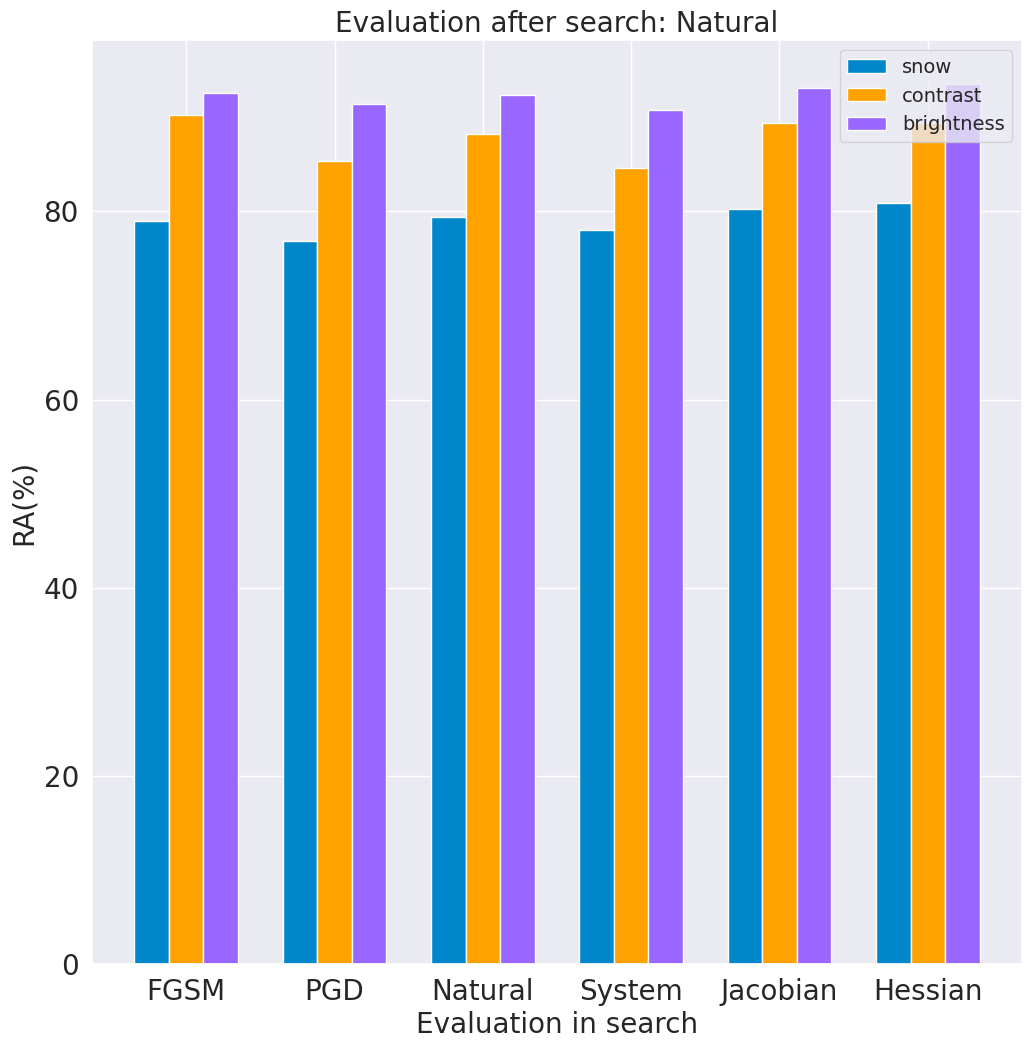}%
	}
	\hfil
	\subfloat{\includegraphics[width=2.3in]{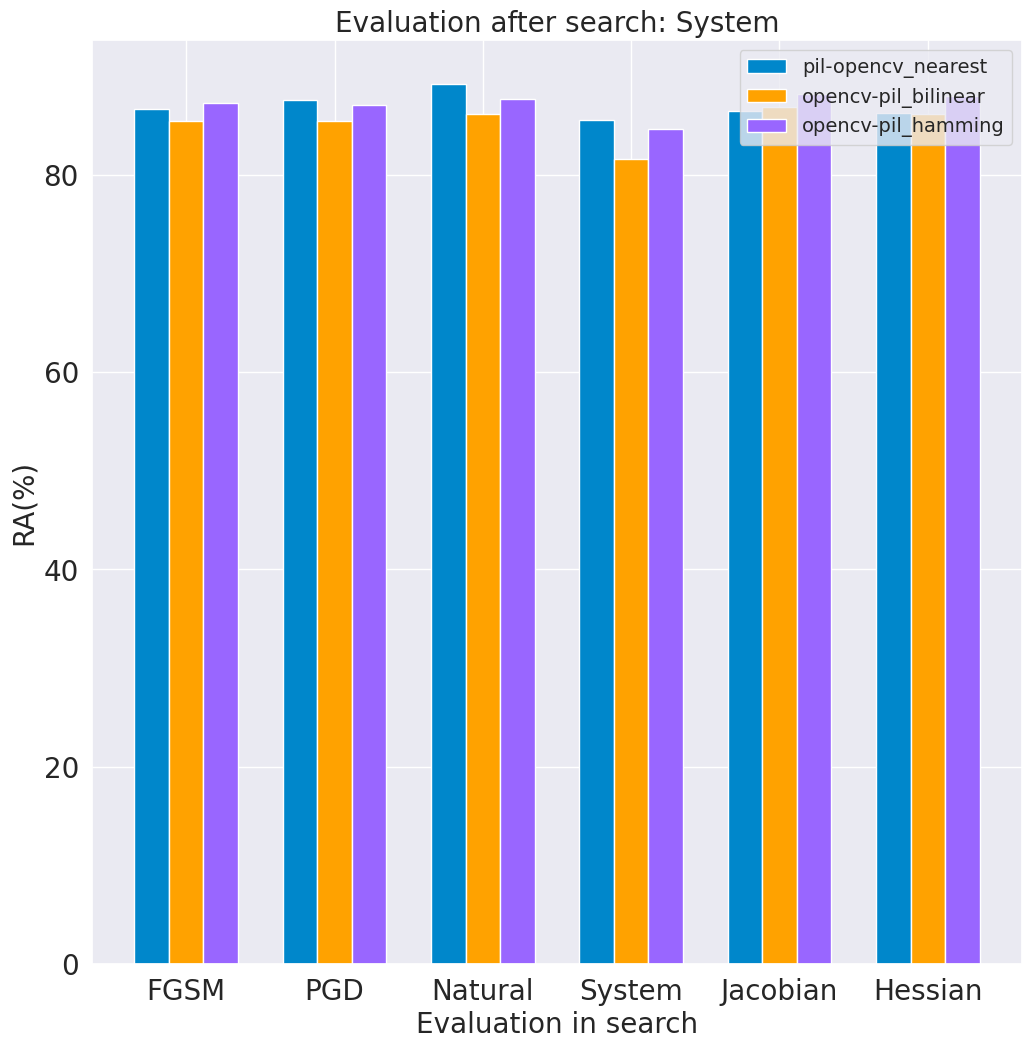}%
	}
	\subfloat{\includegraphics[width=2.3in]{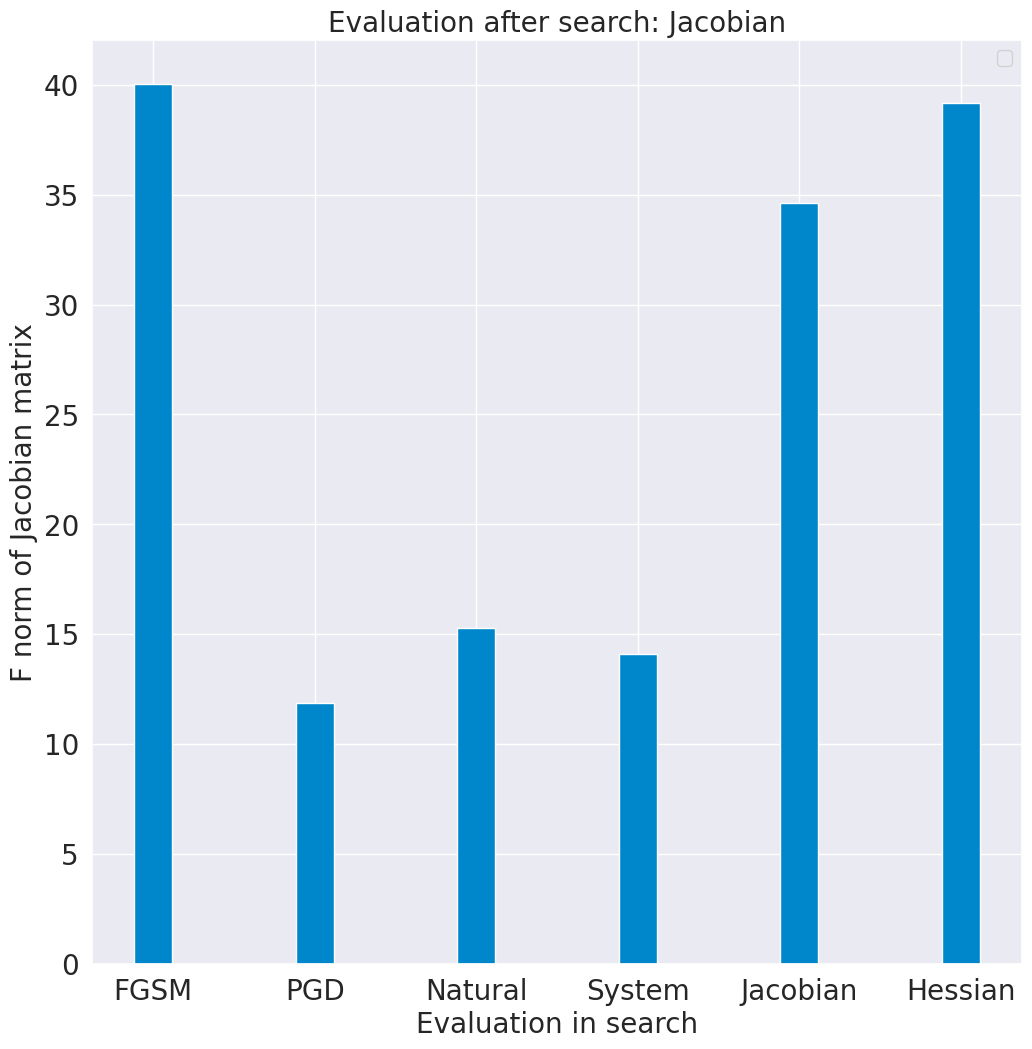}%
	}
	
	\caption{The performance of the searched architectures by SmoothDARTS (first two rows) and Weight Sharing based Random Search (last two rows)}
	\label{SmoothDARTS}
\end{figure*}

\begin{figure*}[h]
	\centering
	\subfloat{\includegraphics[width=2.3in]{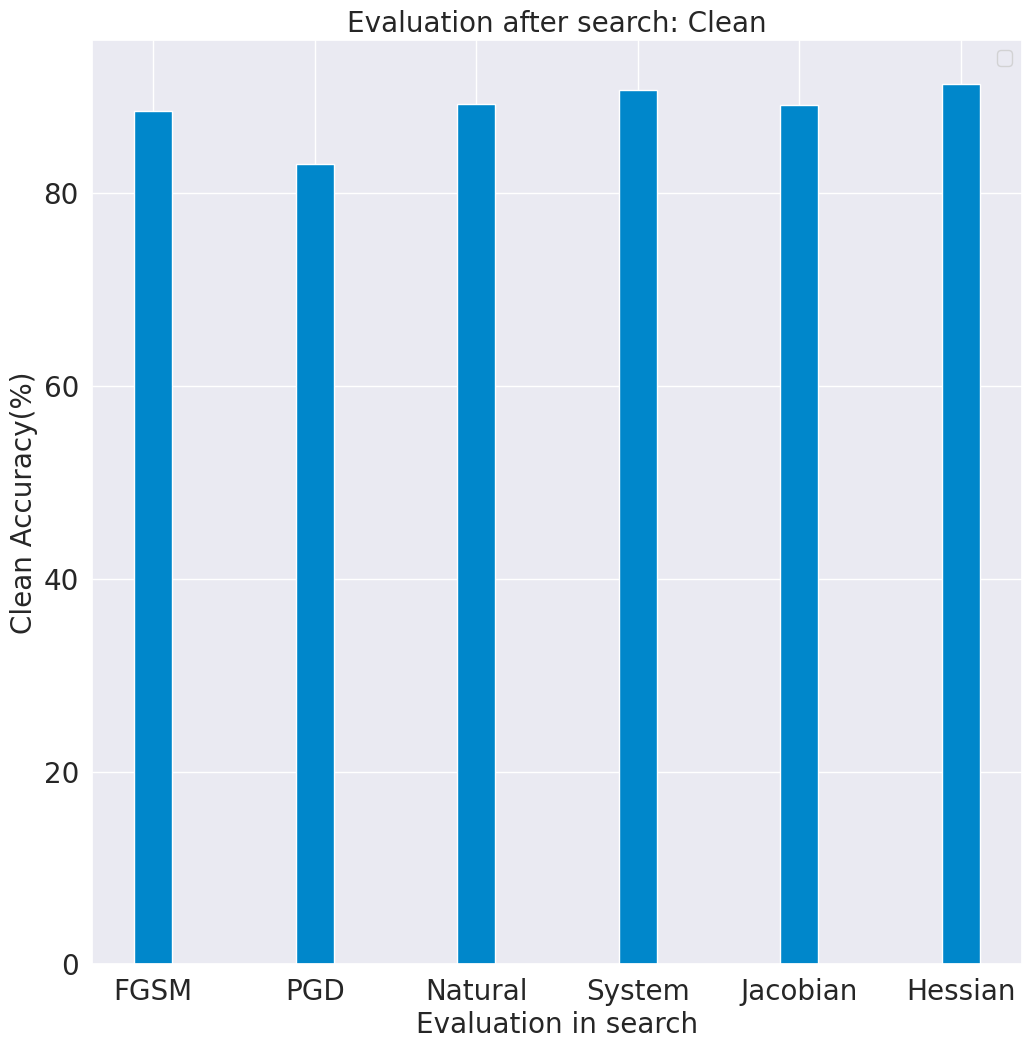}%
	}
	\hfil
	\subfloat{\includegraphics[width=2.3in]{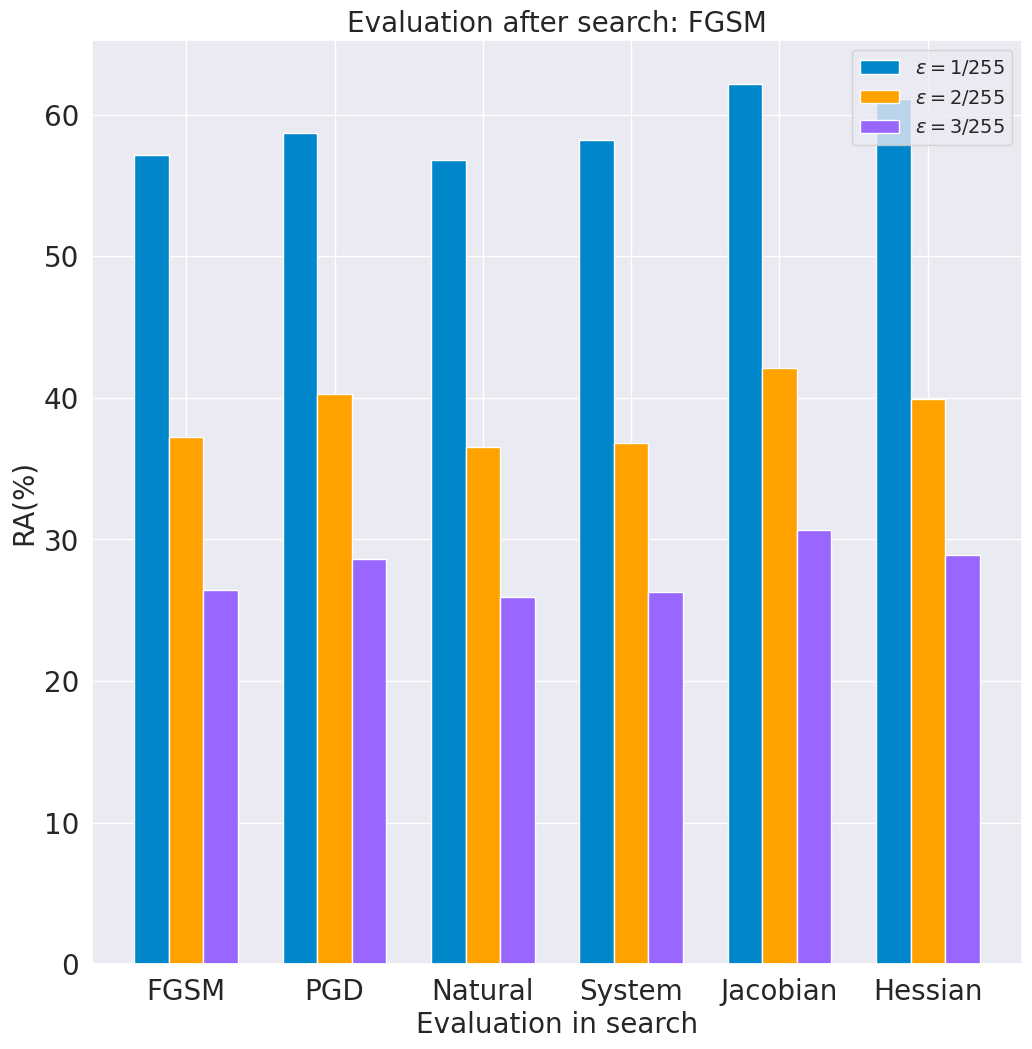}%
	}
	\subfloat{\includegraphics[width=2.3in]{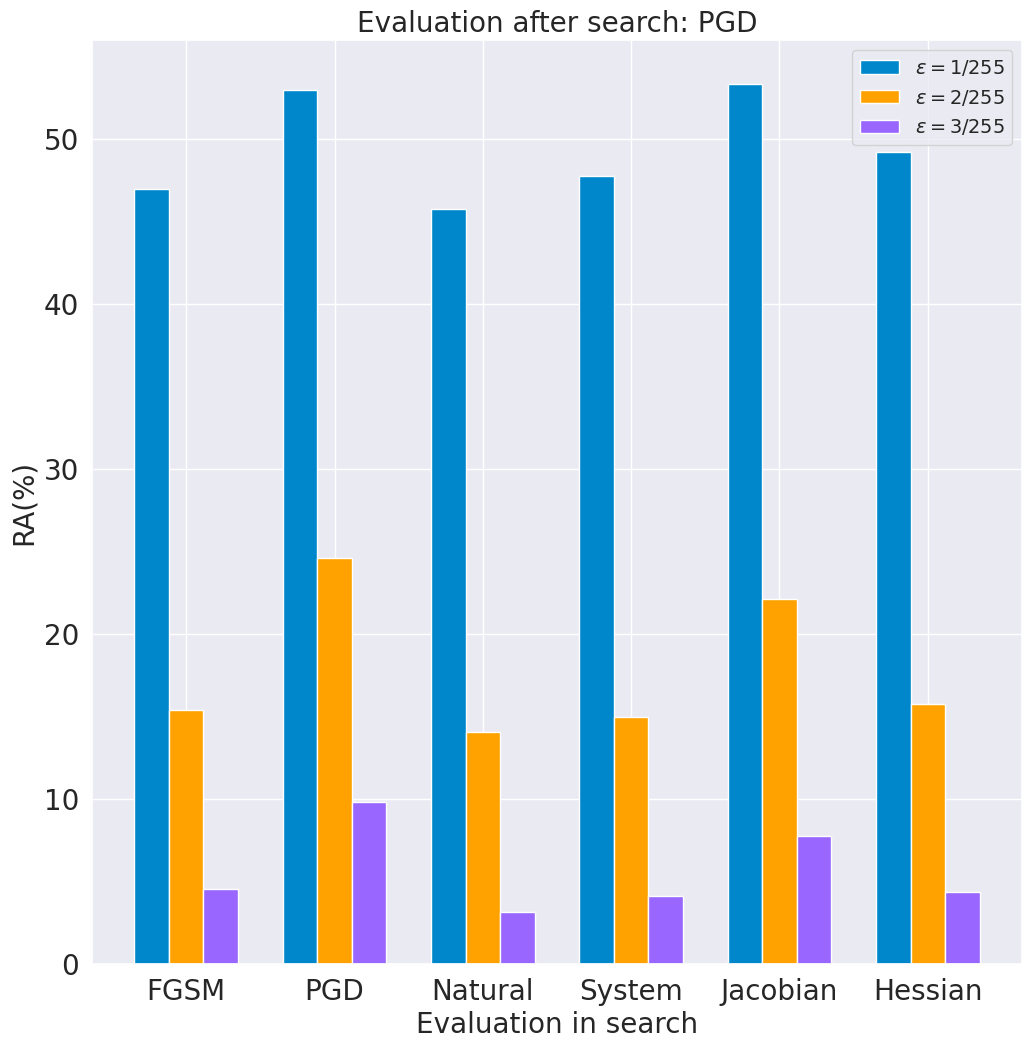}%
	}
	
	\subfloat{\includegraphics[width=2.3in]{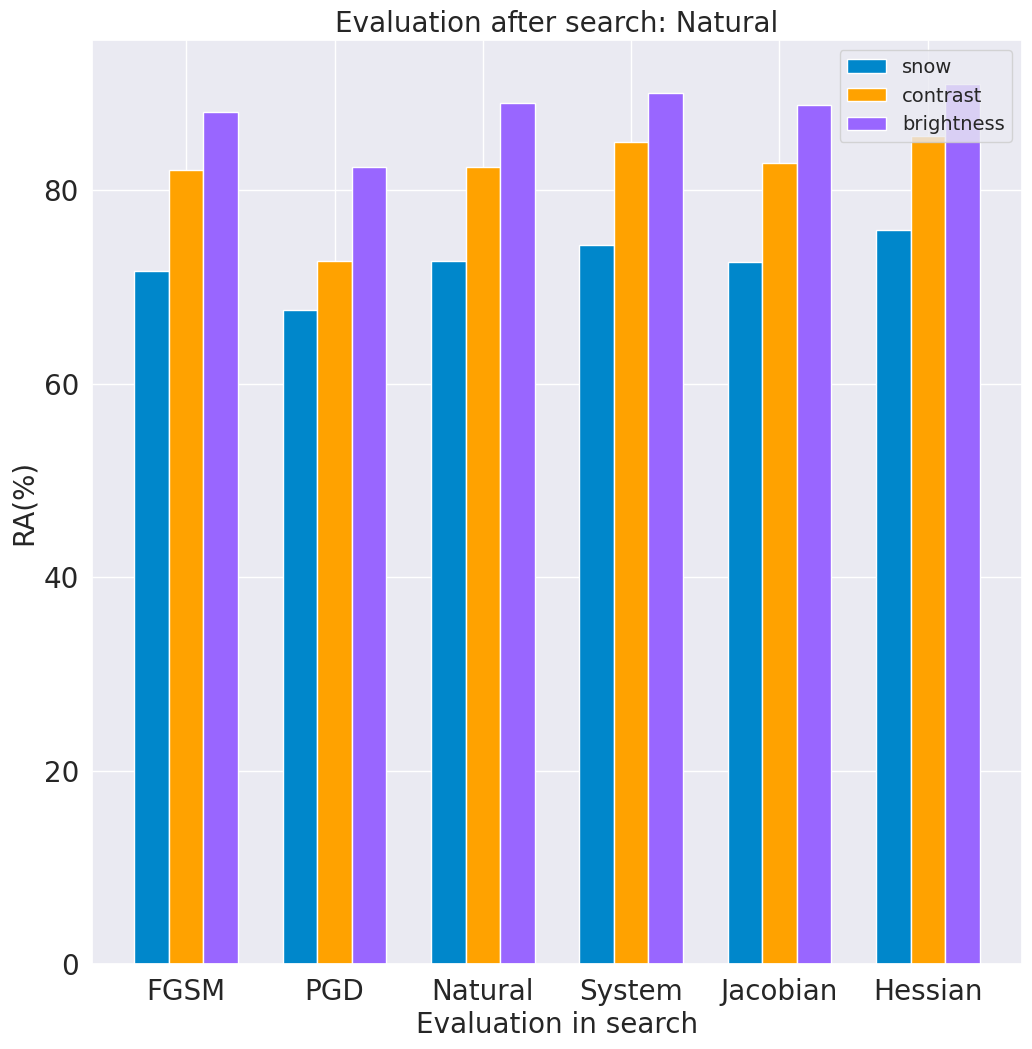}%
	}
	\hfil
	\subfloat{\includegraphics[width=2.3in]{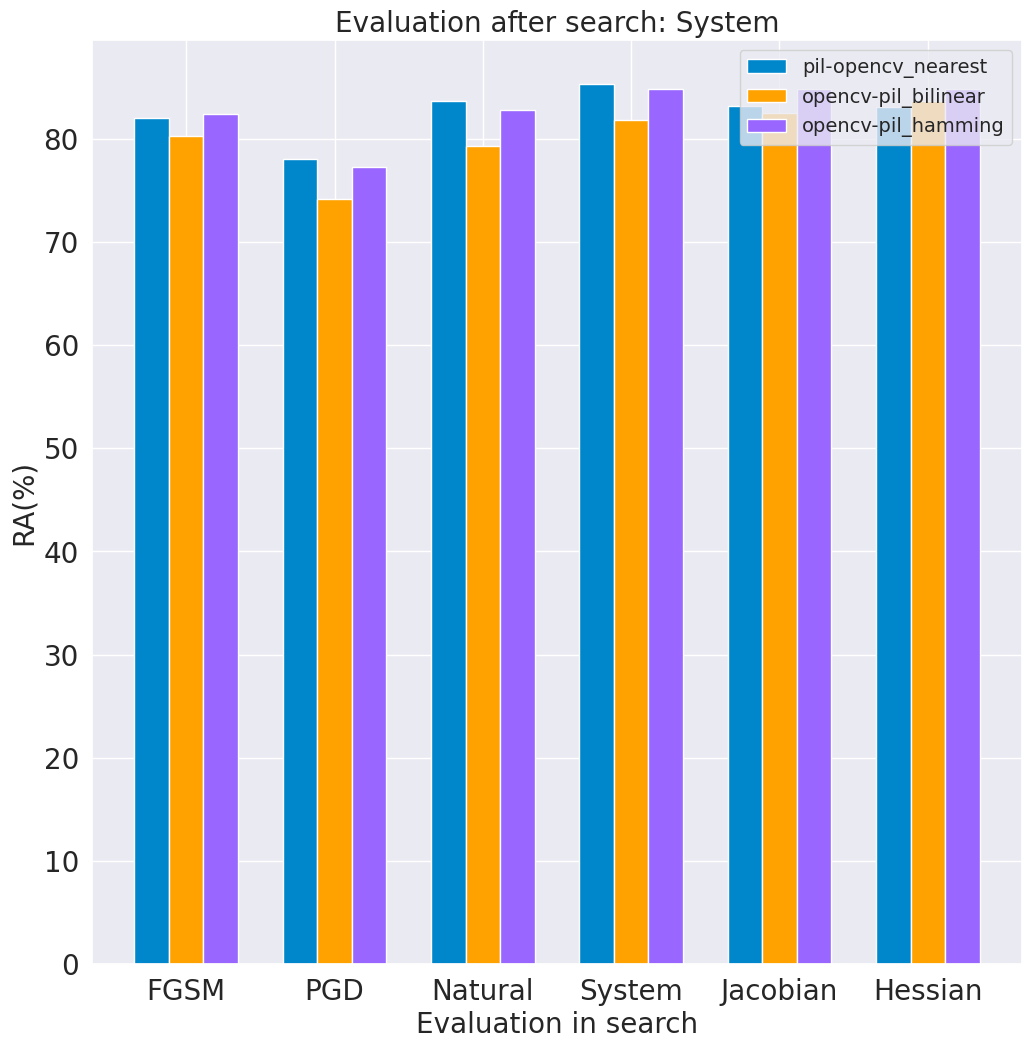}%
	}
	\subfloat{\includegraphics[width=2.3in]{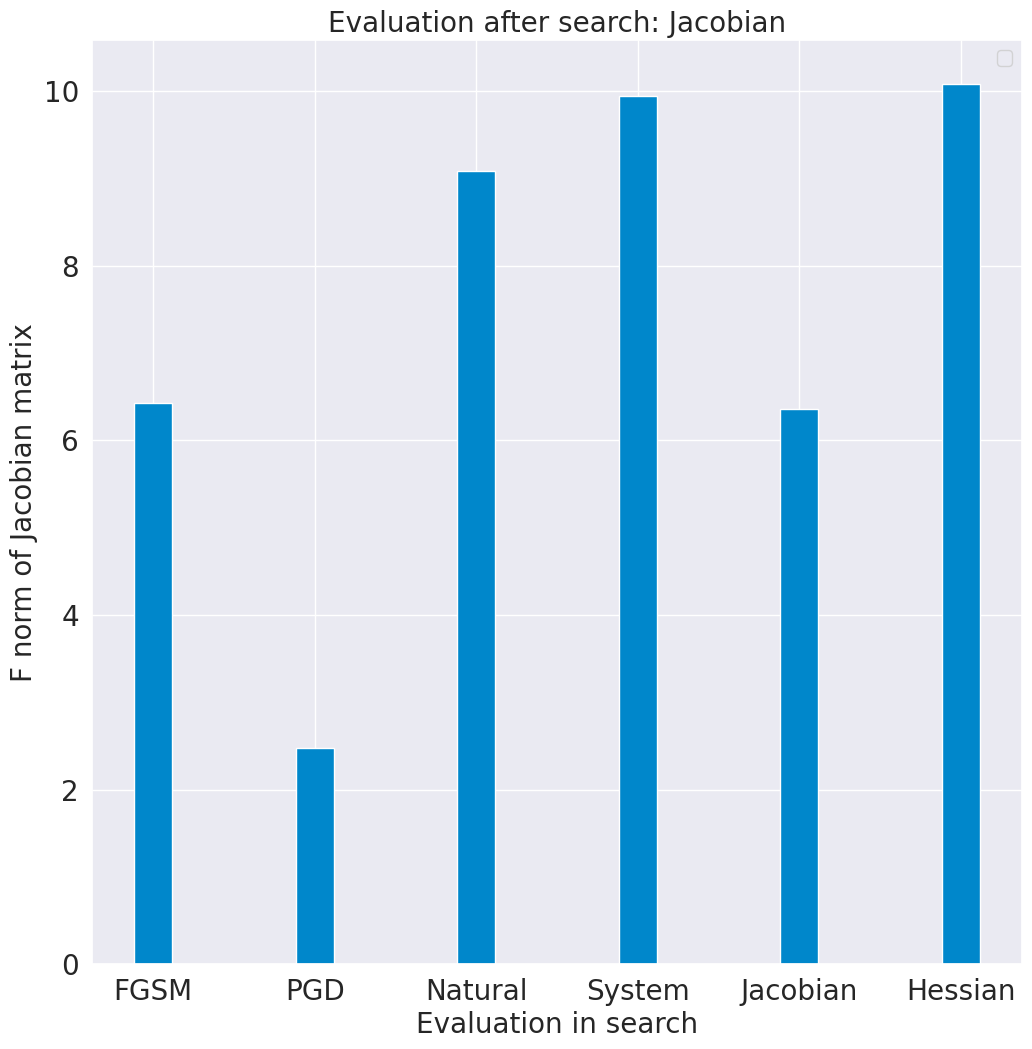}%
	}
	
	\subfloat{\includegraphics[width=2.3in]{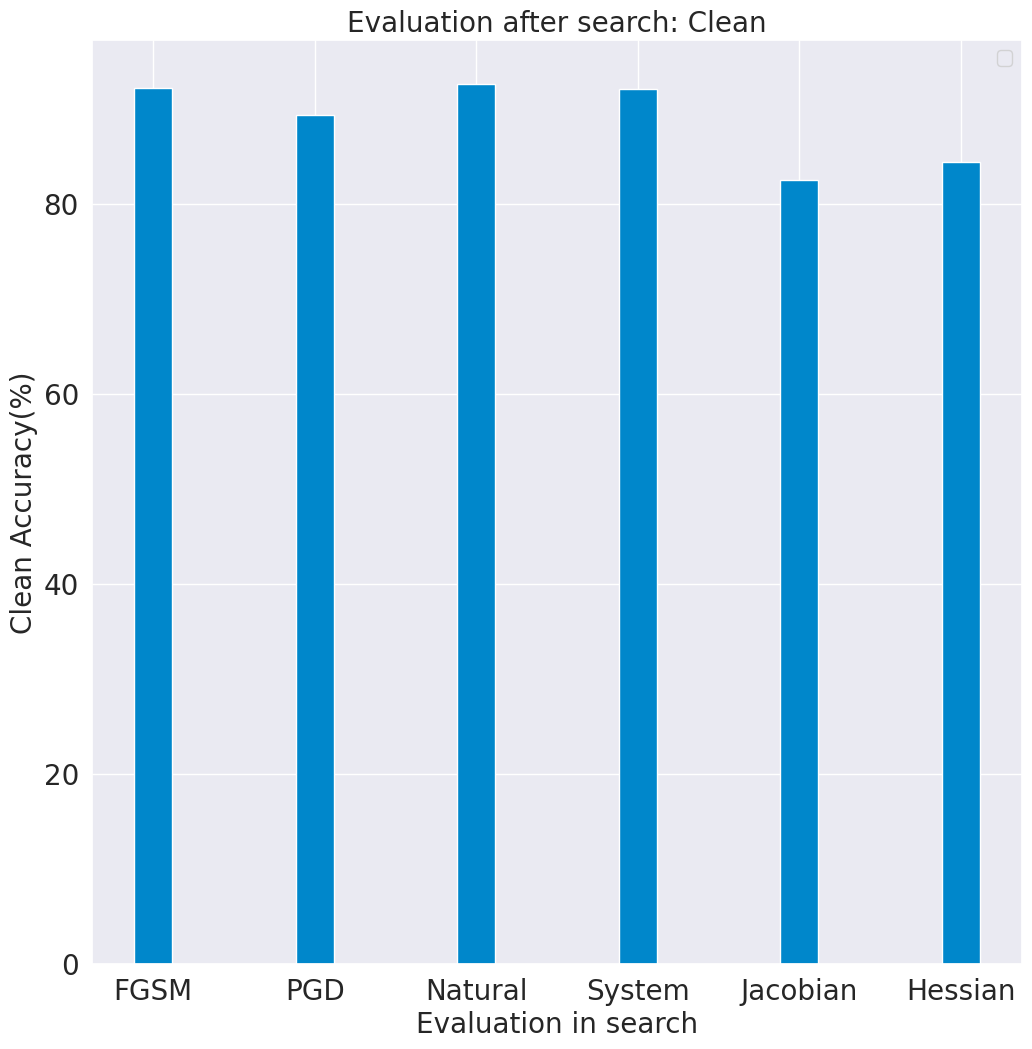}%
	}
	\hfil
	\subfloat{\includegraphics[width=2.3in]{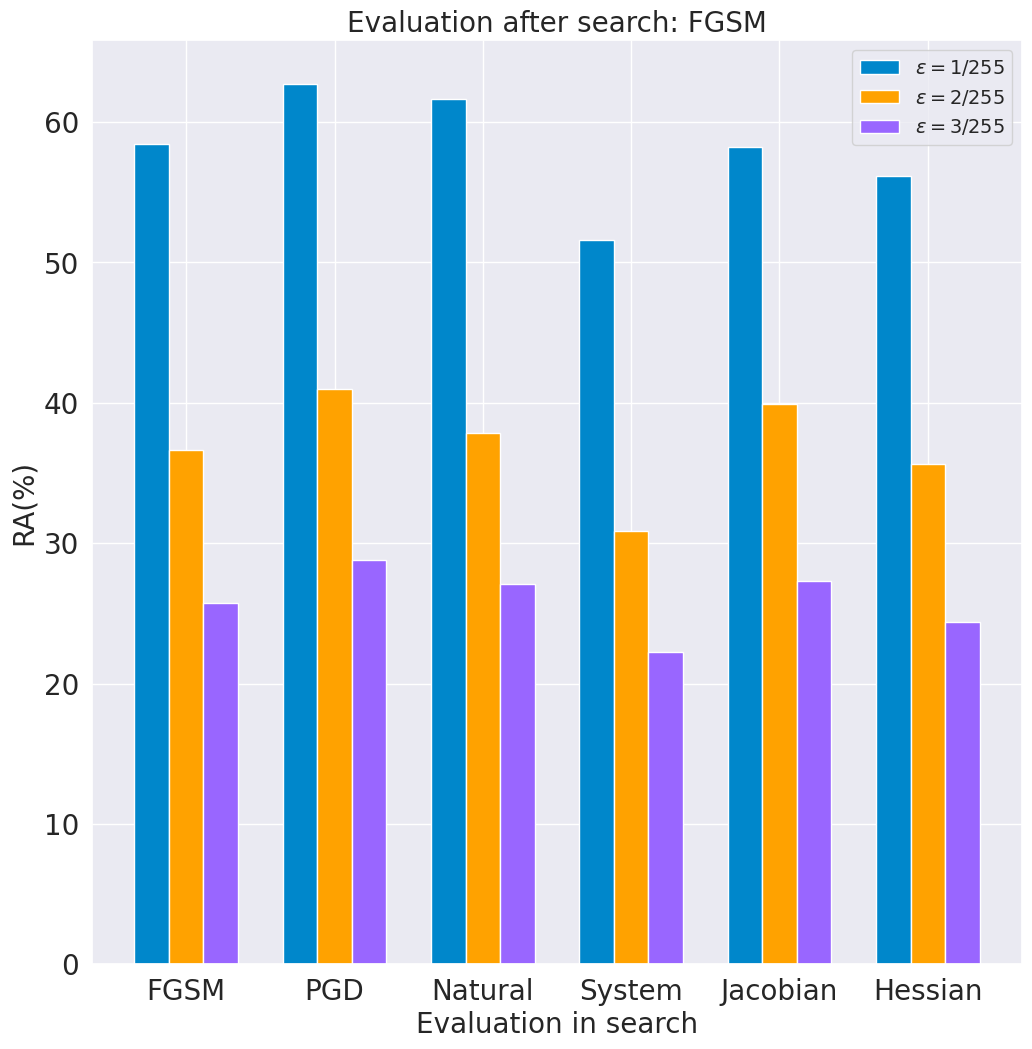}%
	}
	\subfloat{\includegraphics[width=2.3in]{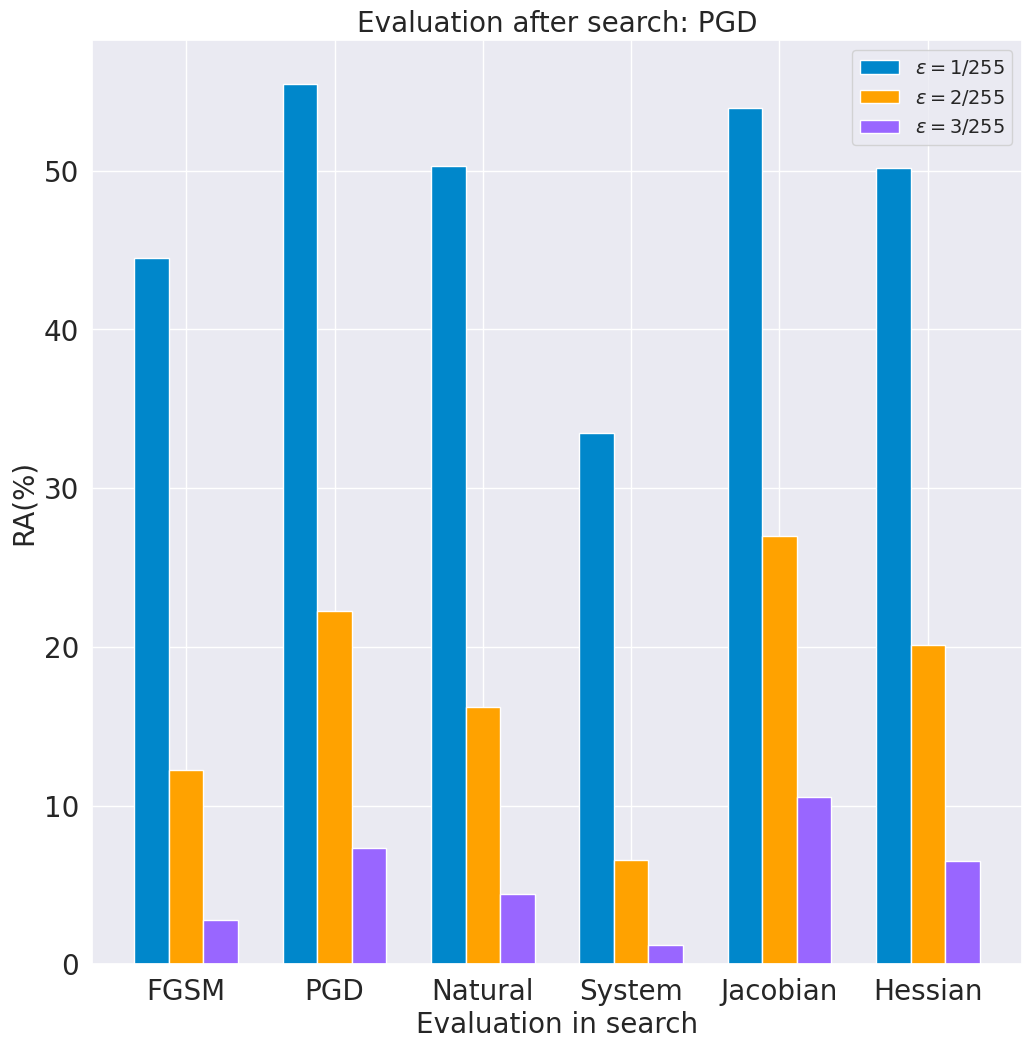}%
	}
	
	\subfloat{\includegraphics[width=2.3in]{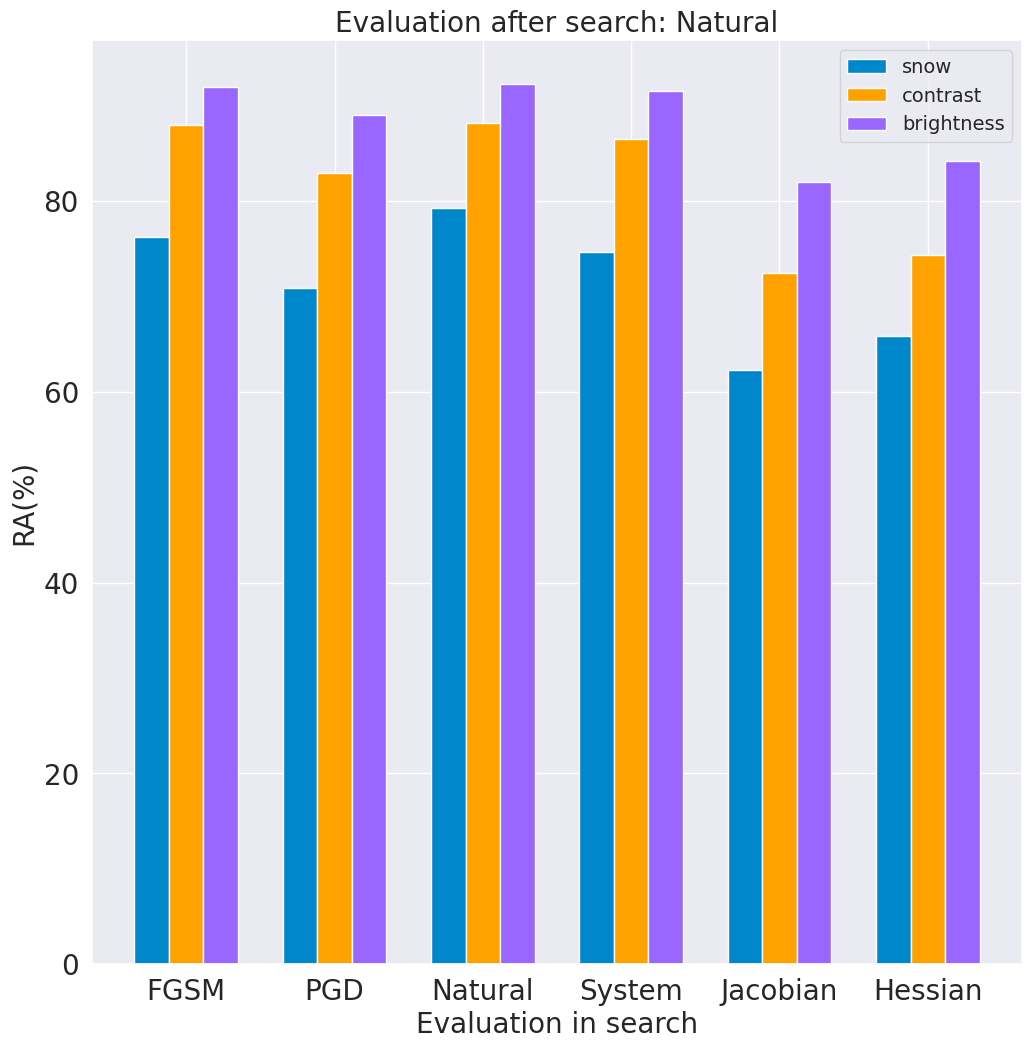}%
	}
	\hfil
	\subfloat{\includegraphics[width=2.3in]{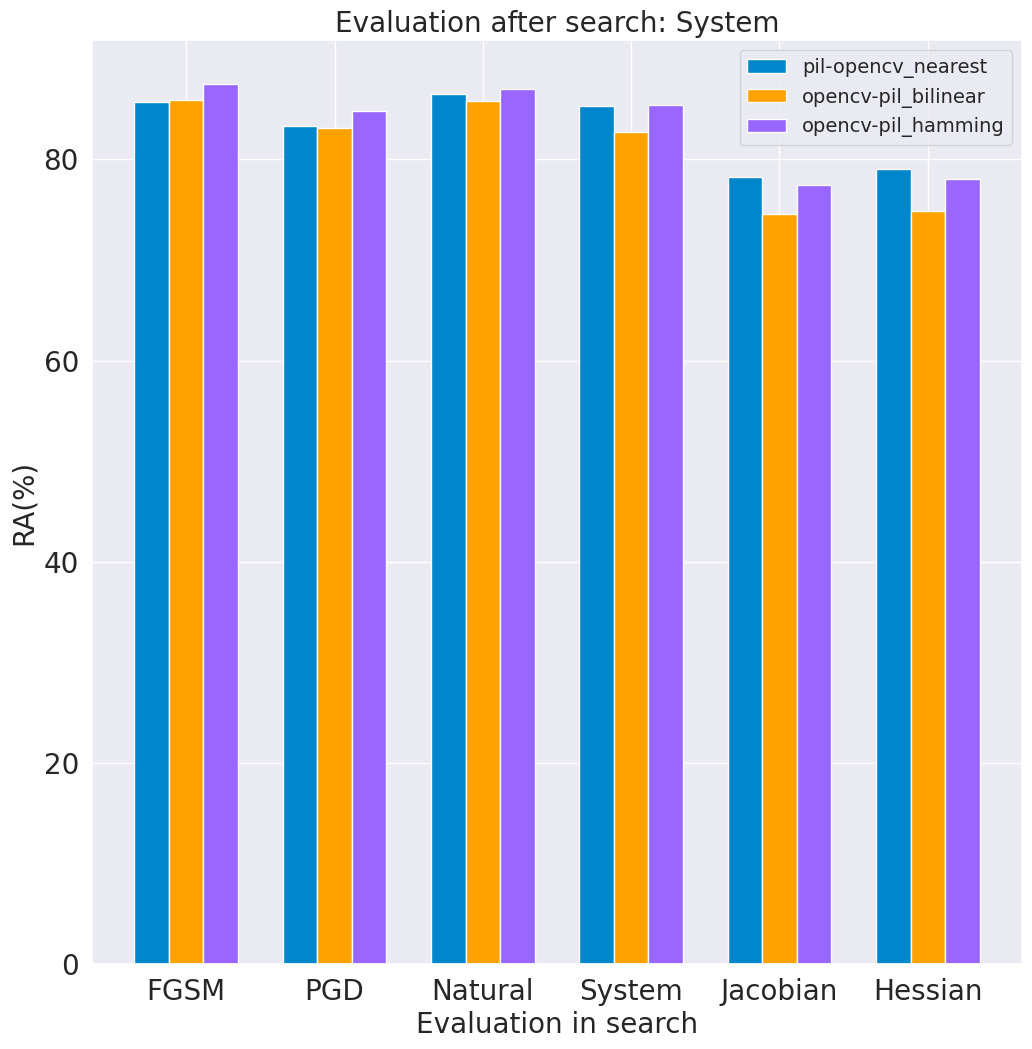}%
	}
	\subfloat{\includegraphics[width=2.3in]{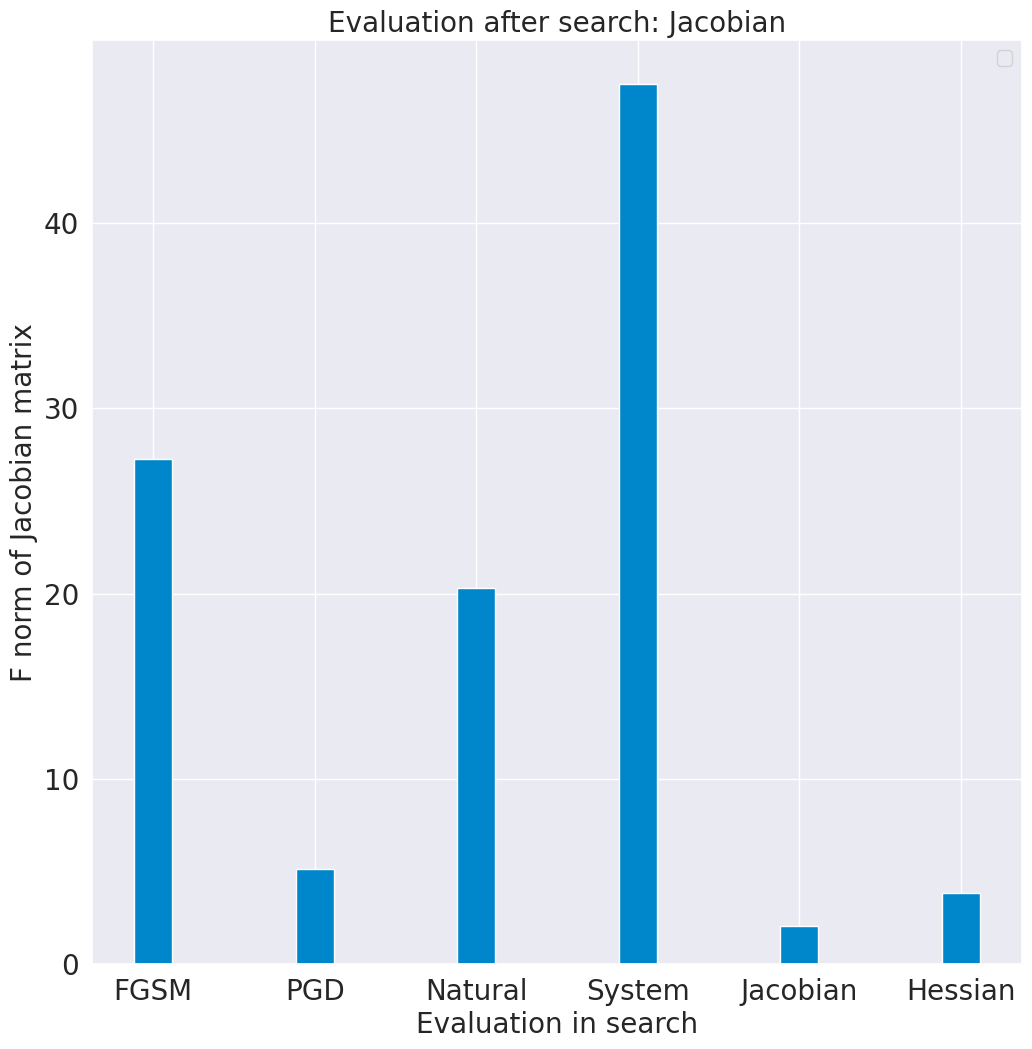}%
	}
	
	\caption{The performance of the searched architectures by Random Search (first two rows) and DE (last two rows)}
	\label{RandomSearch}
\end{figure*}

In this section, we report the performance of the searched architectures using different robustness evaluations, including clean accuracy, robust accuracy under FGSM, PGD, natural noises, system noises, and the $F$ norm of the Jacobian matrix. The performance of them is shown in Fig.~\ref{DARTS}, Fig.~\ref{NASP}, Fig.~\ref{SmoothDARTS}, and Fig.~\ref{RandomSearch}, respectively. 

From Fig.~\ref{DARTS}, we can see that generally, the Jacobian matrix can help find the most robust architectures in DARTS, and the robust accuracy under FGSM or PGD attack is higher than other architectures. From the results of the PGD attack, the searched architecture under natural noises achieves the lowest robustness. From the results of the Jacobian matrix, we can see that DARTS can find the architecture with the lowest Jacobian value, which verifies the effectiveness of NAS methods. However, we can also see that the lowest Jacobian value does not mean the highest robustness. Thus, if we hope to find robust architectures against the adversarial noises, the better method is adopting the adversarial noises as the evaluation metric directly. In addition, from the results of DARTS, we can see that the searched architectures under PGD attack are not more robust under the natural noises and system noises than other robustness metrics. Thus, multiple types of robustness evaluations can be needed to be considered if comprehensively robust architectures against multiple types of noises are hoped to be obtained. In the PC-DARTS method, both the Jacobian matrix and Hessian matrix can help find more robust architectures than other metrics.

\begin{table*}[h]
	\renewcommand\arraystretch{1.5}
	\scriptsize
	\centering
	\caption{Comparison of the parameter number (M) and robust accuracy ($\%$) of adversarial noises of different models on the CIFAR10 dataset.}
	
	\label{Comprehensive}
	\setlength{\textwidth}{24mm}{
		\begin{tabular}{c|c|c|cccccc}			
			\toprule			
			\multirow{1}{*}{ Method}&	\multirow{1}{*}{Type} &	\multirow{1}{*}{Param} &\multicolumn{1}{c}{Clean}&  \multicolumn{1}{c}{FGSM-LinfAttack}  & PGD-LinfAttack	 &   CW-LinfAttack		
			&  MI-LinfAttack      &AAA\\		
			\midrule		
			{ResNet-18}&	{Manual}   &   11.17&93.64&51.4&34.2&34.2&33.6&30.6 \\	
			{MobileNet-v2}&	{Manual}   &  2.30& 92.68&48.6&25.6&24.2&25.0&20.8   \\	
			{DenseNet-121}&	{Manual}   &  6.96&94.31&52.8&31.6&31.0&31.4  &26.8\\
			\midrule
			
			{Random$\_$Search$\_$FGSM}	&	{Automatic}  &  2.64&88.57&54.2&43.6&44.6&43.4   &41.6                 \\
			{Random$\_$Search$\_$AAA}	&	{Automatic}  &3.43&83.00& \textbf{55.8}&\textbf{50.6}&\textbf{50.8}&\textbf{50.6}  &\textbf{49.4} \\
			\bottomrule		
	\end{tabular}}
\end{table*}

\begin{table*}[h]
	\renewcommand\arraystretch{1.5}
	\scriptsize
	\centering
	\caption{The performance of different network architectures under the searched attack on different source models. The models in each row are source models to search attacks, and that in each column are models to evaluate the robustness.} 
	\setlength{\textwidth}{24mm}{
		\begin{tabular}{c|cccccc}
			\hline
			\multicolumn{1}{c}{Models} &Random$\_$Search$\_$FGSM  & PCDARTS$\_$$\_$FGSM & FairDARTS$\_$$\_$FGSM & NASP$\_$$\_$FGSM   & SmoothDARTS$\_$$\_$FGSM &  \\
			\hline
			\multicolumn{1}{c}{DARTS$\_$$\_$FGSM}      & 44.4 &43.0&45.0&42.6&54.0\\
			\multicolumn{1}{c}{SmoothDARTS$\_$$\_$Jacobian}      &\textbf{43.2}&\textbf{42.2} &\textbf{45.2}&\textbf{42.4}&\textbf{52.8}\\
			
			\multicolumn{1}{c}{FairDARTS$\_$$\_$Jacobian}      &\textbf{43.2}&\textbf{42.2} &\textbf{45.2}&\textbf{42.4}&\textbf{52.8}\\
			\hline
		\end{tabular}
	}
	\label{circuit1}
\end{table*}

\begin{table*}[h]
	\scriptsize
	\centering
	\caption{The visualization of the searched $l_{\infty}$ attack by AAA across various models on CIFAR10 dataset. }	
	\label{circuit}
	\setlength{\tabcolsep}{4mm}{
		\begin{tabular}{cc}		
			\toprule
			{Source model}&	{The searched attack} \\
			\midrule		
			{DARTS$\_$$\_$FGSM}  &\makecell[c]{'A': MomentumIterative-LinfAttack, 'M': 0.00294,'I': 10; 'A': MT-LinfAttack,'L':L1, \\'M': 0.00392, 'I': 1;'A': MI-LinfAttack, 'L':Hinge,'M': 0.00294, 'I': 8}  \\	
			\midrule
			
			{SmoothDARTS$\_$$\_$Jacobian}	 & \makecell[c]{'A': MI-LinfAttack, 'L':Hinge$\_$P, 'M': 0.00392,'I': 6; 'A': MI-LinfAttack, \\'L':Hinge$\_$P, 'M': 0.00343,'I': 6;'A': MomentumIterative-LinfAttack, 'L':DLR,'M': 0.00392,'I': 7}  \\
			\midrule	     
			{FairDARTS$\_$$\_$Jacobian}	   & \makecell[c]{'A': MI-LinfAttack, 'L':Hinge$\_$P, 'M': 0.00392,'I': 6; 'A': MI-LinfAttack, \\'L':Hinge$\_$P, 'M': 0.00343,'I': 6;'A': MomentumIterative-LinfAttack, 'L':DLR,'M': 0.00392,'I': 7}     \\
			\bottomrule		
	\end{tabular}}
\end{table*}

We also investigate the time cost of different NAS methods under different robustness evaluations. The comparison of the searching time is listed in Table \ref{NAStime}. In general, differentiable-based architecture search methods are more efficient than non-differentiable based such as DE and random search. Among the differentiable-based methods, some DARTS variants that reduce the time cost of DARTS, such as PC-DARTS and NASP, prove to be still more efficient than the original DARTS method. In the adversarial noises, taking the stronger adversarial attack as the evaluation will bring out more computational burden. For instance, the total time cost of differentiable-based architecture search methods under PGD is about twice that under FGSM. Due to the natural noises under one severe level having only 10,000 images, thus the searching time is less than system noises. Although we use the approximation of the second-order gradient in the process of obtaining the Hessian matrix, the time cost is still more than the Jacobian matrix, which only needs to calculate the one-order gradient. The above results coincide with our perceptions. Therefore, in the search process for robust architectures, apart from the performance of the searched architectures, the searching cost is also an important metric to evaluate the method.

\subsection{The Circuit of Auto Adversarial Attack and Defense}\label{sec44}
Existing works about AutoML for adversarial attack and defense are separate, which may cause the searched architectures to be not robust enough, or the searched attacks are not efficient enough. It is necessary to combine AutoML for adversarial attack and defense together to further improve the performance of architectures or attacks. Therefore, in this section, the mutual relationship between auto adversarial attack and defense is further investigated.

\subsubsection{The performance of the searched robust architectures under the searched adversarial attack}
\begin{figure}[htbp]
	\centering
	\includegraphics[scale=0.43]{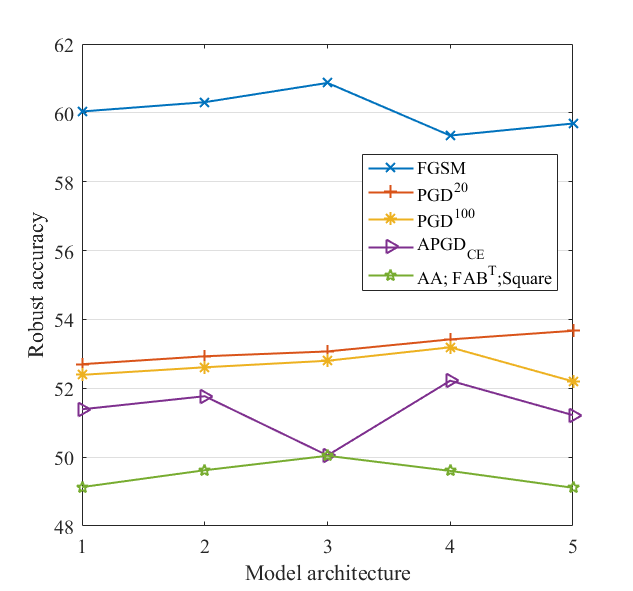}
	\caption{The robust accuracy of five architectures.}
	\label{dadd}
\end{figure}

In current works about NAS for robust architectures, the robust accuracy under FGSM or PGD is often taken as the robustness evaluation to guide the search of NAS, which may result in the searched model architectures are not robust enough. For example, the robust accuracy of five searched model architectures is presented in Fig~\ref{dadd}, which can be obtained from the report of AdvRush \cite{AdvRush2021}. If the evaluation just considers the robust accuracy of FGSM or PGD, it could cause the misjudgment of robust architectures in NAS. In these five architectures, we can find that the order of the robustness of architectures using stronger attacks to evaluate tends to be more accurate. Similar phenomena can also be seen in other related works \cite{LiuMultiobjective,Differentiable2021}. 

In this section, we also conduct the related experiments and report the performance of the searched robust architectures under the searched adversarial attack, FGSM, and PGD. The searched adversarial attack can possess better attack performance than FGSM and PGD. Random search is adopted as the search strategy. The experimental result is listed in Table \ref{Comprehensive}. From Table \ref{Comprehensive}, we can see that by taking the searched stronger attack as the robustness evaluation, NAS can find more robust architectures. The robust accuracy of Random$\_$Search$\_$FGSM under the searched attack, denoted as AAA, is only 41.6$\%$, while that of Random$\_$Search$\_$AAA can reach 49.4$\%$, which illustrates the feasibility and effectiveness of using more stronger attacks as the robustness evaluation in NAS methods.

\subsubsection{The performance of the searched attacks under under the searched robust architecture}

Apart from the effect of the searched stronger attack on the performance of NAS methods for robust architectures, we also investigate the performance of the searched attacks by using the searched more robust architecture as the source model. We select three searched models, including the DARTS$\_$$\_$FGSM, SmoothDARTS$\_$$\_$Jacobian, and FairDARTS$\_$$\_$Jacobian, as the source model to conduct the search for more efficient attacks, respectively. The robustness of three models is listed in Table \ref{NAS}, which shows that the order is DARTS$\_$$\_$FGSM, SmoothDARTS$\_$$\_$Jacobian, and FairDARTS$\_$$\_$Jacobian from the highest to lowest robustness. Random search is adopted as the search strategy. The searched attacks are evaluated on other five models, including Random$\_$Search$\_$FGSM, PCDARTS$\_$$\_$FGSM, FairDARTS$\_$$\_$FGSM, NASP$\_$$\_$FGSM, and SmoothDARTS$\_$$\_$FGSM. The performance of different attacks are presented in Table \ref{circuit1}. The visualization of the searched attacks is presented in Table \ref{circuit}.

From Table \ref{circuit1}, in general, we can see that by selecting the more robust architecture as the source model, more efficient adversarial attacks can tend to be found. For instance, the robust accuracy of SmoothDARTS$\_$$\_$FGSM evaluated by the searched attack using DARTS$\_$$\_$FGSM as the source model is 54.0$\%$. However, if we use the more robust architecture FairDARTS$\_$$\_$Jacobian as the source model, we can find the stronger attack, the robust accuracy of SmoothDARTS$\_$$\_$FGSM evaluated by which can reach 52.8$\%$. Due to the relatively easier implementation of random search, we can also see that using the SmoothDARTS$\_$$\_$Jacobian and  FairDARTS$\_$$\_$Jacobian, the searched attacks are the same, but these results can still verify the necessity of using more robust architectures as the source model to conduct the search process for efficient attacks.
\section{Conclusion}\label{sec5}
In this paper, we propose a platform called auto adversarial attack and defense. In this platform, we can alleviate the disadvantages of existing adversarial robustness open-sourced platforms that they can not optimize the architectures of DNN models or the performance of existing adversarial attacks. On the one hand, we provide the auto adversarial defense, where four types of robustness evaluation, including adversarial noises, system noises, natural noises, and quantified metrics, are incorporated in various NAS methods to find more robust architectures. On the other hand, we provide auto adversarial attack, which ensembles multiple evolutionary algorithms to find the near-optimal configuration of different adversarial attacks. In addition, we make the first step towards combining auto adversarial attack and defense together, forming a unified framework. Among auto adversarial defense, the searched efficient attack can be used as the new robustness evaluation to further enhance the robustness. Among auto adversarial attack, the searched robust architectures can be used as the threat model to help find stronger adversarial attacks. On the one hand, our proposed platform can provide a fair comparison environment for the research about AutoML for adversarial attack and defense. On the other hand, it can provide a tool for researchers to improve the robustness of neural architectures and the effectiveness of adversarial attacks.

\ifCLASSOPTIONcaptionsoff
  \newpage
\fi

{
\bibliography{reference}
\bibliographystyle{unsrt}
}


\newpage
\newpage
\clearpage

\appendices
\section{Limitations and broader impacts}

\text{$A^{3}D$} has some limitations, and we list them as follows. (1) Although we include some evolutionary algorithms or NAS methods to search for the near-optimal efficient attack or robust architectures, due to the rapid emergence of evolutionary algorithms or NAS methods, there might be some relevant works that are not implemented. (2) We focus on the performance of different search strategies and evaluations on the final searched attacks or architectures. In future, we also further study the performance of different search spaces.

\section{The performance of the searched adversarial attack on different datasets}
We report the performance of the searched adversarial attack on different datasets, including CIFAR10, CIFAR100 and ImageNet datasets. The selected norm of adversarial attacks includes $l_{2}$ and $l_{\infty}$. The other three optimiztaion algorithms, including local search, particle swarm optimization and random search are presented in Algorithm \ref{alglocalsearch}, Algorithm \ref{alglocalpso}, and Algorithm \ref{algrandomsearch}. Among them, local search and PSO are single-objective optimization algorithms, and random search is the multi-objective optimization algorithm.

\begin{algorithm}[!ht]	
	\caption{Local Search for Efficient Attacks}
	\label{alglocalsearch}
	\LinesNumbered
	\KwIn{ 
		The search space, maximum epoch of searching $T$.
	}
	\KwOut{The searched near-optimal attack  $\textbf{x}$.}
	
	Encode the individual according to the defined search space; \\ 
	$\textbf{x}_{best}$ $\gets$ generate the random initial individual according to the encoding;\\
	Searching epoch $t=1$; \\ 
	\While{t<T}{
		
		Construct the neighborhood by randomly changing the loss function, magnitude, iteration number and attacker operation based on $\textbf{x}_{best}$;\\
	Evaluate the individuals generated by local search and select the best one to substitute $\textbf{x}_{best}$;\\
	}
$\textbf{x}$ = $\textbf{x}_{best}$;\\
	\textbf{return}  $\textbf{x}$.
	
\end{algorithm}

\begin{algorithm}[!ht]	
	\caption{Particle Swarm Optimization for Efficient Attacks}
	\label{alglocalpso}
	\LinesNumbered
	\KwIn{ 
		The population size ${n}$, search space, maximum epoch of searching $T$.
	}
	\KwOut{The searched near-optimal attack  $\textbf{x}$.}
	
	Encode the individual according to the defined search space; \\ 
	$\textbf{P}$ $\gets$ generate the initial population according to the encoding;\\
	Searching epoch $t=1$; \\ 
	\While{t<T}{
		
		$\textbf{x}_{p}$  $\leftarrow$ Obtain the best particle for each individual;\\
		$\textbf{x}_{g}$  $\leftarrow$ Obtain the beat particle in the whole population;\\
				Update the position and velocity of all particles;\\

		Evaluate the fitness velues of all particles in the population;\\
		$t=t+1$;\\
	}
	\textbf{return}  $\textbf{x}$.
	
\end{algorithm}

\begin{algorithm}[!ht]	
	\caption{Random Search for Efficient Attacks}
		\label{algrandomsearch}
	\LinesNumbered
	\KwIn{ 
		The population size ${n}$, search space.
	}
	\KwOut{The searched near-optimal attack  $\textbf{x}$.}
	
	Encode the individual according to the defined search space; \\ 
	$\textbf{P}$ $\gets$ generate the initial population randomly according to the encoding;\\
		
				Evaluate the fitness velues of all particles in the population;\\
		
		Obtain the individual  $\textbf{x}$ with the lowest robustness $\leftarrow$ $argmin$($fitness$(\textbf{P}))\\
	
	\textbf{return}  $\textbf{x}$.
	
\end{algorithm}

\begin{table*}[h]
	\renewcommand\arraystretch{1.5}
	\scriptsize
	\centering
	\caption{Comparison of the robust accuracy ($\%$) and time cost (min) under the $l_{\infty}$ attack of APGD, the searched attack by local search, PSO, DE, NSGA-II across various defense strategies with the magnitude set as 8/255 using 100 images on CIFAR100 dataset.}
	
	\label{CIFAR100Linf}
	\setlength{\tabcolsep}{3mm}{
		\begin{tabular}{c|c|ccccccccccc}			
			\toprule			
			\multirow{2}{*}{Defense Method}&	\multirow{2}{*}{Model} &\multicolumn{2}{c}{Clean}&  \multicolumn{2}{c}{FGSM-LinfAttack} &  \multicolumn{2}{c}{PGD-LinfAttack-7} &\multicolumn{2}{c}{CW-LinfAttack-50} &\multicolumn{2}{c}{MI-LinfAttack-50}\\
			
			\cmidrule(r){3-4}   \cmidrule(r){5-6}  \cmidrule(r){7-8}
			\cmidrule(r){9-10} 	\cmidrule(r){11-12}
			&	& acc & time	 &   acc		
			&  time      &  acc   &   time  &  acc   &   time  & acc   &   time  \\		
			\midrule		
			{Fixing Data \cite{rebuffi2021fixing}}&	{WRN-28-10}   &  59.2&0.85&37.6&2.56&35.8&9.41&32.6&51.34&35.0&51.40\\	
			{OAAT \cite{addepalli2021towards}}	&	\text {ResNet-18}     & 60.8 &0.204&35.6&0.70&33.6&2.49&27.2&14.01&32.8&13.86\\
				{OAAT \cite{addepalli2021towards}}	&	\text {WRN-34}   &64.2 &0.54&43.0&2.32&38.6&9.55&34.0&54.63&36.8&55.37\\
			
		LBGAT \cite{cui2021learnable} & 	WRN-34-10 &59.4&0.487&36.8&2.14&35.0&8.91&31.2&54.50&34.6&55.58\\
		{AWP \cite{wu2020adversarial}}  & 	WRN-34-10 & 59.2 &0.503&34.4&2.13&33.2&8.99&31.2&56.46&33.2&57.03\\
			{IAR \cite{bernhard2020luring}}	 & WRN-34-10   &61.6&0.486&32.2&2.16&28.0&9.10&27.4&58.11&26.4&58.11\\
		{Overfit \cite{rice2020overfitting}} & {PAResNet-18 }  & 53.2&0.22&27.4&0.63&23.0&2.47&22.6&14.34&22.4&14.10\\
			\midrule
			
			\multirow{2}{*}{Defense Method}&	\multirow{2}{*}{Model} &\multicolumn{2}{c}{Local search}&  \multicolumn{2}{c}{PSO} &  \multicolumn{2}{c}{DE} &\multicolumn{2}{c}{Random Search} &\multicolumn{2}{c}{NSGA-II}\\

			\cmidrule(r){3-4}  \cmidrule(r){5-6}  \cmidrule(r){7-8}
			\cmidrule(r){9-10} 	\cmidrule(r){11-12}
			&	& acc & time	 &   acc		
			&  time      &  acc   &   time  &  acc   &   time  & acc   &   time  \\	
			\midrule		
				{Fixing Data \cite{rebuffi2021fixing}}&	{WRN-28-10}   &  32.6&50.37&32.6&230.58&31.0&306.37&32.4&56.47&32.0&38.40\\	
			{OAAT \cite{addepalli2021towards}}	&	\text {ResNet-18}     & 27.2 &16.14&27.0&68.00&26.2&106.21&27.0&17.16&27.2&11.87\\
			{OAAT \cite{addepalli2021towards}}	&	\text {WRN-34}   &33.6&50.08&34.0&260.19&32.6&352.58&33.2&60.45&33.6&40.99\\
			
			LBGAT \cite{cui2021learnable} & 	WRN-34-10 &31.2&56.00&30.6&247.48&30.0&322.22&31.2&59.47&30.6&38.92\\
			{AWP \cite{wu2020adversarial}}  & 	WRN-34-10 & 31.2 &57.32&30.4&247.39&29.4&323.53&31.2&62.27&30.2&40.67\\
			{IAR \cite{bernhard2020luring}}	 & WRN-34-10   &27.4&56.25&26.4&247.60&24.4&299.83&27.6&63.45&26.0&41.37\\
			{Overfit \cite{rice2020overfitting}} & {PAResNet-18 }  & 22.2&15.64&22.4&66.72&20.4&96.87&22.4&16.99&21.0&12.52\\
			
			\bottomrule		
	\end{tabular}}
\end{table*}

\begin{table*}[h]
	\renewcommand\arraystretch{1.5}
	\scriptsize
	\centering
	\caption{Comparison of the robust accuracy ($\%$) and time cost (min) under the $l_{2}$ attack of APGD, the searched attack by local search, PSO, DE, NSGA-II across various defense strategies with the magnitude set as 0.5 using 500 images on CIFAR100 dataset.}
	
	\label{CIFAR100L2}
	\setlength{\tabcolsep}{3mm}{
		\begin{tabular}{c|c|ccccccccccc}			
			\toprule			
			\multirow{2}{*}{Defense Method}&	\multirow{2}{*}{Model} &\multicolumn{2}{c}{Clean}&  \multicolumn{2}{c}{Momentum-L2Attack-50} &  \multicolumn{2}{c}{PGD-L2Attack-7} &\multicolumn{2}{c}{CW-L2Attack-50} &\multicolumn{2}{c}{MI-Attack-50}\\
			
			\cmidrule(r){3-4}   \cmidrule(r){5-6}  \cmidrule(r){7-8}
			\cmidrule(r){9-10} 	\cmidrule(r){11-12}
			&	& acc & time	 &   acc		
			&  time      &  acc   &   time  &  acc   &   time  & acc   &   time  \\		
			\midrule		
			{Fixing Data \cite{rebuffi2021fixing}}&	{WRN-28-10}   & 59.2&0.85 &50.8&48.46&41.2&9.75&37.8&52.52&40.4&52.97\\	
			{OAAT \cite{addepalli2021towards}}	&	\text {ResNet-18}     & 60.8&0.204&50.0&11.09&40.0&2.79&37.2&15.34&39.6&15.03\\
			{OAAT \cite{addepalli2021towards}}	&	\text {WRN-34}   &64.2&0.54&55.0&50.76&43.8&9.99&38.8&57.39&42.0&60.32\\
			
			LBGAT \cite{cui2021learnable} & 	WRN-34-10 &59.4&0.487&49.2&52.18&38.8&9.39&35.4&57.08&37.8&58.13\\
			{AWP \cite{wu2020adversarial}}  & 	WRN-34-10 & 59.2 &0.503&53.2&53.81&39.2&9.59&37.0&58.26&38.8&58.90\\
			{IAR \cite{bernhard2020luring}}	 & WRN-34-10   &61.6&0.486&49.4&55.46&35.6&10.03&33.6&59.36&33.6&59.56\\
			{Overfit \cite{rice2020overfitting}} & {PAResNet-18 }  &53.2&0.22 &44.4&11.85&30.2&2.76&29.0&16.54&28.6&15.26\\
			\midrule
			
			\multirow{2}{*}{Defense Method}&	\multirow{2}{*}{Model} &\multicolumn{2}{c}{Local search}&  \multicolumn{2}{c}{PSO} &  \multicolumn{2}{c}{DE} &\multicolumn{2}{c}{Random Search} &\multicolumn{2}{c}{NSGA-II}\\

			\cmidrule(r){3-4}  \cmidrule(r){5-6}  \cmidrule(r){7-8}
			\cmidrule(r){9-10} 	\cmidrule(r){11-12}
			&	& acc & time	 &   acc		
			&  time      &  acc   &   time  &  acc   &   time  & acc   &   time  \\	
			\midrule		
			{Fixing Data \cite{rebuffi2021fixing}}&	{WRN-28-10}    &37.0&373.29&37.0&989.39&37.4&319.57&37.0&373.29&37.2&1003.94 \\	
			{OAAT \cite{addepalli2021towards}}	&	\text {ResNet-18}     & 36.2&124.64&36.2&348.39&36.4&100.32&36.2&124.64&36.2&349.99 \\
			{OAAT \cite{addepalli2021towards}}	&	\text {WRN-34}   &37.4&438.61&37.2&1092.79&37.6&356.67&37.4&438.61&37.4&1112.54\\
			
			LBGAT \cite{cui2021learnable} & 	WRN-34-10 &34.2&407.04&34.0&1011.72&34.2&334.54&34.2&407.04&33.8&1023.48\\
			{AWP \cite{wu2020adversarial}}  & 	WRN-34-10   &35.6&405.26&35.6&1011.07&36.2&338.82&35.6&405.26&35.6&1033.69\\
			{IAR \cite{bernhard2020luring}}	 & WRN-34-10   &32.4&410.50&31.8&942.98&31.8&337.32&32.4&410.50&32.0&1014.04\\
			{Overfit \cite{rice2020overfitting}} & {PAResNet-18 }  &27.8&116.80&27.4&290.00&27.8&96.51&27.8&116.80&27.8&315.31 \\
			
			\bottomrule		
	\end{tabular}}
\end{table*}

\begin{table*}[t]
	\renewcommand\arraystretch{1.5}
	\scriptsize
	\centering
	\caption{Comparison of the robust accuracy ($\%$) and time cost (min) under the $l_{\infty}$ attack of APGD, the searched attack by local search, PSO, DE, NSGA-II across various defense strategies with the magnitude set as 4/255 using 1280 images on ImageNet dataset.}
	
	\label{ImagenetLinf}
	\setlength{\tabcolsep}{3mm}{
		\begin{tabular}{c|c|ccccccccccc}			
			\toprule			
			\multirow{2}{*}{Defense Method}&	\multirow{2}{*}{Model} &\multicolumn{2}{c}{Clean}&  \multicolumn{2}{c}{FGSM-LinfAttack} &  \multicolumn{2}{c}{PGD-LinfAttack-7} &\multicolumn{2}{c}{CW-LinfAttack-50} &\multicolumn{2}{c}{MI-LinfAttack-50}\\
			
			\cmidrule(r){3-4}   \cmidrule(r){5-6}  \cmidrule(r){7-8}
			\cmidrule(r){9-10} 	\cmidrule(r){11-12}
			&	& acc & time	 &   acc		
			&  time      &  acc   &   time  &  acc   &   time  & acc   &   time  \\		
			\midrule		
			{FastAT \cite{wong2020fast}}&	{WRN-28-10}    & 48.75&31.57 &28.52&39.81&24.77&83.00&23.28&321.94&23.36&351.55\\	
			{Salman \cite{salman2020adversarially}}	&	{ResNet-18}       & 48.67&28.52&27.97&31.76 &25.94&52.49&23.44&113.98&25.23&160.88\\
			{Salman \cite{salman2020adversarially}}	&	{ResNet-50}  &60.62&30.55&39.92&41.68&36.25&94.07&34.77&400.51&35.39&396.61\\
			{Salman \cite{salman2020adversarially}}&{WideResNet-50}&66.25&33.34&43.75&55.53&38.75&152.58&38.36&740.14&37.81&735.32\\
			{Label Smoothing \cite{shafahi2019label}}&InceptionV3&63.44&14.92&53.20&39.53&2.27&71.66&0.00&280.14&0.234&276.16\\
			EnsembleAT \cite{tramer2017ensemble}& Inception$\_$resnetV2&69.77&18.93&41.0&52.01&6.48&128.19&0.63&608.87&0.703&621.08\\
			
			\midrule
			
			\multirow{2}{*}{Defense Method}&	\multirow{2}{*}{Model} &\multicolumn{2}{c}{Local search}&  \multicolumn{2}{c}{PSO} &  \multicolumn{2}{c}{DE} &\multicolumn{2}{c}{Random Search} &\multicolumn{2}{c}{NSGA-II}\\

			\cmidrule(r){3-4}  \cmidrule(r){5-6}  \cmidrule(r){7-8}
			\cmidrule(r){9-10} 	\cmidrule(r){11-12}
			&	& acc & time	 &   acc		
			&  time      &  acc   &   time  &  acc   &   time  & acc   &   time  \\	
					\midrule		
		{FastAT \cite{wong2020fast}}&	{WRN-28-10}    &22.89 &152.97&21.41&678.10&21.33&840.40&23.44&172.34&21.72&119.14 \\	
		{Salman \cite{salman2020adversarially}}	&	{ResNet-18}       & 22.89&70.40&21.64&281.56&21.64&361.85&23.52&78.49&22.27&56.41\\
		{Salman \cite{salman2020adversarially}}	&	{ResNet-50}  &34.45&196.89&31.88&822.60&31.80&1058.43&34.69&220.16&32.89&148.70\\
		{Salman \cite{salman2020adversarially}}&{WideResNet-50}&37.34&380.60&35.55&1599.29&35.47&2100.64&38.36&425.46&36.17&286.95\\
		{Label Smoothing \cite{shafahi2019label}}&InceptionV3&0.00&146.21&0.00&1013.90&0.00&367.67&0.00&255.62&0.00&169.45&\\
		EnsembleAT \cite{tramer2017ensemble}& Inception$\_$resnetV2&1.09&384.78&0.00&2214.45&0.00&909.86&0.70&587.96&0.00&383.57\\
			
			\bottomrule		
	\end{tabular}}
\end{table*}

\begin{table*}[t]
	\renewcommand\arraystretch{1.5}
	\scriptsize
	\centering
	\caption{Comparison of the robust accuracy ($\%$) and time cost (min) under the $l_{2}$ attack of APGD, the searched attack by local search, PSO, DE, NSGA-II across various defense strategies with the magnitude set as 3 using 1280 images on ImageNet dataset.}
	
	\label{ImagenetL2}
	\setlength{\tabcolsep}{3mm}{
		\begin{tabular}{c|c|ccccccccccc}			
			\toprule			
			\multirow{2}{*}{Defense Method}&	\multirow{2}{*}{Model} &\multicolumn{2}{c}{Clean}&   \multicolumn{2}{c}{Momentum-L2Attack-50} &  \multicolumn{2}{c}{PGD-L2Attack-7} &\multicolumn{2}{c}{CW-L2Attack-50} &\multicolumn{2}{c}{MI-Attack-50}\\
			
			\cmidrule(r){3-4}   \cmidrule(r){5-6}  \cmidrule(r){7-8}
			\cmidrule(r){9-10} 	\cmidrule(r){11-12}
			&	& acc & time	 &   acc		
			&  time      &  acc   &   time  &  acc   &   time  & acc   &   time  \\		
			\midrule		
			{FastAT \cite{wong2020fast}}&	{WRN-28-10}    &48.75& 31.57&35.31&319.22&7.19&89.78&33.20&275.78&3.59&389.37\\	
			{Salman \cite{salman2020adversarially}}	&	{ResNet-18}       &48.67&28.52& 35.94&128.11&5.63&57.15&36.48&149.35&3.52&197.68\\
			{Salman \cite{salman2020adversarially}}	&	{ResNet-50}  &60.62&30.55&48.44&363.60&8.36&98.42&45.23&391.84&4.53&232.99\\
			{Salman \cite{salman2020adversarially}}&{WideResNet-50}&66.25&33.34&52.81&700.11&13.28&157.97&50.47&731.67&6.88&411.13\\
			{Label Smoothing \cite{shafahi2019label}}&InceptionV3&63.44&14.92&0.47&262.62&3.05&57.61&2.50&286.93&0.47&290.17\\
	EnsembleAT \cite{tramer2017ensemble}& Inception$\_$resnetV2&69.77&18.93&1.02&619.06&5.16&115.21&1.64&633.85&0.78&659.74\\
			
			\midrule
			
			\multirow{2}{*}{Defense Method}&	\multirow{2}{*}{Model} &\multicolumn{2}{c}{Local search}&  \multicolumn{2}{c}{PSO} &  \multicolumn{2}{c}{DE} &\multicolumn{2}{c}{Random Search} &\multicolumn{2}{c}{NSGA-II}\\

			\cmidrule(r){3-4}  \cmidrule(r){5-6}  \cmidrule(r){7-8}
			\cmidrule(r){9-10} 	\cmidrule(r){11-12}
			&	& acc & time	 &   acc		
			&  time      &  acc   &   time  &  acc   &   time  & acc   &   time  \\	
			\midrule		
			{FastAT \cite{wong2020fast}}&	{WRN-28-10}    & 2.89&878.63 &2.58&3206.61&3.05&1436.86&2.73&1043.05&2.58&3195.9\\	
			{Salman \cite{salman2020adversarially}}	&	{ResNet-18}       &2.5&383.71 &2.27&1532.61&2.27&720.49&2.27&571.76&2.19&1571.31\\
			{Salman \cite{salman2020adversarially}}	&	{ResNet-50}  &3.75&1006.15&3.28&3572.37&3.75&1660.15&3.52&1362.70&3.20&2733.42\\
			{Salman \cite{salman2020adversarially}}&{WideResNet-50}&5.94&1878.39&5.55&6380.59&6.09&3009.5&5.86&2616.4&5.55&4403.56\\
			{Label Smoothing \cite{shafahi2019label}}&InceptionV3&0.00&1450.25&0.00&2720.38&0.00&3730.54&0.00&1848.48&0.00&2638.56&\\
			EnsembleAT \cite{tramer2017ensemble}& Inception$\_$resnetV2&0.00&3047.54&0.00&5814.95&0.00&5950.62&0.00&5323.06&0.00&5742.43&\\
			
			\bottomrule		
	\end{tabular}}
\end{table*}

\section{The comparison of the searched architectures}

We report the robustness performance of different searched architectures under different types of evaluations and present their visualization of them. In addition, we report the robustness performance in each type of robustness in detail.
\label{sec:limit}

Other five alforithms of NAS for robust architectures are presented in detail, including NASP, PC-DARTS, SmoothDARTS, FairDARTS and Random Search, as presented in Algorithm \ref{algnasp}, Algorithm \ref{algpcdarts}, Algorithm \ref{algfairdarts}, Algorithm \ref{algsmoothdarts} and Algorithm \ref{algrandom_search}, respectively.

\begin{algorithm}[!ht]
	\caption{Random Search for Robust Architectures}
	\label{algrandom_search}
	\LinesNumbered
	\KwIn{ 
		The initial weight ${w}$, initial architecture $\alpha$, search space,  maximum epoch of searching $T$, the number of warm$\_$epoch $T_{warm}$, robustness  evaluation metric.
	}
	\KwOut{The searched near optimal network architecture  $\alpha$.}
	
	Determine the $\mathcal{L}_{\text {train }}$ according to the robustness metric by Eq.~\ref{eq1}, Eq.~\ref{eq2}, Eq.~\ref{eq3}, Eq.~\ref{eq5d5} and Eq.~\ref{eq5} ;\\
	
Encode the individual according to the defined search space; \\ 
$\textbf{P}$ $\gets$ generate the initial population randomly according to the encoding;\\

Evaluate the fitness velues of all individuals in the population;\\

Obtain the individual  $\textbf{x}$ with the highest robustness $\leftarrow$ $argmax$($fitness$(\textbf{P}))\\

	\textbf{return} $\alpha$.
	
\end{algorithm}

\begin{algorithm}[htbp]	
	\caption{NASP for Robust Architectures}
		\label{algnasp}
	\LinesNumbered
	\KwIn{ 
		The initial weight ${w}$, initial architecture $\alpha$, search space,  maximum epoch of searching $T$, the number of warm$\_$epoch $T_{warm}$, robustness  evaluation metric.
	}
	\KwOut{The searched near optimal network architecture  $\alpha$.}
	
	Create the mixed operation according to the defined search space; \\ 
	Determine the $\mathcal{L}_{\text {train }}$ according to the robustness metric by Eq.~\ref{eee}, Eq.~\ref{mix} or Eq.~\ref{eq111};\\
	Training epoch $t=1$; \\ 
	\While{t<T}{
		\If{$t$< $T_{warm}$}{
			Update ${w}$ by calculating $\nabla_{w} \mathcal{L}_{\text {train }}(w, \alpha)$;\\
		}
		\Else{
			Get discrete representation of old architecture $\overline{{a}}_t^{(i, j)}=\operatorname{prox}_{\mathcal{C}_1}\left({a}_t^{(i, j)}\right)$;\\
			Update the architecture  $
			{a}_{t+1}=\operatorname{prox}_{\mathcal{C}_2}\left({a}_t-\nabla_{\overline{{a}}^t} \mathcal{L}_{\mathrm{val}}\left(w^t, \overline{{a}}_t\right)\right)
			$;\\
			Get discrete representation of new architecture $\overline{{a}}_{t+1}^{(i, j)}=\operatorname{prox}_{\mathcal{C}_1}\left({a}_{t+1}^{(i, j)}\right)$;\\
			Update ${w}$ by calculating $\nabla_{w} \mathcal{L}_{\text {train }}(w, \alpha)$;\\
		}
		$t=t+1$;\\
	}
	\textbf{return} $\alpha$.
	
\end{algorithm}

\begin{algorithm}[!ht]	
	\caption{PC-DARTS for Robust Architectures}
	\label{algpcdarts}
	\LinesNumbered
	\KwIn{ 
		The initial weight ${w}$, initial architecture $\alpha$, search space,  maximum epoch of searching $T$, the number of warm$\_$epoch $T_{warm}$, robustness  evaluation metric.
	}
	\KwOut{The searched near optimal network architecture  $\alpha$.}
	
	Create the mixed operation according to the defined search space by	$
	x^j=\sum_{o \in \mathcal{O}} S\left(\beta^o\right) o\left(M^o * I^j\right)+\left(1-M^o * I^j\right)
	$; \\ 
	Determine the $\mathcal{L}_{\text {train }}$ according to the robustness metric by Eq.~\ref{eee}, Eq.~\ref{mix} or Eq.~\ref{eq111};\\
	Training epoch $t=1$; \\ 
	\While{t<T}{
		\If{$t$< $T_{warm}$}{
			Update ${w}$ by calculating $\nabla_{w} \mathcal{L}_{\text {train }}(w, \alpha)$;\\
		}
		\Else{
			Update the architecture  $\alpha$ by calculating gradient $\nabla_{\alpha} \mathcal{L}_{v a l}\left(w-\xi \nabla_{w} \mathcal{L}_{\text {train }}(w, \alpha), \alpha\right)$;\\
			Update ${w}$ by calculating $\nabla_{w} \mathcal{L}_{\text {train }}(w, \alpha)$;\\
		}
		$t=t+1$;\\
	}
	\textbf{return} $\alpha$.
	
\end{algorithm}

\begin{algorithm}[h]	
	\caption{FairDARTS for Robust Architectures}
	\label{algfairdarts}
	\LinesNumbered
	\KwIn{ 
		The initial weight ${w}$, initial architecture $\alpha$, search space,  maximum epoch of searching $T$, the number of warm$\_$epoch $T_{warm}$, robustness  evaluation metric.
	}
	\KwOut{The searched near optimal network architecture  $\alpha$.}
	
	Create the mixed operation according to the defined search space; \\ 
	The zero-one loss $
	L_{0-1}=-\frac{1}{N} \sum_i^N\left(\sigma\left(\alpha_i\right)-0.5\right)^2
	$;\\
	Determine the $\mathcal{L}_{\text {train }}$ according to the robustness metric by Eq.~\ref{eee}, Eq.~\ref{mix} or Eq.~\ref{eq111};\\
	Training epoch $t=1$; \\ 
	\While{t<T}{
		\If{$t$< $T_{warm}$}{
			Update ${w}$ by calculating $\nabla_{w} \mathcal{L}_{\text {train }}(w, \alpha)$;\\
		}
		\Else{
			Update the architecture  $\alpha$ by calculating gradient $\nabla_{\alpha} \mathcal{L}_{v a l}\left(w-\xi \nabla_{w} \mathcal{L}_{\text {train }}(w, \alpha), \alpha\right)$;\\
			Update ${w}$ by calculating $\nabla_{w} \mathcal{L}_{\text {train }}(w, \alpha)$;\\
		}
		$t=t+1$;\\
	}
	\textbf{return} $\alpha$.
	
\end{algorithm}

\begin{algorithm}[!ht]
	\caption{SmoothDARTS for Robust Architectures}
		\label{algsmoothdarts}
	\LinesNumbered
	\KwIn{ 
		The initial weight ${w}$, initial architecture $\alpha$, search space,  maximum epoch of searching $T$, the number of warm$\_$epoch $T_{warm}$, robustness  evaluation metric.
	}
	\KwOut{The searched near optimal network architecture  $\alpha$.}
	
	Create the mixed operation according to the defined search space; \\ 
	Determine the $\mathcal{L}_{\text {train }}$ according to the robustness metric by Eq.~\ref{eee}, Eq.~\ref{mix} or Eq.~\ref{eq111};\\
	Training epoch $t=1$; \\ 
	\While{t<T}{
		\If{$t$< $T_{warm}$}{
			Update ${w}$ by calculating $\nabla_{w} \mathcal{L}_{\text {train }}(w, \alpha)$;\\
		}
		\Else{
			Update the architecture  $\alpha$ by calculating gradient $\nabla_{\alpha} \mathcal{L}_{v a l}\left(w-\xi \nabla_{w} \mathcal{L}_{\text {train }}(w, \alpha), \alpha\right)$;\\
			Compute $\delta$ according to $
			\delta \sim U_{[-\epsilon, \epsilon]}
			$ or $
			\delta^{n+1}=\mathcal{P}\left(\delta^n+l r * \nabla_{\delta^n} L_{\text {train }}\left(w, a+\delta^n\right)\right)
			$;\\
			Update ${w}$ by calculating $\nabla_{w} \mathcal{L}_{\text {train }}(w, \alpha + \delta)$;\\
		}
		$t=t+1$;\\
	}
	\textbf{return} $\alpha$.
	
\end{algorithm}

\begin{figure*}[h]
	\centering
	\subfloat[Normal Cell in DARTS$\_$FGSM]{\includegraphics[width=2.35in]{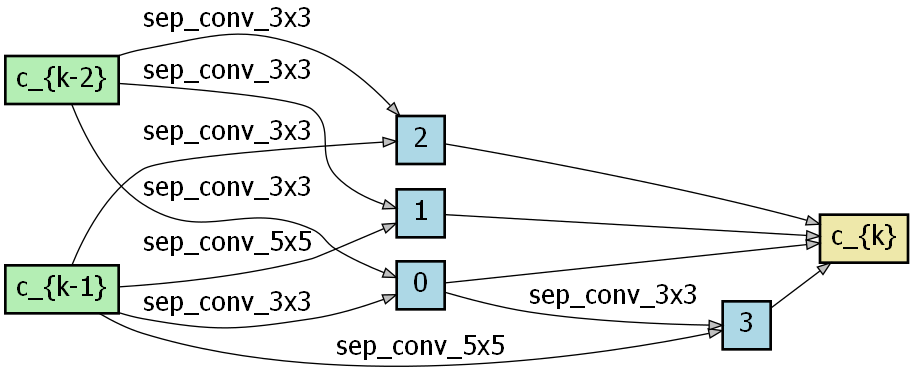}%
	}
	\hfil
	\subfloat[Normal Cell in DARTS$\_$PGD]{\includegraphics[width=2.35in]{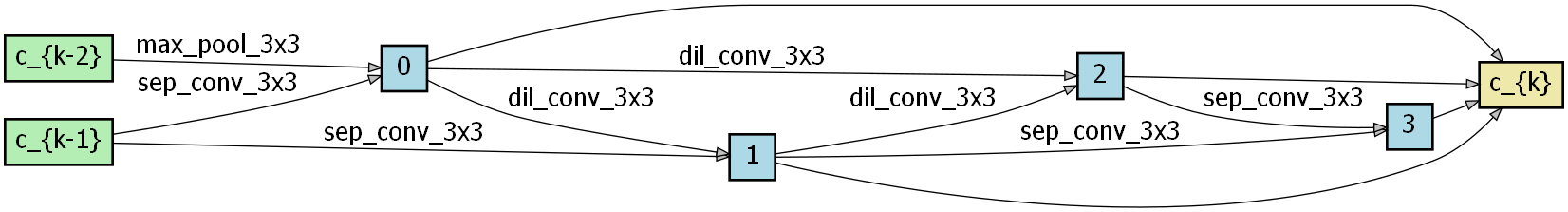}%
	}
	\subfloat[Normal Cell in DARTS$\_$Natural]{\includegraphics[width=2.35in]{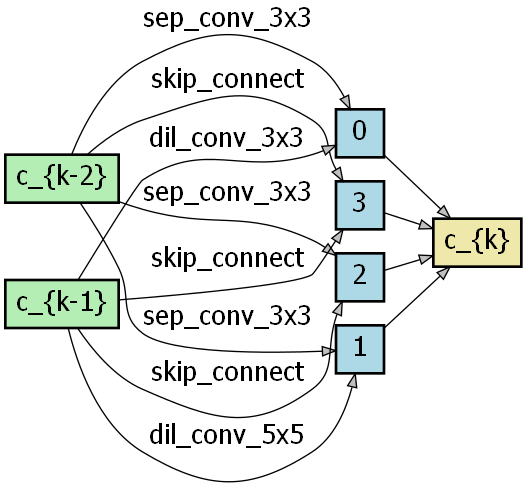}%
	}
	
	\subfloat[Reduction Cell in DARTS$\_$FGSM]{\includegraphics[width=2.35in]{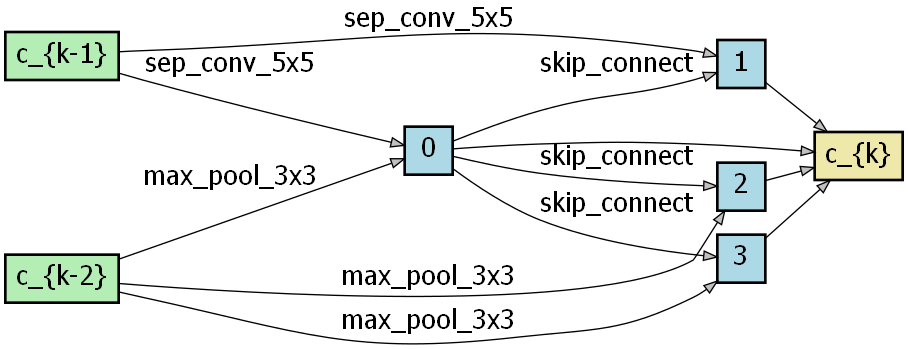}%
	}
	\hfil
	\subfloat[Reduction Cell in DARTS$\_$PGD]{\includegraphics[width=2.35in]{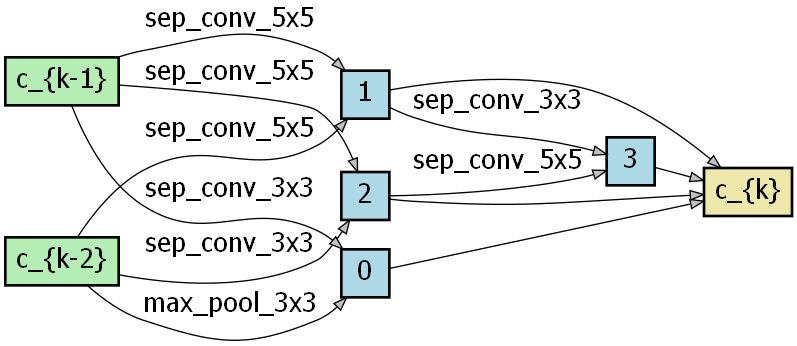}%
	}
	\subfloat[Reduction Cell in DARTS$\_$Natural]{\includegraphics[width=2.35in]{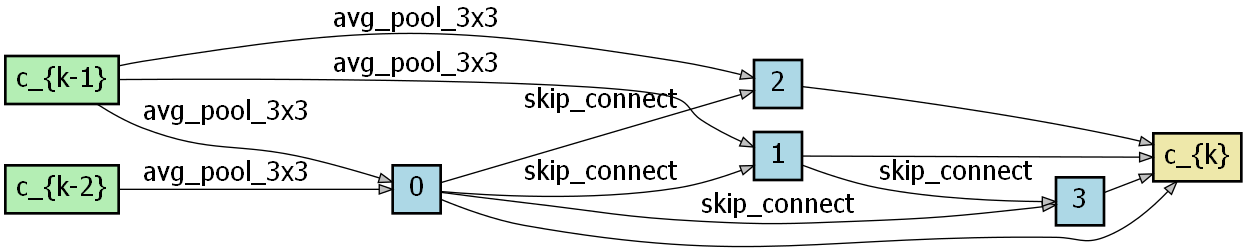}%
	}
	
	\subfloat[Normal Cell in DARTS$\_$System]{\includegraphics[width=2.35in]{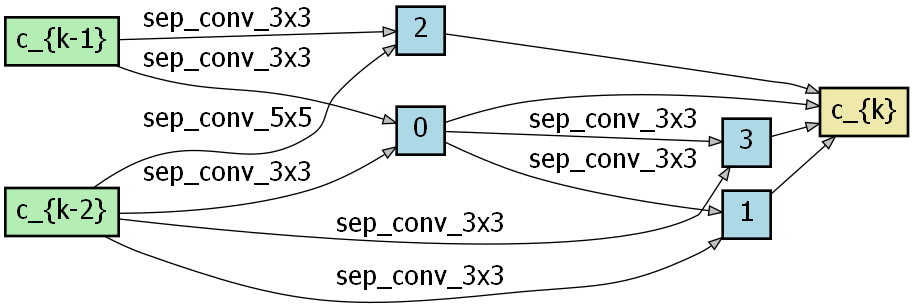}%
	}
	\hfil
	\subfloat[Normal Cell in DARTS$\_$Jacobian]{\includegraphics[width=2.35in]{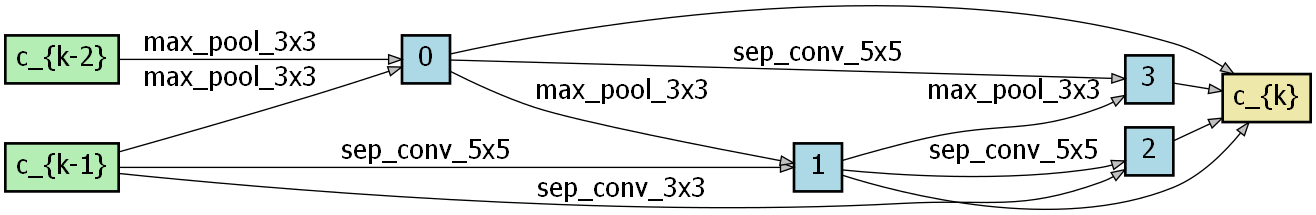}%
	}
	\subfloat[Normal Cell in DARTS$\_$Hessian]{\includegraphics[width=2.35in]{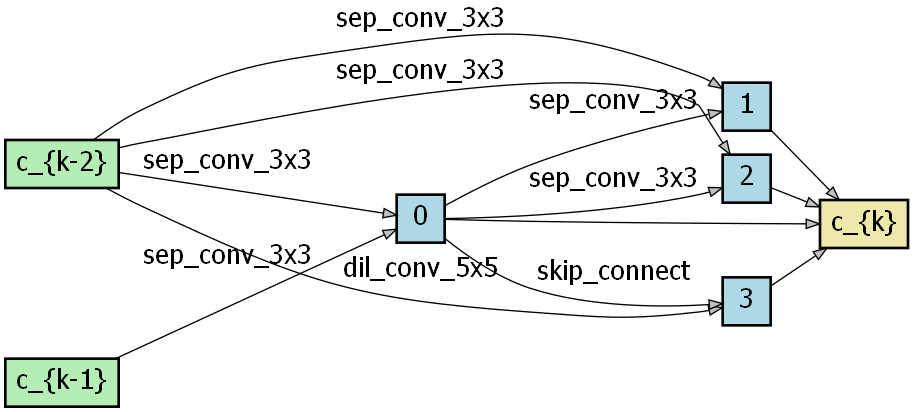}%
	}
	
	\subfloat[Reduction Cell in DARTS$\_$System]{\includegraphics[width=2.35in]{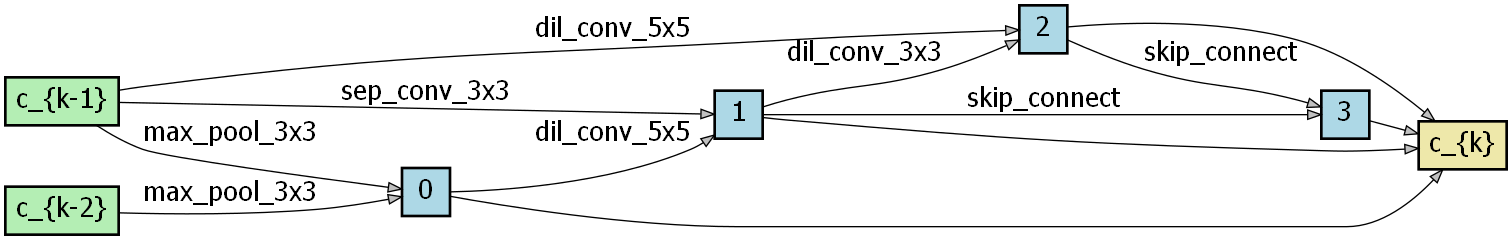}%
	}
	\hfil
	\subfloat[Reduction Cell in DARTS$\_$Jacobian]{\includegraphics[width=2.35in]{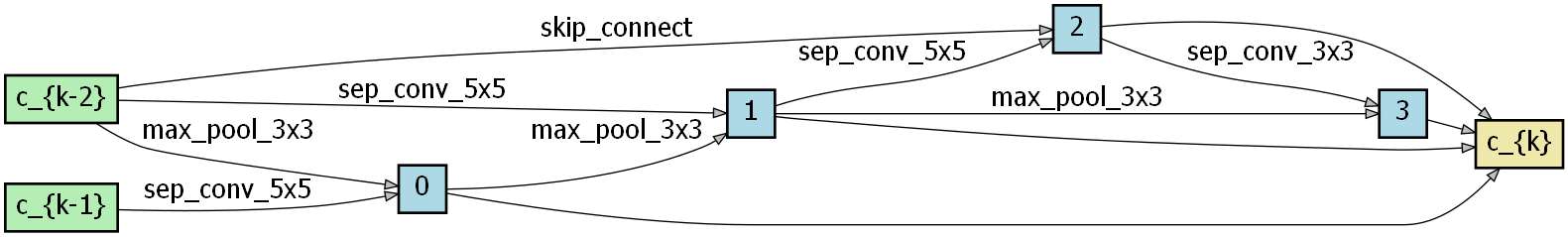}%
	}
	\subfloat[Reduction Cell in DARTS$\_$Hessian]{\includegraphics[width=2.35in]{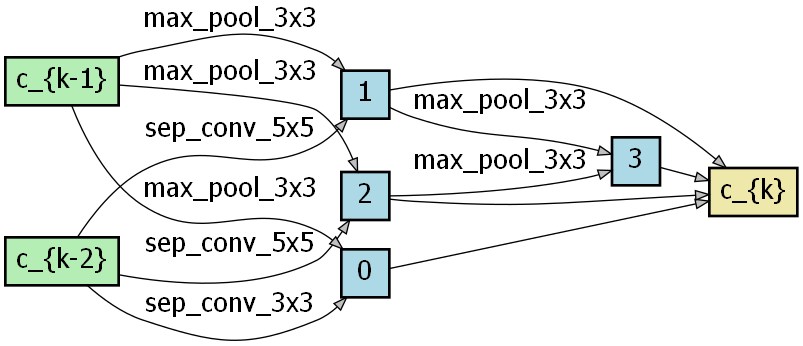}%
	}
	\caption{The visualization of the searched architectures using DARTS under different robustness evaluation}
	\label{arch1}
\end{figure*}

\begin{figure*}[h]
	\centering
	\subfloat[Normal Cell in PC-DARTS$\_$FGSM]{\includegraphics[width=2.35in]{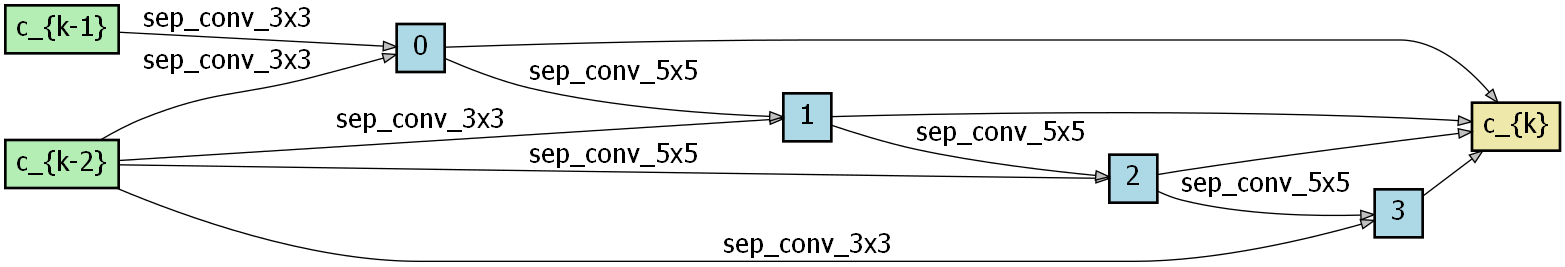}%
	}
	\hfil
	\subfloat[Normal Cell in PC-DARTS$\_$PGD]{\includegraphics[width=2.35in]{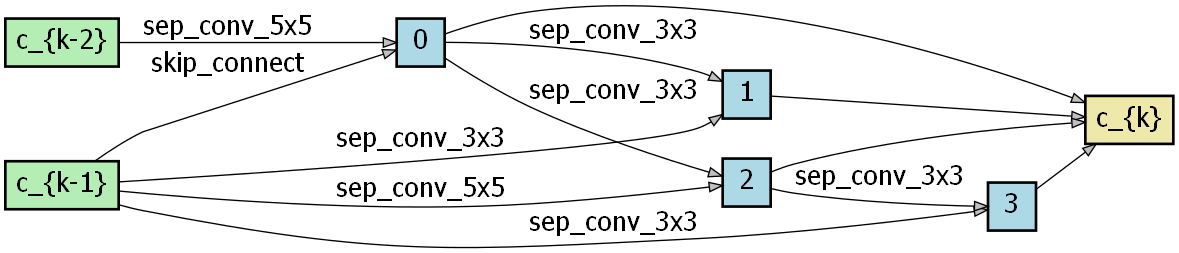}%
	}
	\subfloat[Normal Cell in PC-DARTS$\_$Natural]{\includegraphics[width=2.35in]{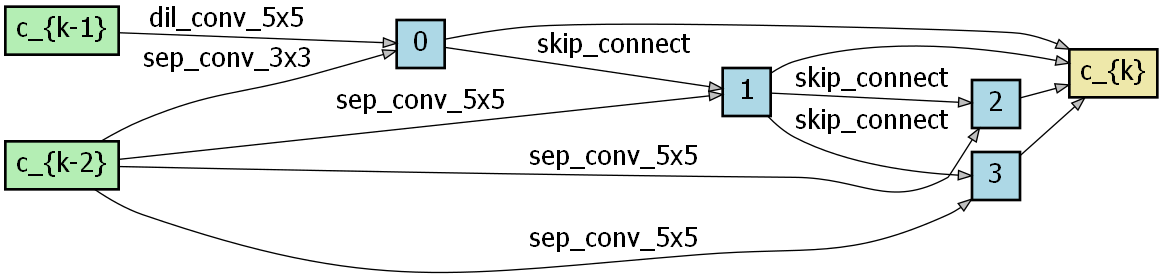}%
	}
	
	\subfloat[Reduction Cell in PC-DARTS$\_$FGSM]{\includegraphics[width=2.35in]{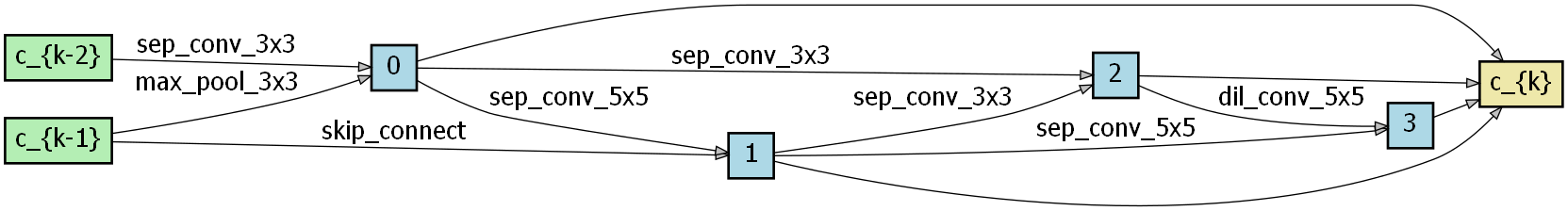}%
	}
	\hfil
	\subfloat[Reduction Cell in PC-DARTS$\_$PGD]{\includegraphics[width=2.35in]{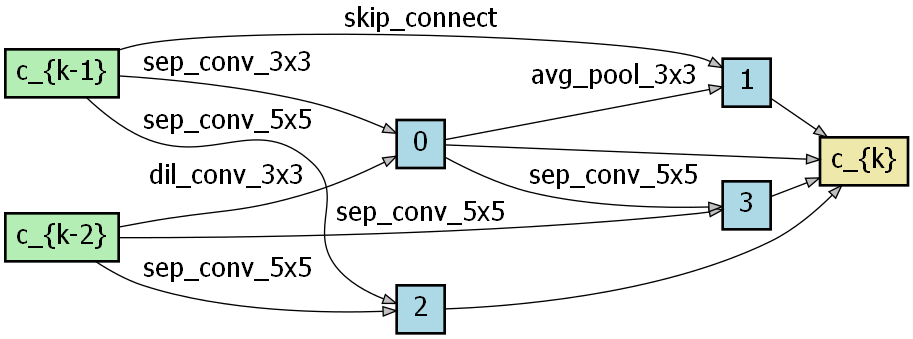}%
	}
	\subfloat[Reduction Cell in PC-DARTS$\_$Natural]{\includegraphics[width=2.35in]{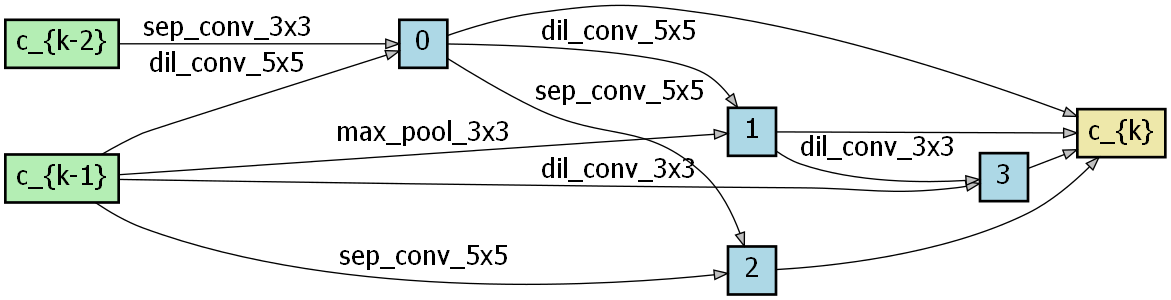}%
	}
	
	\subfloat[Normal Cell in PC-DARTS$\_$System]{\includegraphics[width=2.35in]{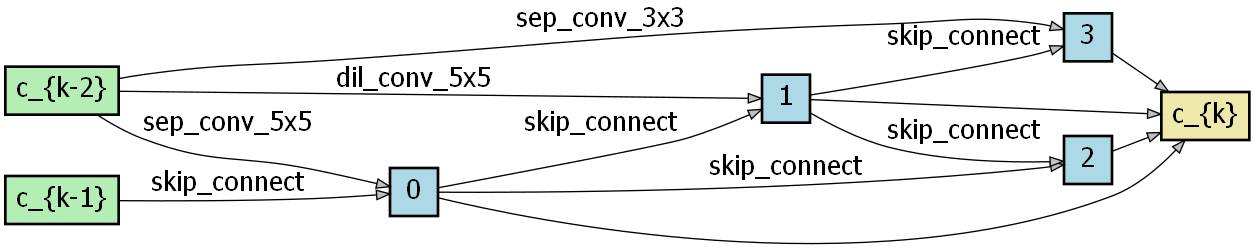}%
	}
	\hfil
	\subfloat[Normal Cell in PC-DARTS$\_$Jacobian]{\includegraphics[width=2.35in]{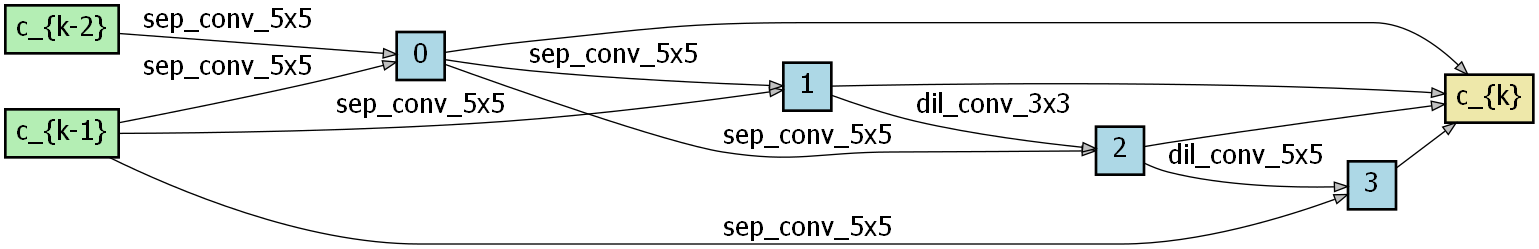}%
	}
	\subfloat[Normal Cell in PC-DARTS$\_$Hessian]{\includegraphics[width=2.35in]{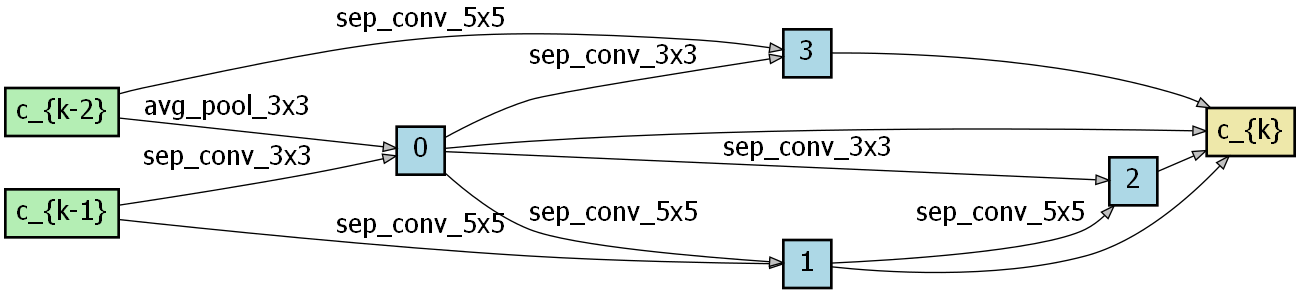}%
	}
	
	\subfloat[Reduction Cell in PC-DARTS$\_$System]{\includegraphics[width=2.35in]{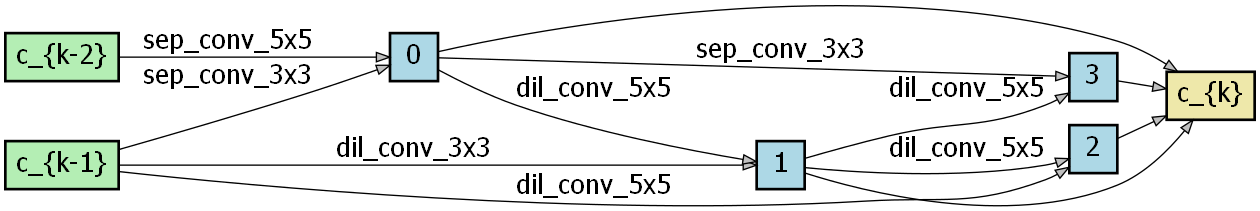}%
	}
	\hfil
	\subfloat[Reduction Cell in PC-DARTS$\_$Jacobian]{\includegraphics[width=2.35in]{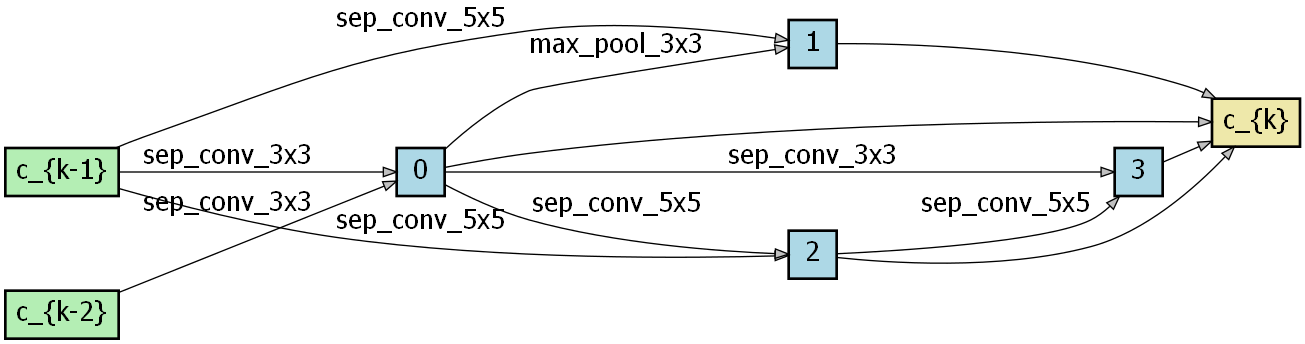}%
	}
	\subfloat[Reduction Cell in PC-DARTS$\_$Hessian]{\includegraphics[width=2.35in]{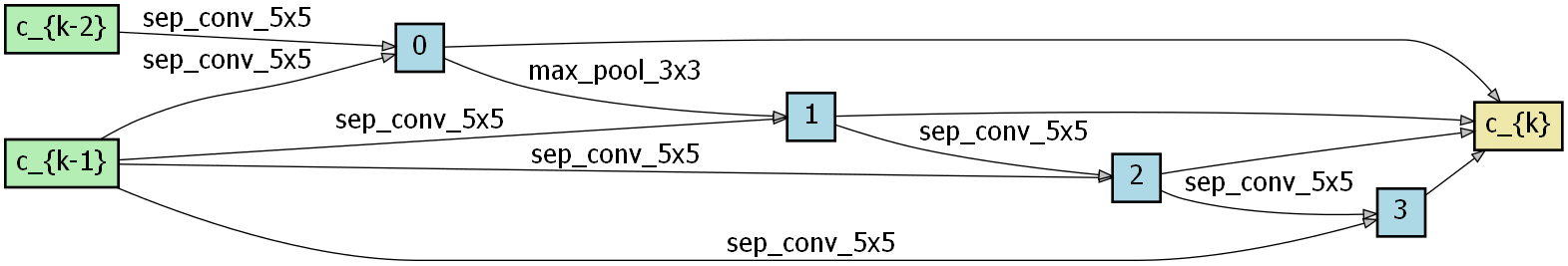}%
	}
	\caption{The visualization of the searched architectures using PC-DARTS under different robustness evaluation}
	\label{arch2}
\end{figure*}

\begin{figure*}[h]
	\centering
	\subfloat[Normal Cell in NASP$\_$FGSM]{\includegraphics[width=2.35in]{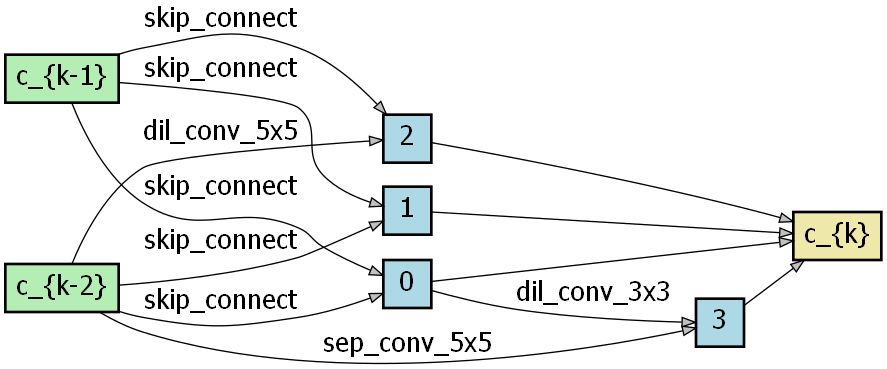}%
	}
	\hfil
	\subfloat[Normal Cell in NASP$\_$PGD]{\includegraphics[width=2.35in]{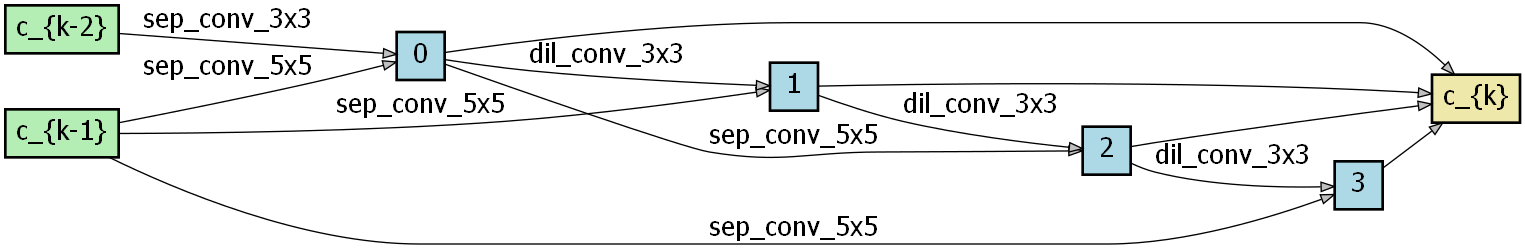}%
	}
	\subfloat[Normal Cell in NASP$\_$Natural]{\includegraphics[width=2.35in]{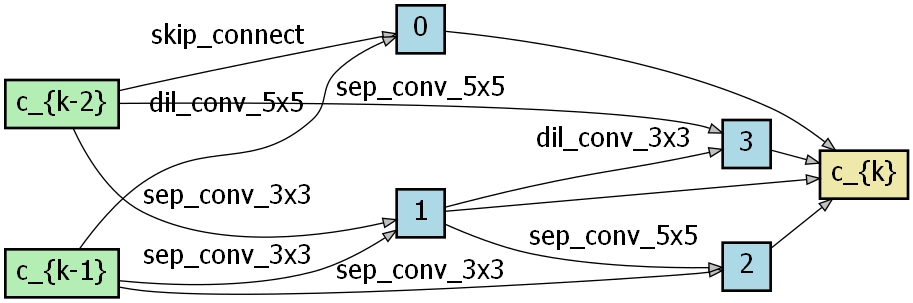}%
	}
	
	\subfloat[Reduction Cell in NASP$\_$FGSM]{\includegraphics[width=2.35in]{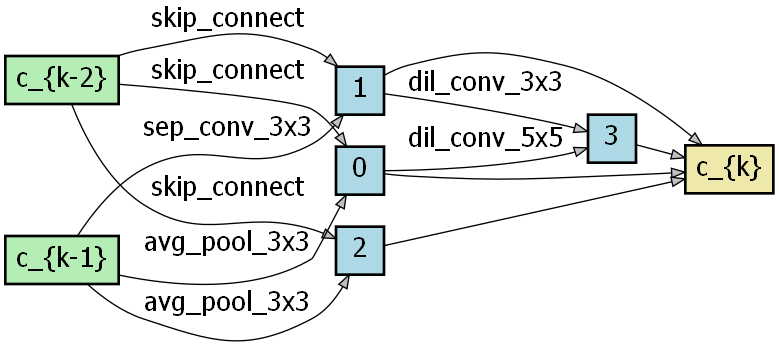}%
	}
	\hfil
	\subfloat[Reduction Cell in NASP$\_$PGD]{\includegraphics[width=2.35in]{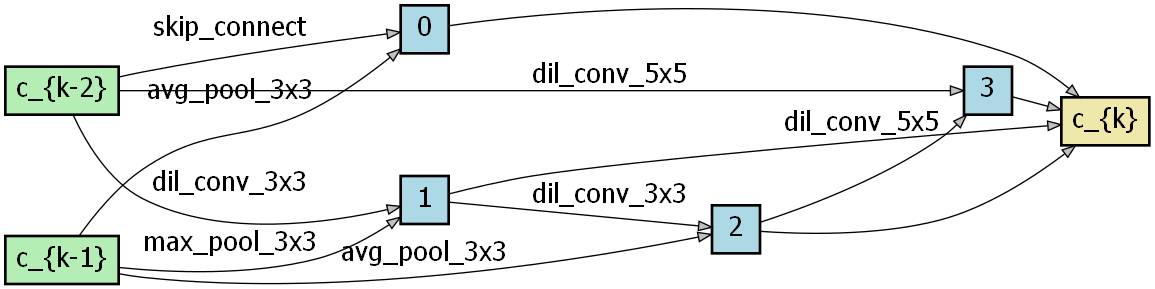}%
	}
	\subfloat[Reduction Cell in NASP$\_$Natural]{\includegraphics[width=2.35in]{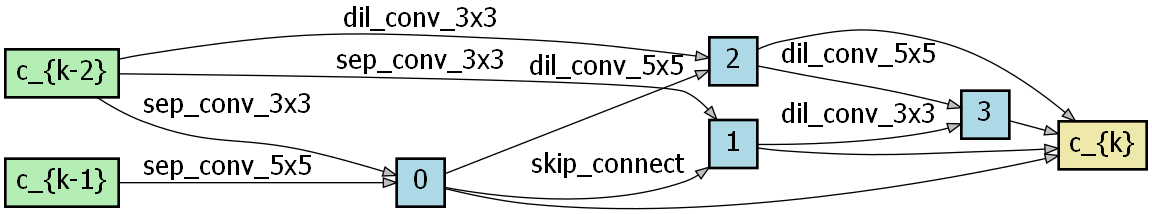}%
	}
	
	\subfloat[Normal Cell in NASP$\_$System]{\includegraphics[width=2.35in]{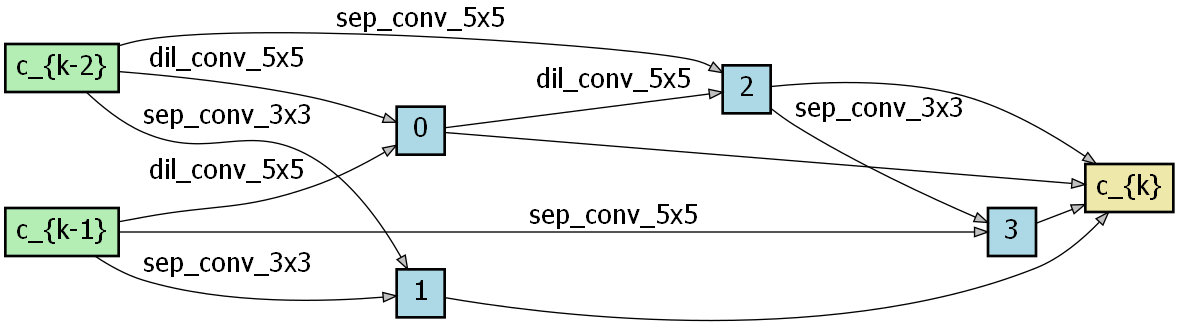}%
	}
	\hfil
	\subfloat[Normal Cell in NASP$\_$Jacobian]{\includegraphics[width=2.35in]{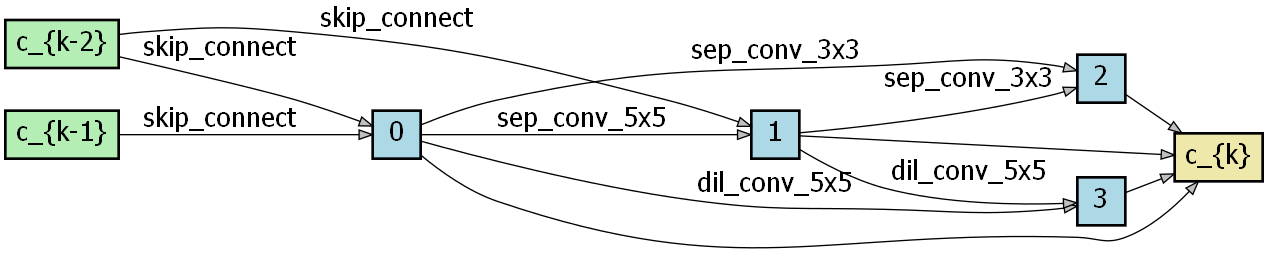}%
	}
	\subfloat[Normal Cell in NASP$\_$Hessian]{\includegraphics[width=2.35in]{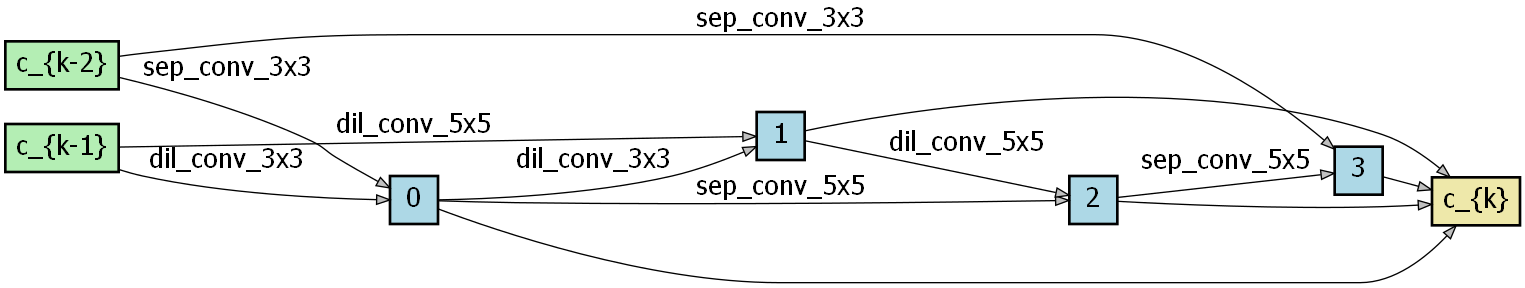}%
	}
	
	\subfloat[Reduction Cell in NASP$\_$System]{\includegraphics[width=2.35in]{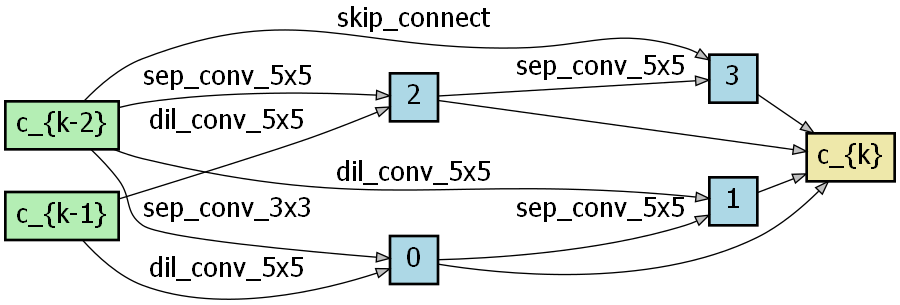}%
	}
	\hfil
	\subfloat[Reduction Cell in NASP$\_$Jacobian]{\includegraphics[width=2.35in]{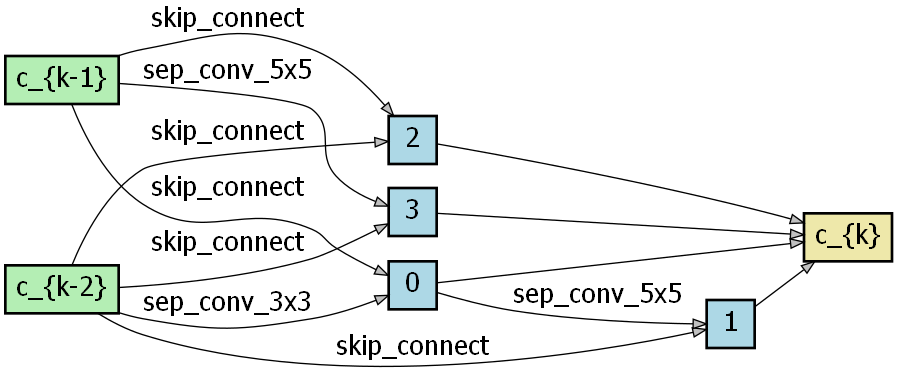}%
	}
	\subfloat[Reduction Cell in NASP$\_$Hessian]{\includegraphics[width=2.35in]{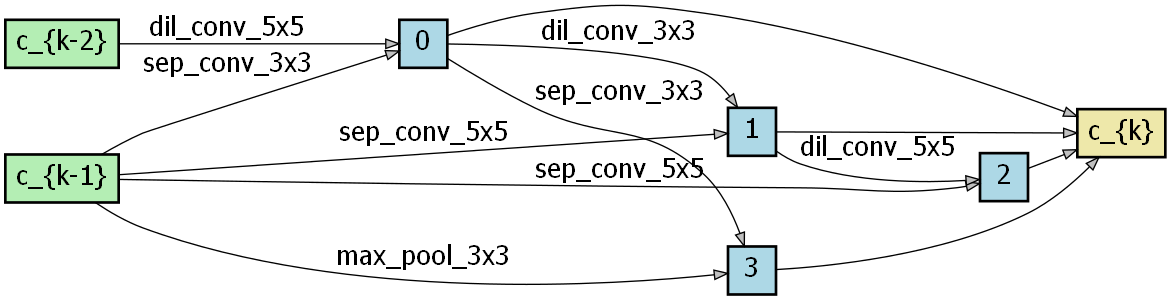}%
	}
	\caption{The visualization of the searched architectures using NASP under different robustness evaluation}
	\label{arch3}
\end{figure*}

\begin{figure*}[h]
	\centering
	\subfloat[Normal Cell in FairDARTS$\_$FGSM]{\includegraphics[width=2.35in]{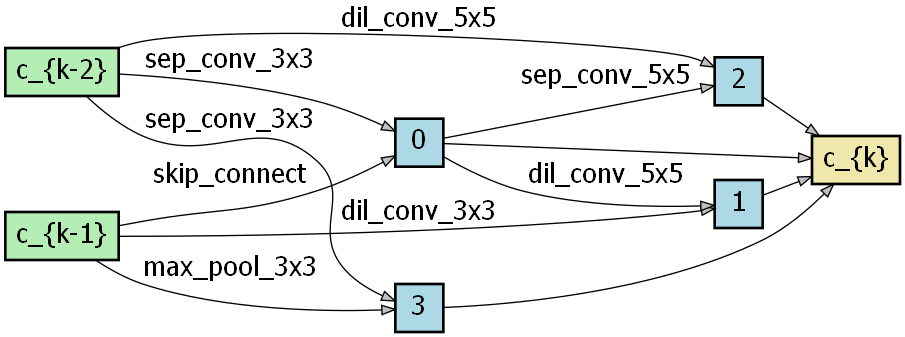}%
	}
	\hfil
	\subfloat[Normal Cell in FairDARTS$\_$PGD]{\includegraphics[width=2.35in]{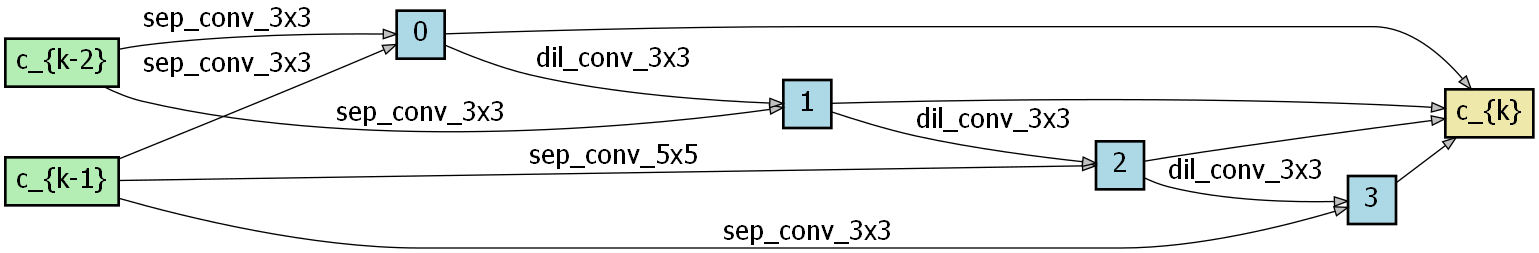}%
	}
	\subfloat[Normal Cell in FairDARTS$\_$Natural]{\includegraphics[width=2.35in]{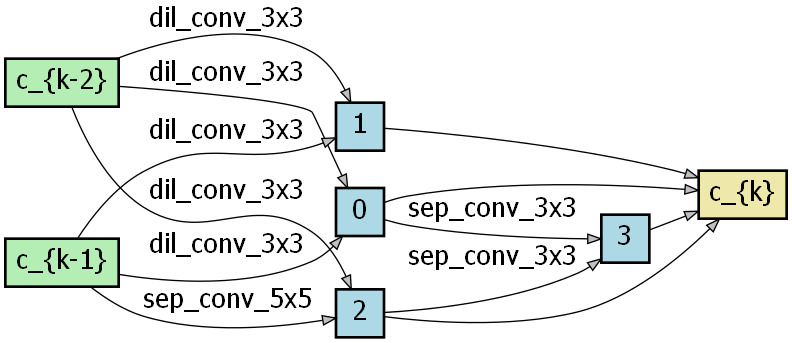}%
	}
	
	\subfloat[Reduction Cell in FairDARTS$\_$FGSM]{\includegraphics[width=2.35in]{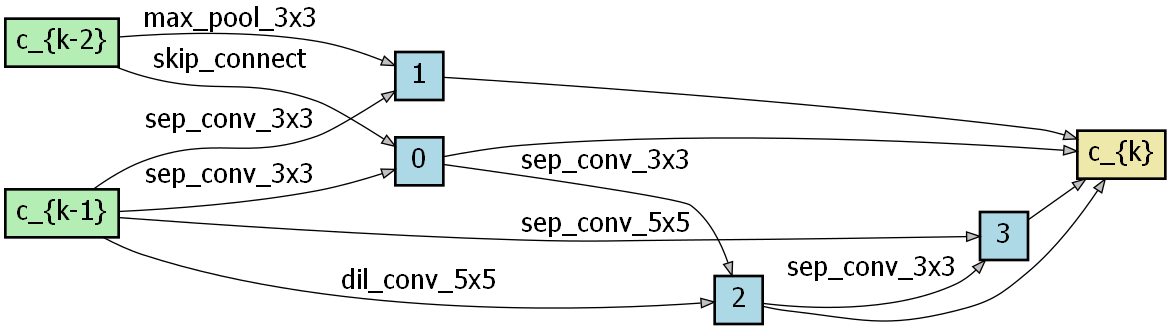}%
	}
	\hfil
	\subfloat[Reduction Cell in FairDARTS$\_$PGD]{\includegraphics[width=2.35in]{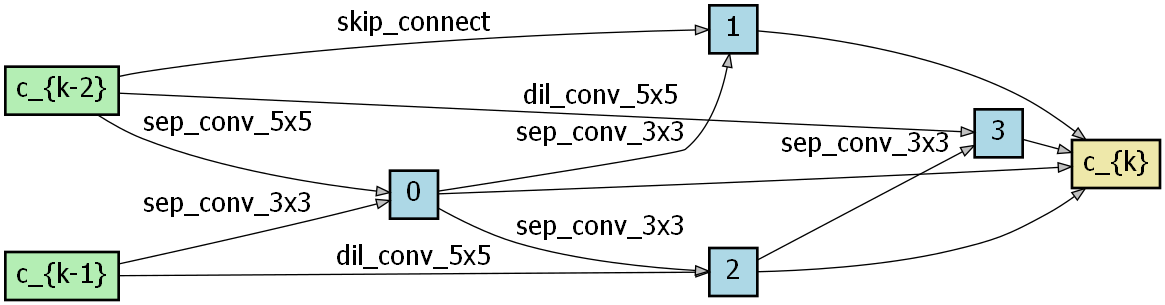}%
	}
	\subfloat[Reduction Cell in FairDARTS$\_$Natural]{\includegraphics[width=2.35in]{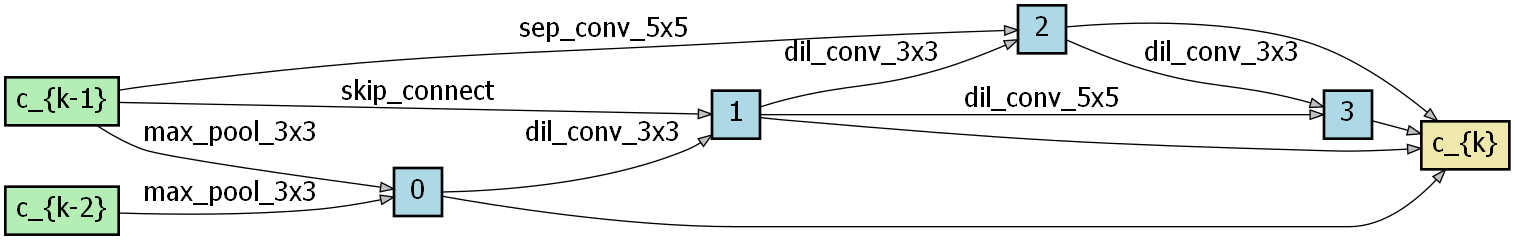}%
	}
	
	\subfloat[Normal Cell in FairDARTS$\_$System]{\includegraphics[width=2.35in]{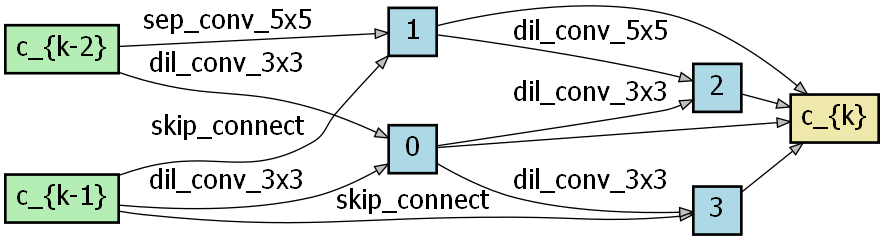}%
	}
	\hfil
	\subfloat[Normal Cell in FairDARTS$\_$Jacobian]{\includegraphics[width=2.35in]{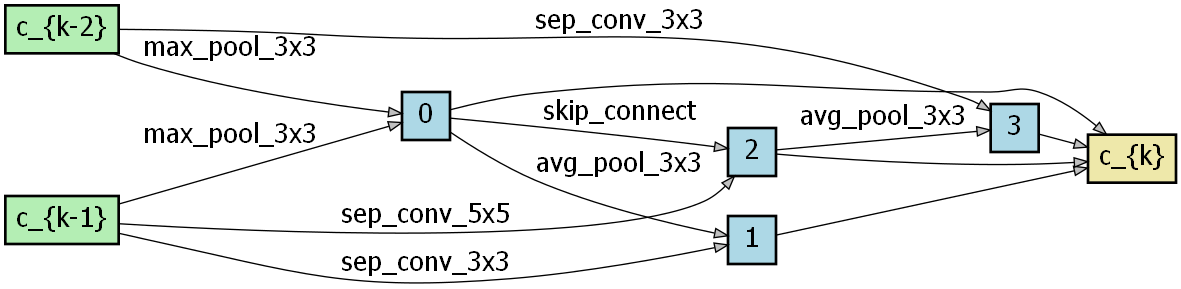}%
	}
	\subfloat[Normal Cell in FairDARTS$\_$Hessian]{\includegraphics[width=2.35in]{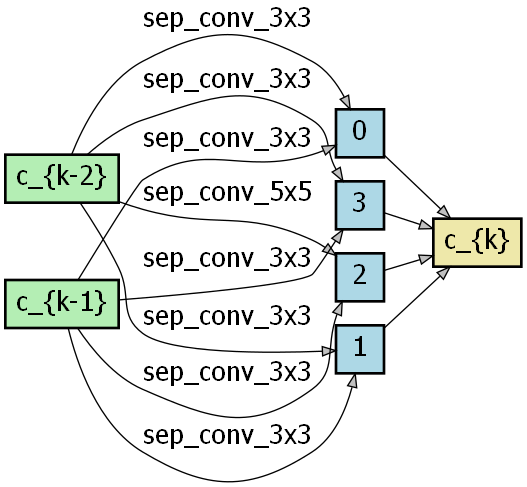}%
	}
	
	\subfloat[Reduction Cell in FairDARTS$\_$System]{\includegraphics[width=2.35in]{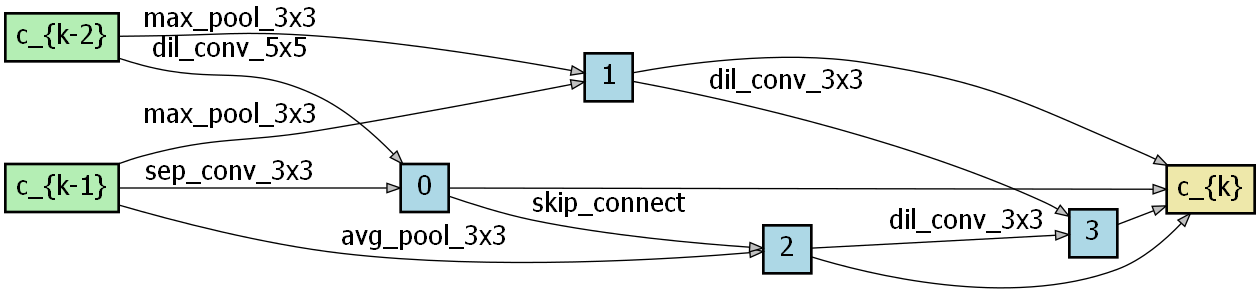}%
	}
	\hfil
	\subfloat[Reduction Cell in FairDARTS$\_$Jacobian]{\includegraphics[width=2.35in]{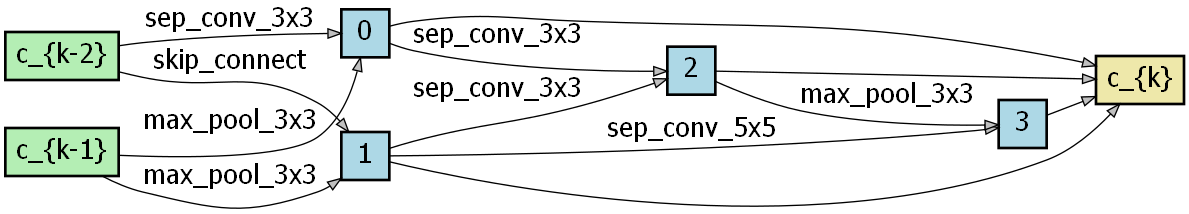}%
	}
	\subfloat[Reduction Cell in FairDARTS$\_$Hessian]{\includegraphics[width=2.35in]{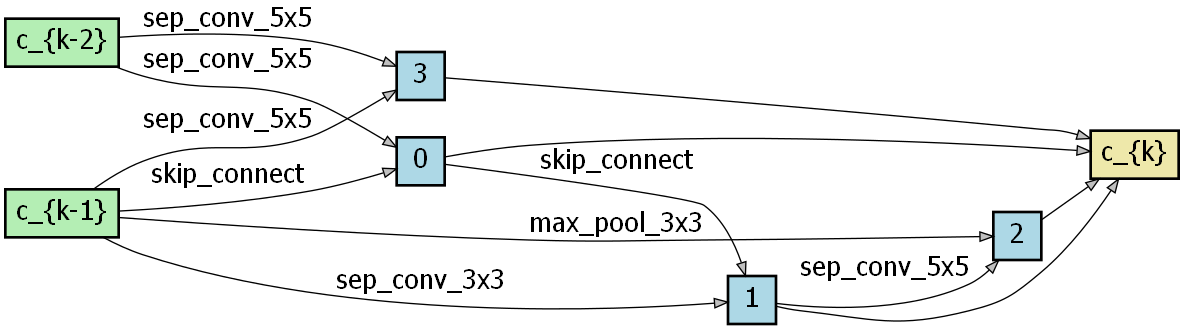}%
	}
	\caption{The visualization of the searched architectures using FairDARTS under different robustness evaluation}
	\label{arch4}
\end{figure*}

\begin{figure*}[h]
	\centering
	\subfloat[Normal Cell in SmoothDARTS$\_$FGSM]{\includegraphics[width=2.35in]{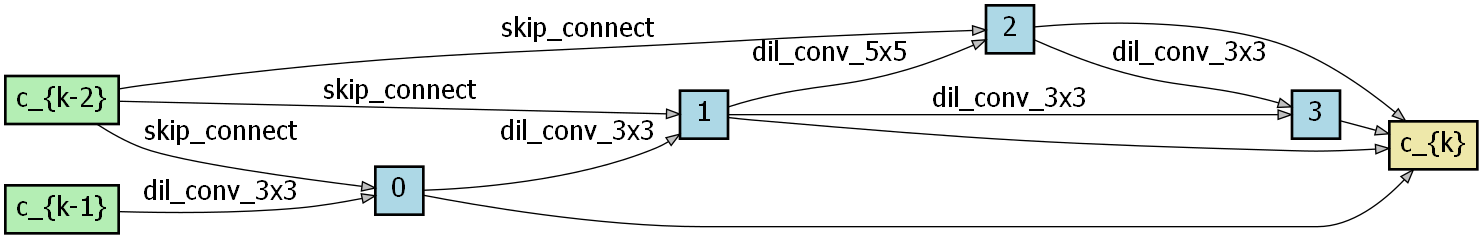}%
	}
	\hfil
	\subfloat[Normal Cell in SmoothDARTS$\_$PGD]{\includegraphics[width=2.35in]{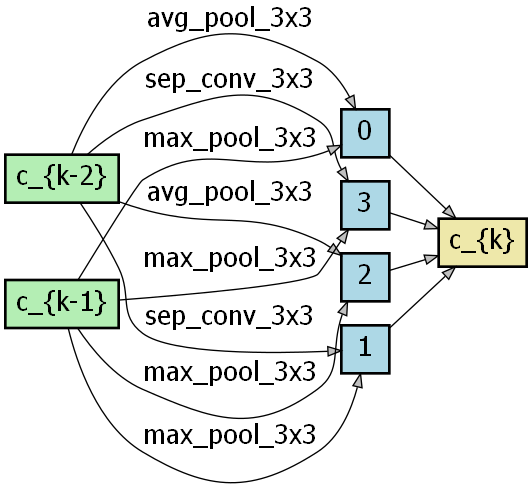}%
	}
	\subfloat[Normal Cell in SmoothDARTS$\_$Natural]{\includegraphics[width=2.35in]{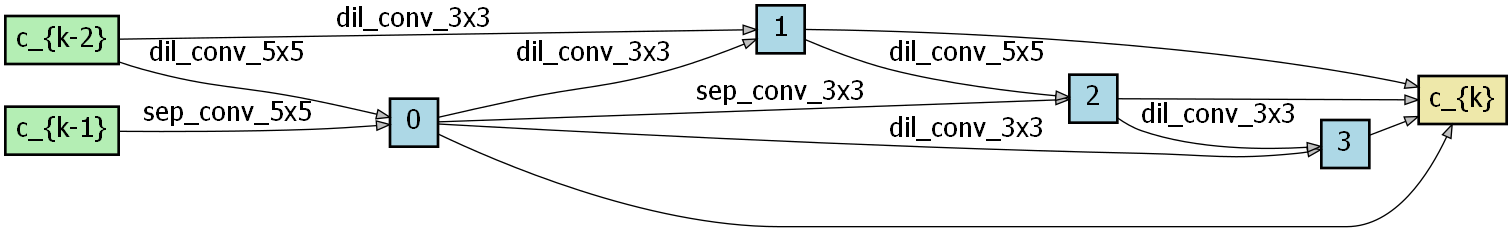}%
	}
	
	\subfloat[Reduction Cell in SmoothDARTS$\_$FGSM]{\includegraphics[width=2.35in]{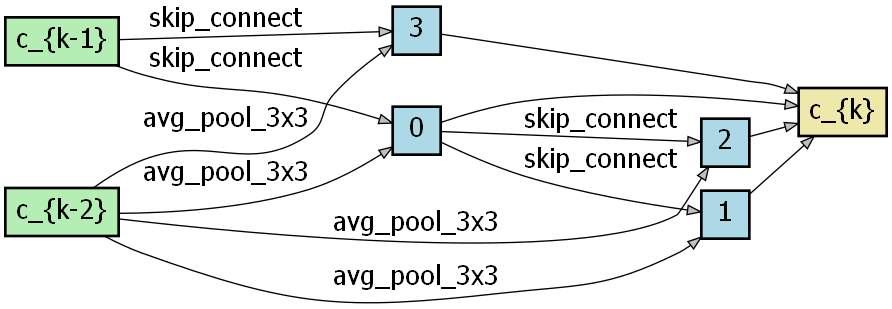}%
	}
	\hfil
	\subfloat[Reduction Cell in SmoothDARTS$\_$PGD]{\includegraphics[width=2.35in]{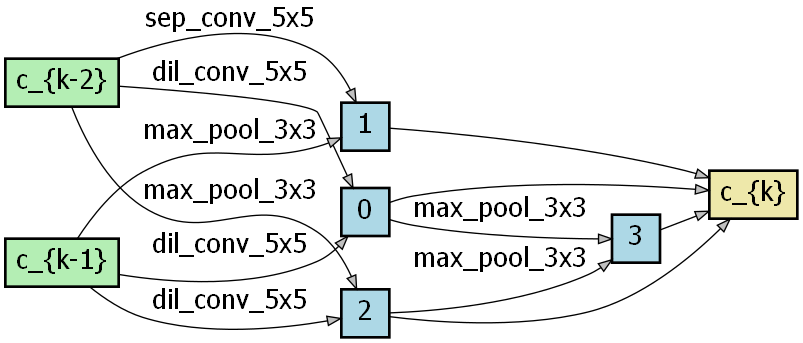}%
	}
	\subfloat[Reduction Cell in SmoothDARTS$\_$Natural]{\includegraphics[width=2.35in]{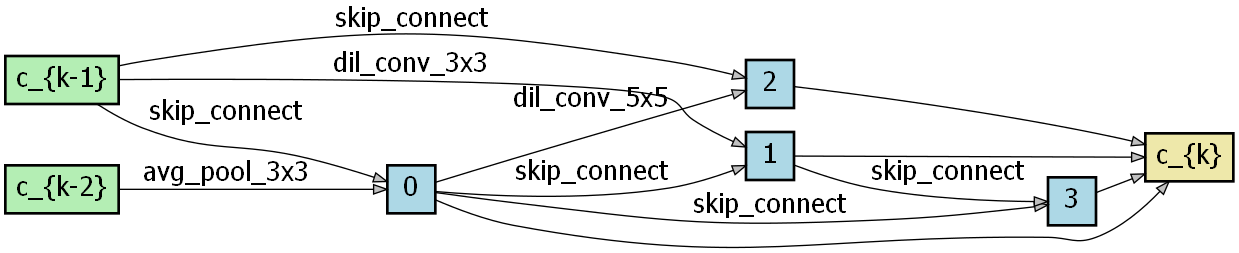}%
	}
	
	\subfloat[Normal Cell in SmoothDARTS$\_$System]{\includegraphics[width=2.35in]{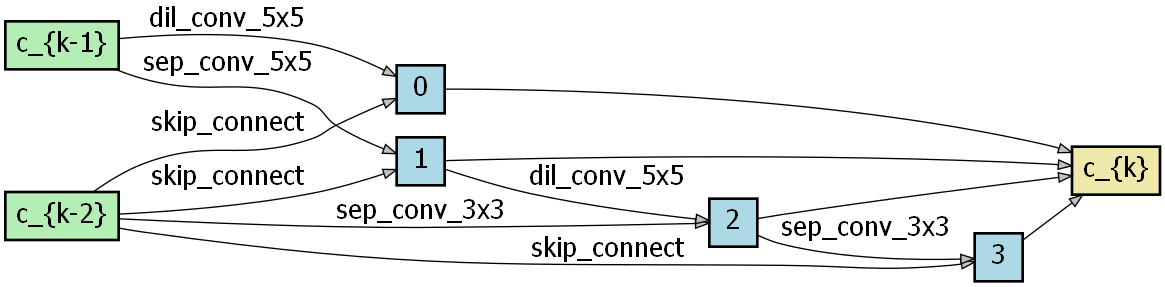}%
	}
	\hfil
	\subfloat[Normal Cell in SmoothDARTS$\_$Jacobian]{\includegraphics[width=2.35in]{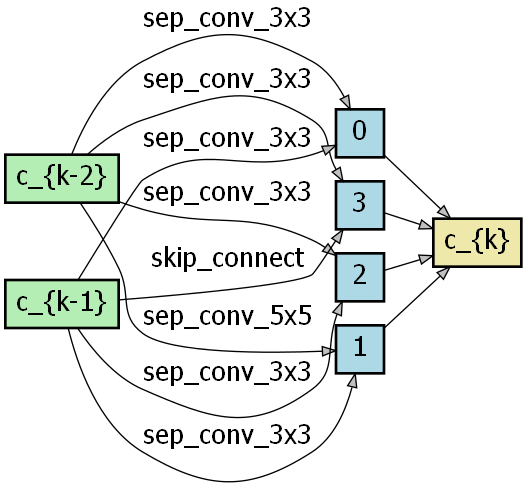}%
	}
	\subfloat[Normal Cell in SmoothDARTS$\_$Hessian]{\includegraphics[width=2.35in]{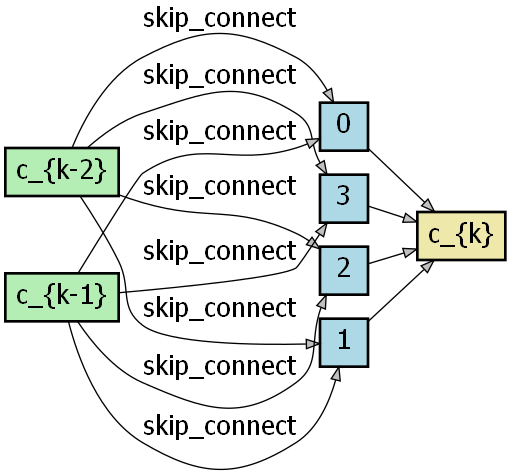}%
	}
	
	\subfloat[Reduction Cell in SmoothDARTS$\_$System]{\includegraphics[width=2.35in]{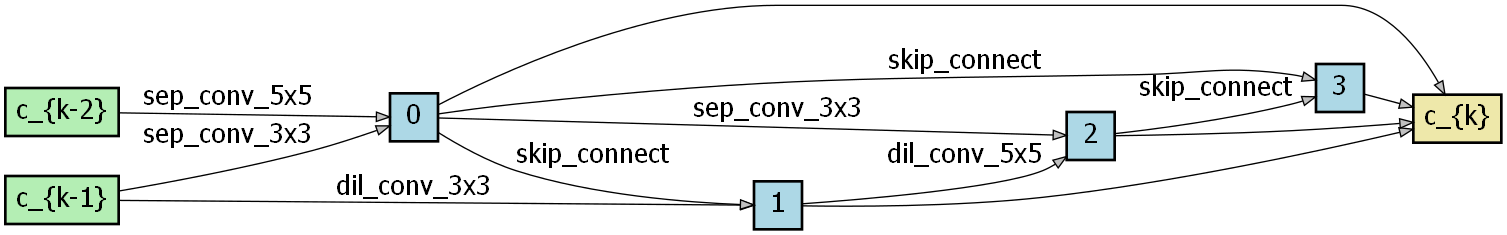}%
	}
	\hfil
	\subfloat[Reduction Cell in SmoothDARTS$\_$Jacobian]{\includegraphics[width=2.35in]{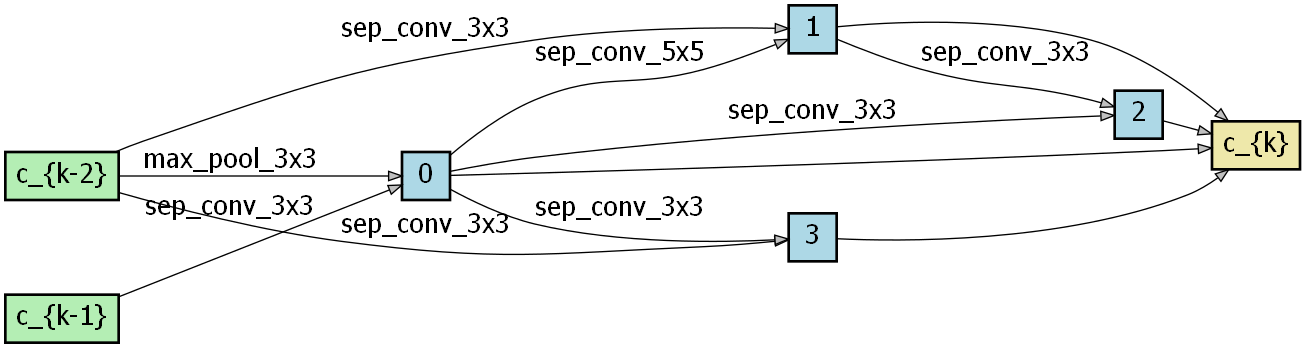}%
	}
	\subfloat[Reduction Cell in SmoothDARTS$\_$Hessian]{\includegraphics[width=2.35in]{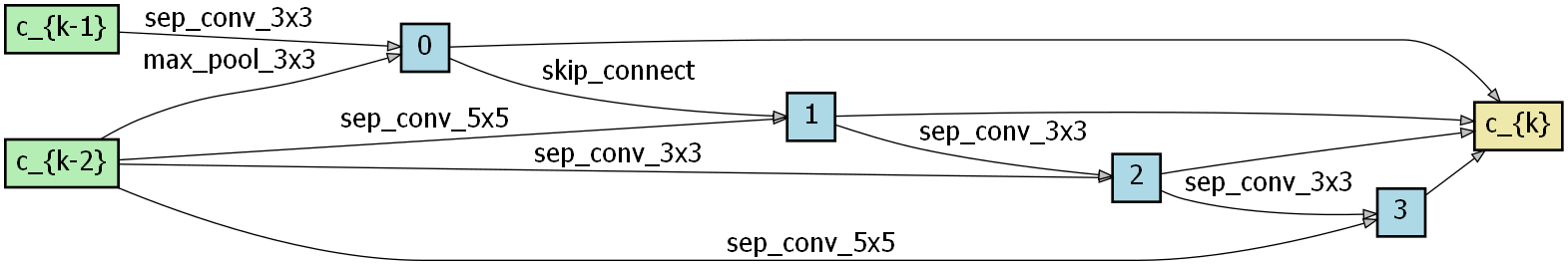}%
	}
	\caption{The visualization of the searched architectures using SmoothDARTS under different robustness evaluation}
	\label{arch5}
\end{figure*}

\begin{figure*}[h]
	\centering
	\subfloat[Normal Cell in Random$\_$search$\_$FGSM]{\includegraphics[width=2.35in]{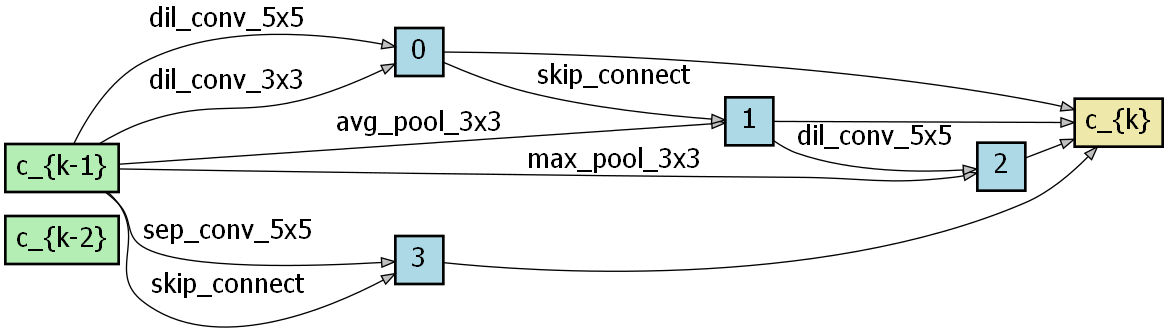}%
	}
	\hfil
	\subfloat[Normal Cell in Random$\_$search$\_$PGD]{\includegraphics[width=2.35in]{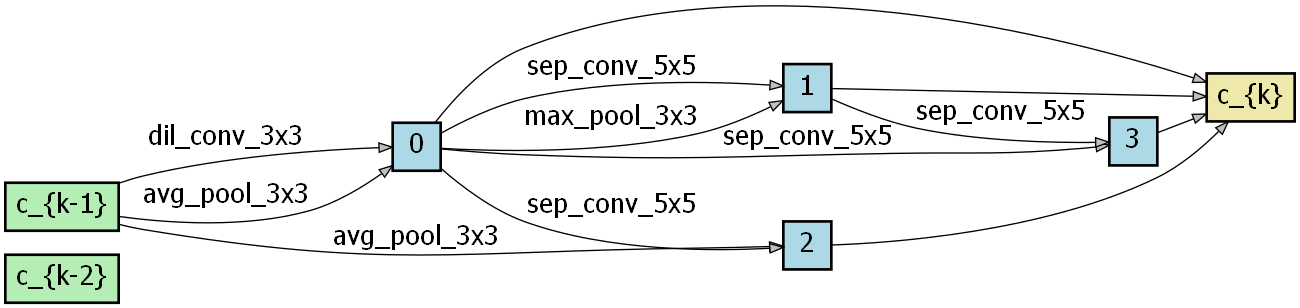}%
	}
	\subfloat[Normal Cell in Random$\_$search$\_$Natural]{\includegraphics[width=2.35in]{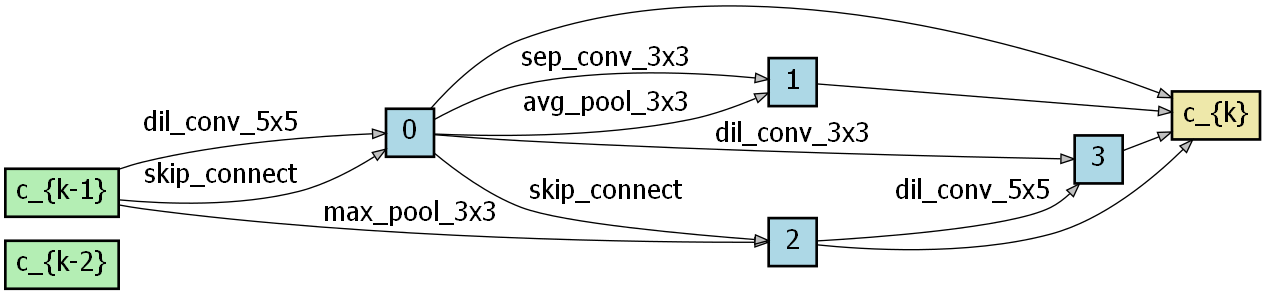}%
	}
	
	\subfloat[Reduction Cell in Random$\_$search$\_$FGSM]{\includegraphics[width=2.35in]{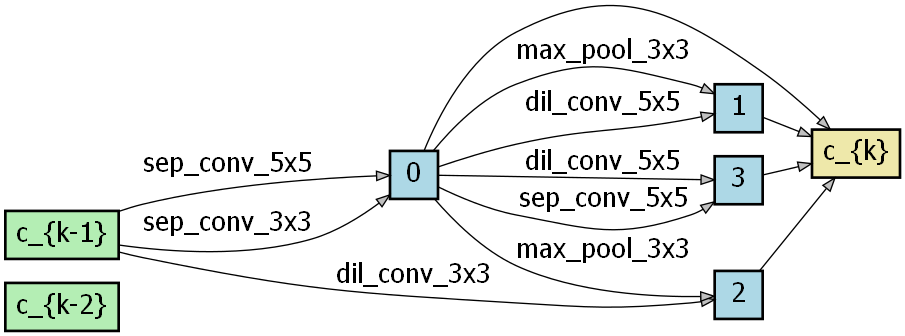}%
	}
	\hfil
	\subfloat[Reduction Cell in Random$\_$search$\_$PGD]{\includegraphics[width=2.35in]{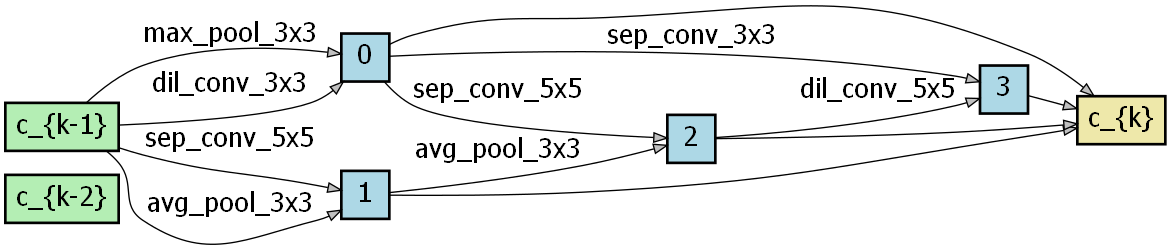}%
	}
	\subfloat[Reduction Cell in Random$\_$search$\_$Natural]{\includegraphics[width=2.35in]{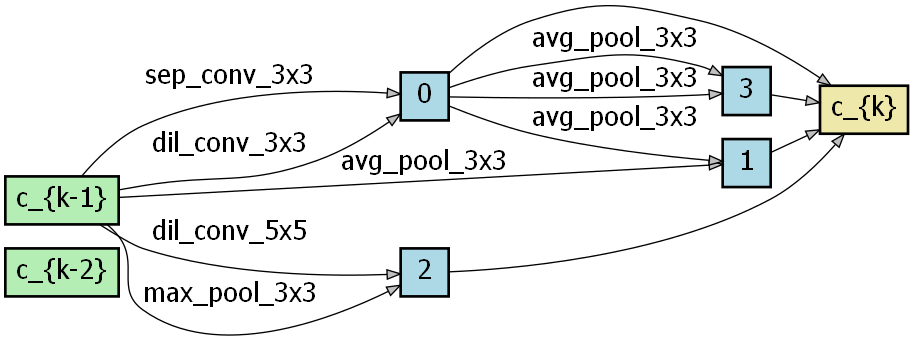}%
	}
	
	\subfloat[Normal Cell in Random$\_$search$\_$System]{\includegraphics[width=2.35in]{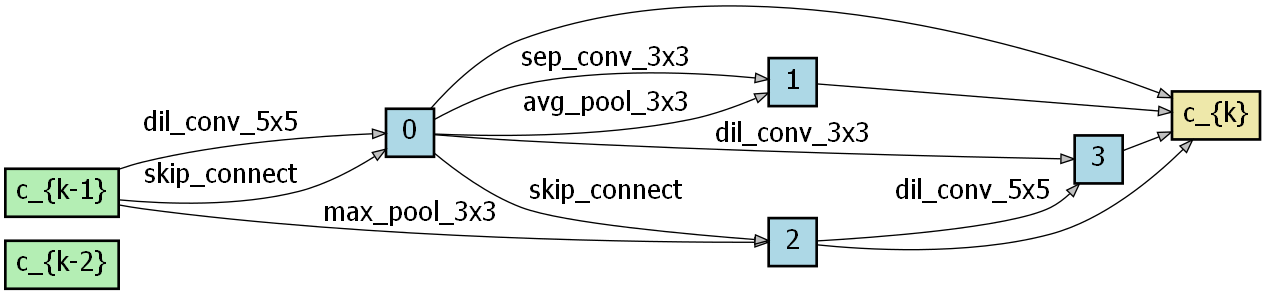}%
	}
	\hfil
	\subfloat[Normal Cell in Random$\_$search$\_$Jacobian]{\includegraphics[width=2.35in]{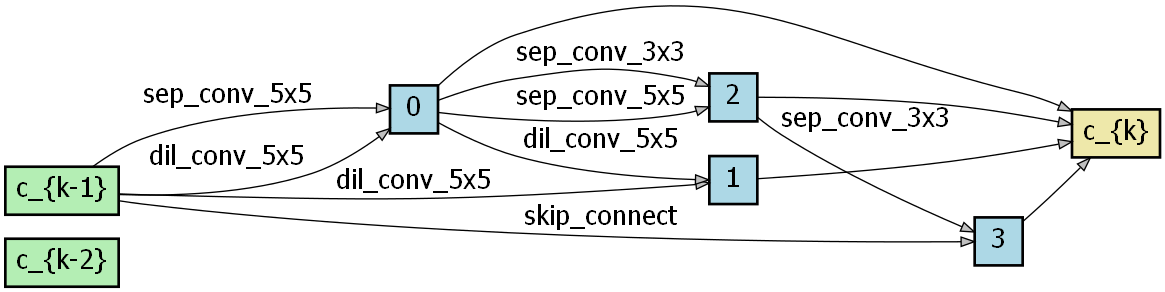}%
	}
	\subfloat[Normal Cell in Random$\_$search$\_$Hessian]{\includegraphics[width=2.35in]{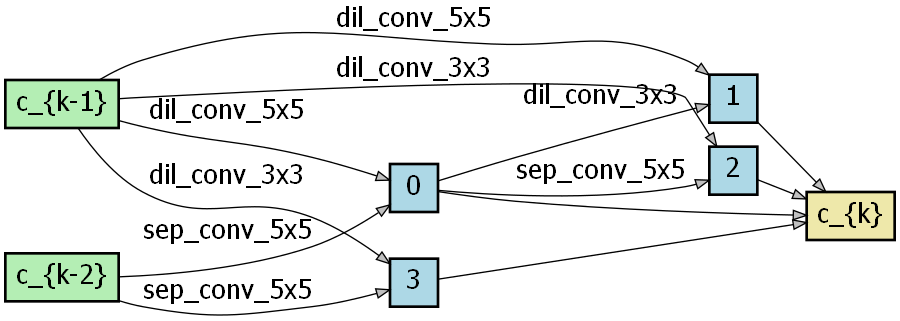}%
	}
	
	\subfloat[Reduction Cell in Random$\_$search$\_$System]{\includegraphics[width=2.35in]{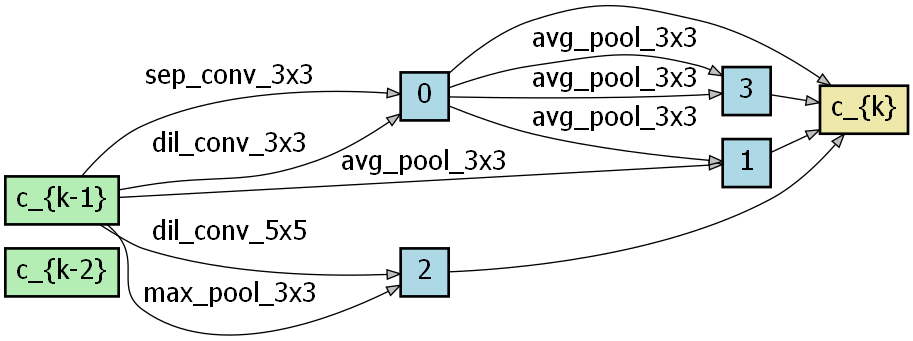}%
	}
	\hfil
	\subfloat[Reduction Cell in Random$\_$search$\_$Jacobian]{\includegraphics[width=2.35in]{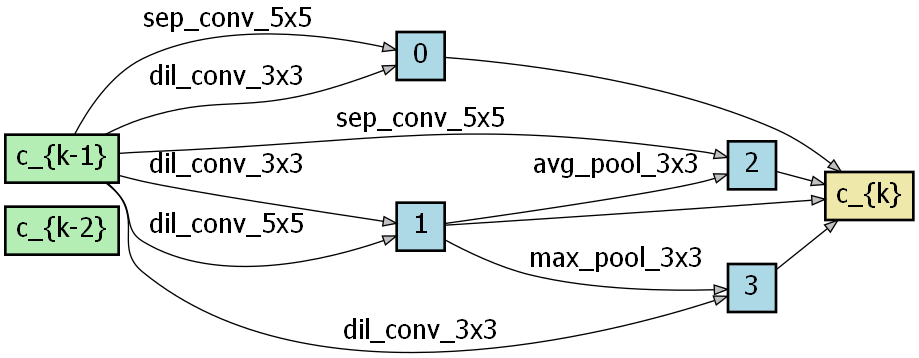}%
	}
	\subfloat[Reduction Cell in Random$\_$search$\_$Hessian]{\includegraphics[width=2.35in]{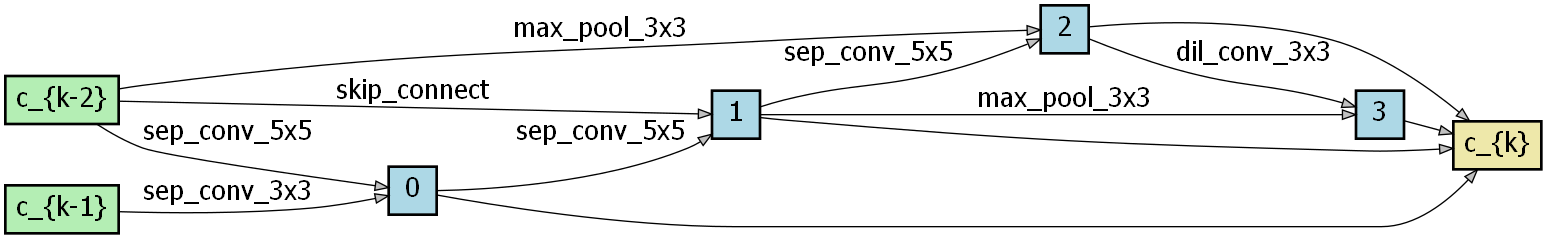}%
	}
	\caption{The visualization of the searched architectures using Random Search under different robustness evaluation}
	\label{arch6}
	
\end{figure*}

\begin{figure*}[h]
	\centering
	\subfloat[Normal Cell in DE$\_$FGSM]{\includegraphics[width=2.35in]{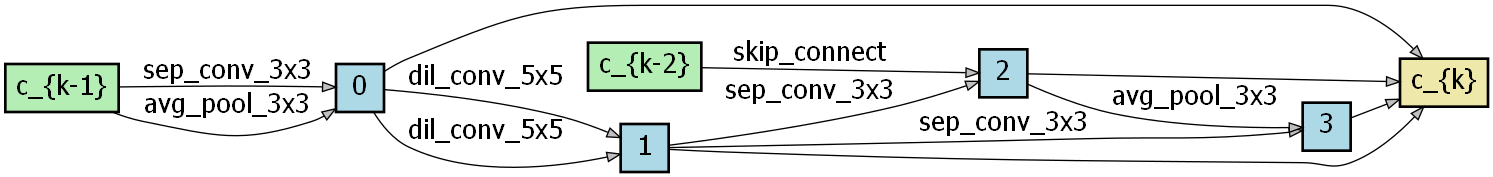}%
	}
	\hfil
	\subfloat[Normal Cell in DE$\_$PGD]{\includegraphics[width=2.35in]{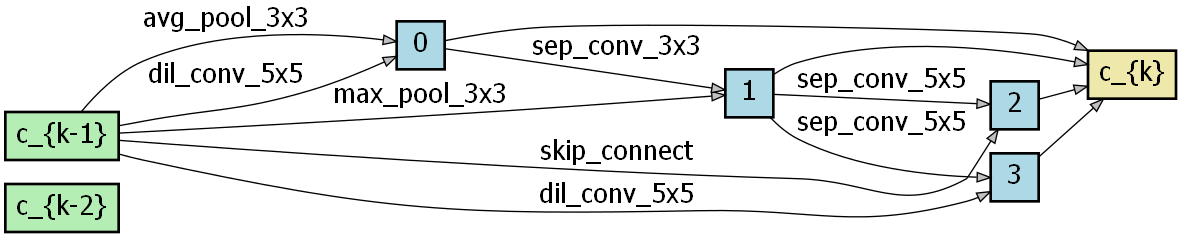}%
	}
	\subfloat[Normal Cell in Random$\_$search$\_$Natural]{\includegraphics[width=2.35in]{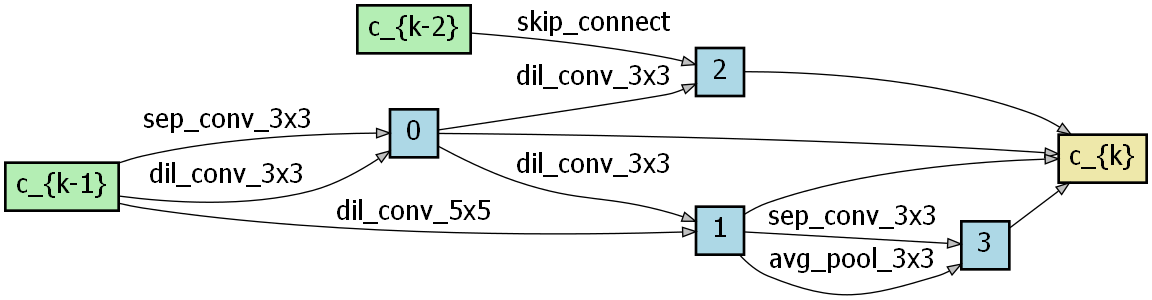}%
	}
	
	\subfloat[Reduction Cell in DE$\_$FGSM]{\includegraphics[width=2.35in]{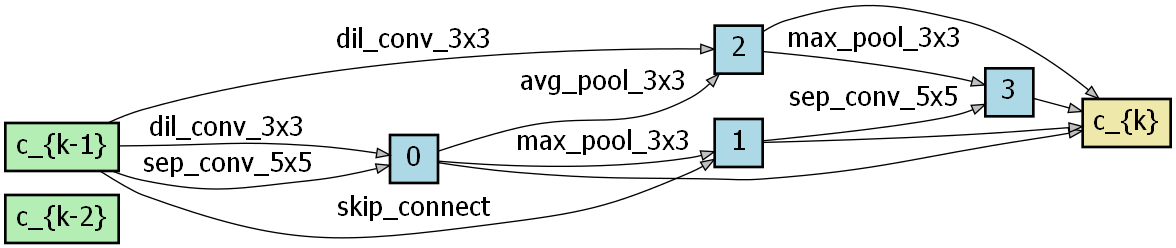}%
	}
	\hfil
	\subfloat[Reduction Cell in DE$\_$PGD]{\includegraphics[width=2.35in]{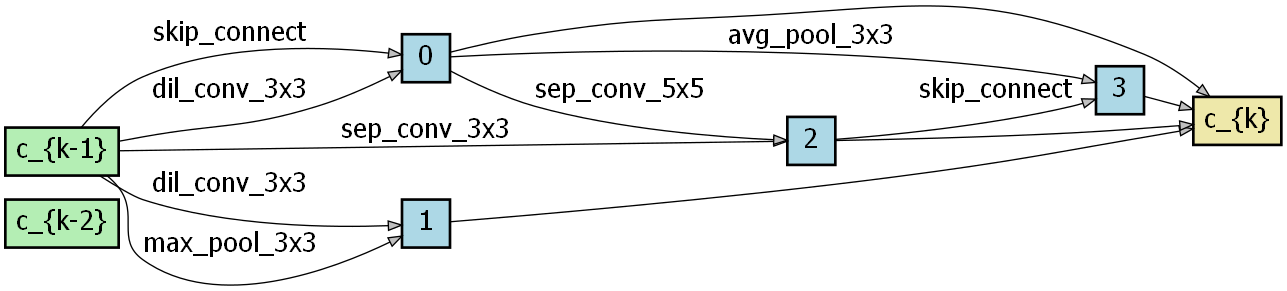}%
	}
	\subfloat[Reduction Cell in DE$\_$Natural]{\includegraphics[width=2.35in]{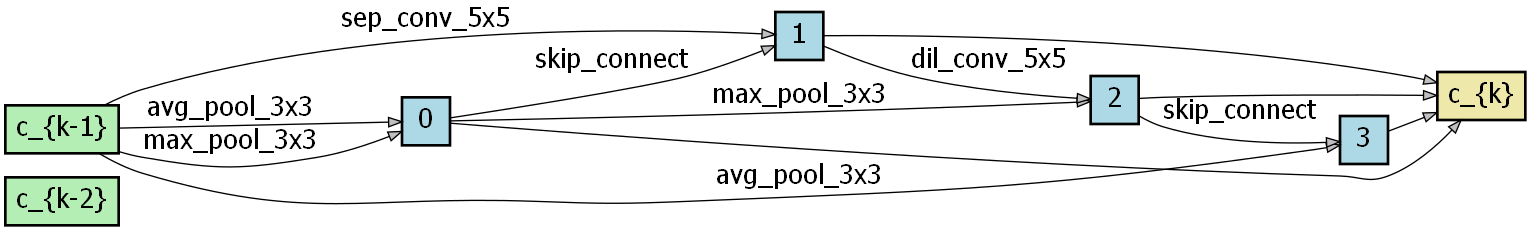}%
	}
	
	\subfloat[Normal Cell in DE$\_$System]{\includegraphics[width=2.35in]{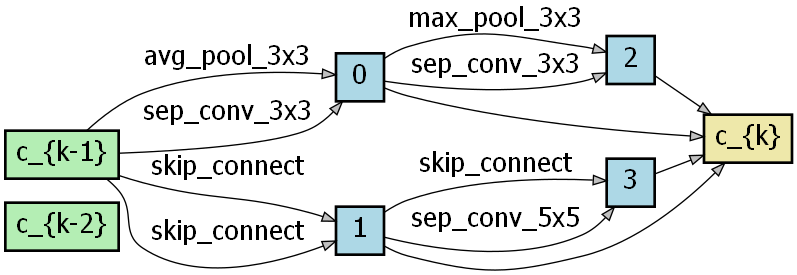}%
	}
	\hfil
	\subfloat[Normal Cell in DE$\_$Jacobian]{\includegraphics[width=2.35in]{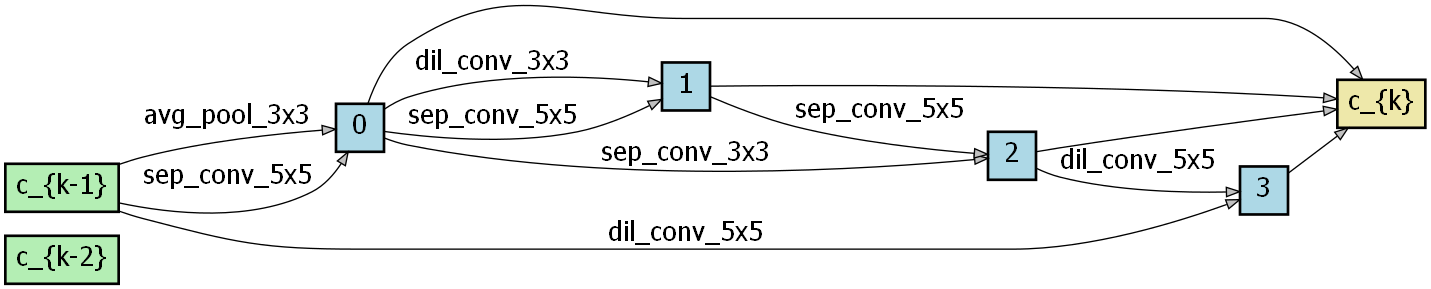}%
	}
	\subfloat[Normal Cell in DE$\_$Hessian]{\includegraphics[width=2.35in]{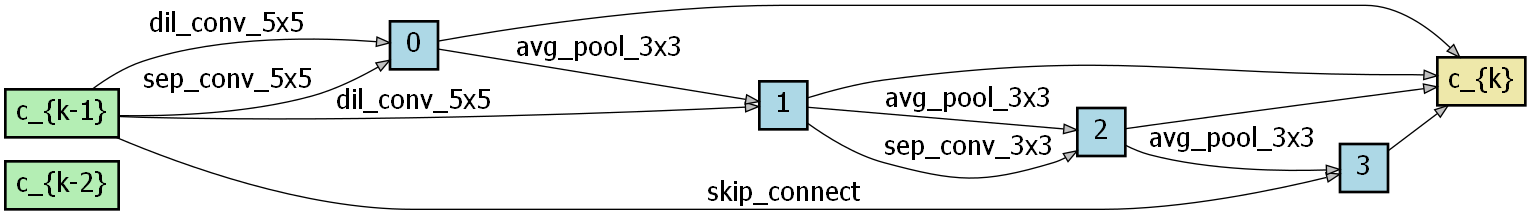}%
	}
	
	\subfloat[Reduction Cell in DE$\_$System]{\includegraphics[width=2.35in]{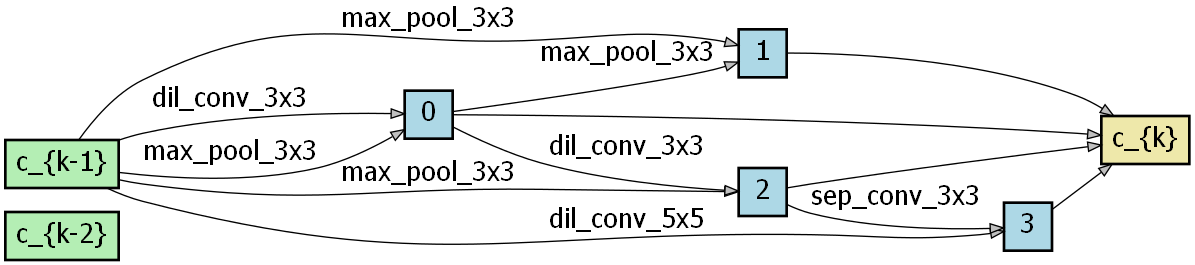}%
	}
	\hfil
	\subfloat[Reduction Cell in DE$\_$Jacobian]{\includegraphics[width=2.35in]{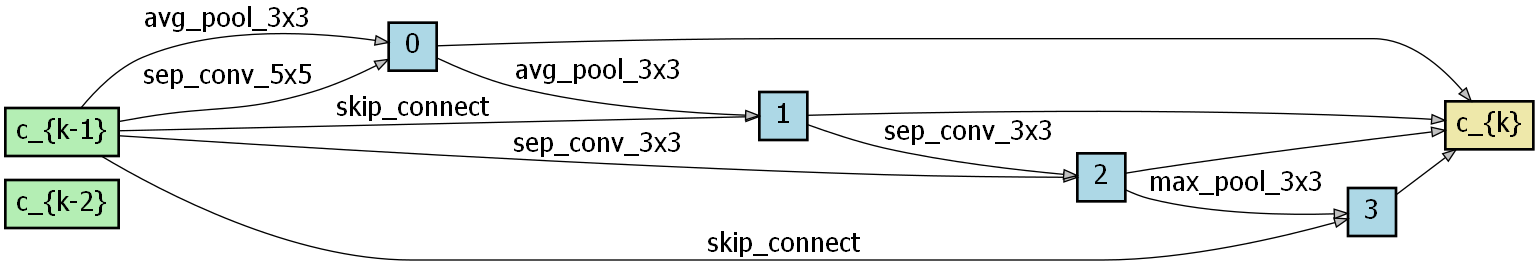}%
	}
	\subfloat[Reduction Cell in DE$\_$Hessian]{\includegraphics[width=2.35in]{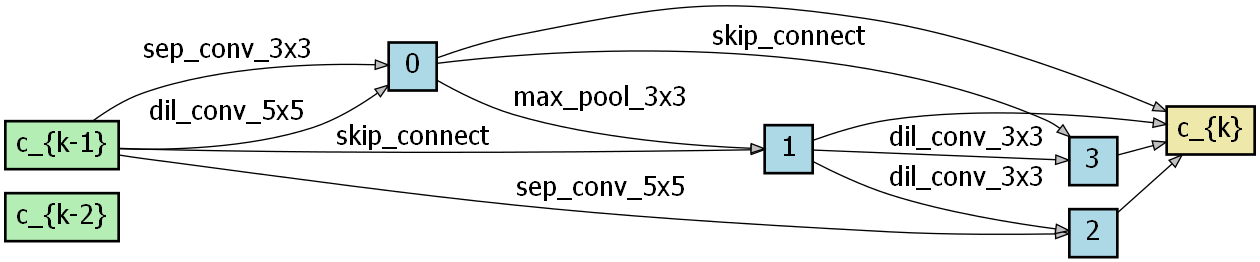}%
	}
	\caption{The visualization of the searched architectures using DE under different robustness evaluation}
	\label{arch8}
\end{figure*}

\begin{figure*}[h]
	\centering
	\subfloat[Normal Cell in Weight Sharing based Random$\_$Search$\_$FGSM]{\includegraphics[width=2.35in]{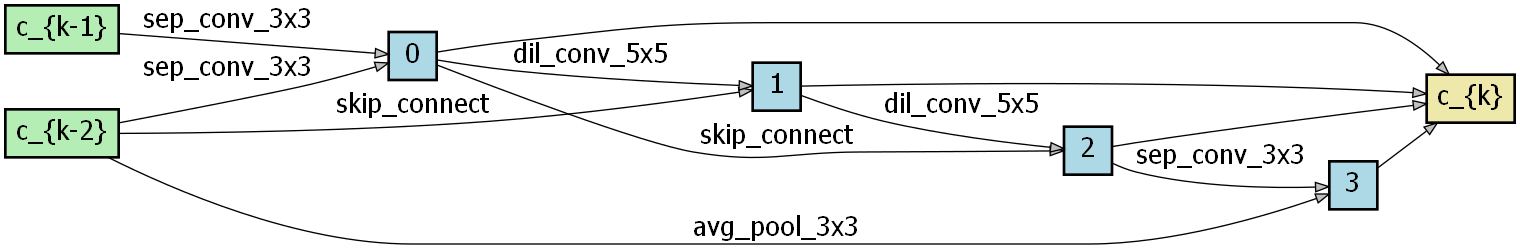}%
	}
	\hfil
	\subfloat[Normal Cell in Weight Sharing based Random$\_$Search$\_$PGD]{\includegraphics[width=2.35in]{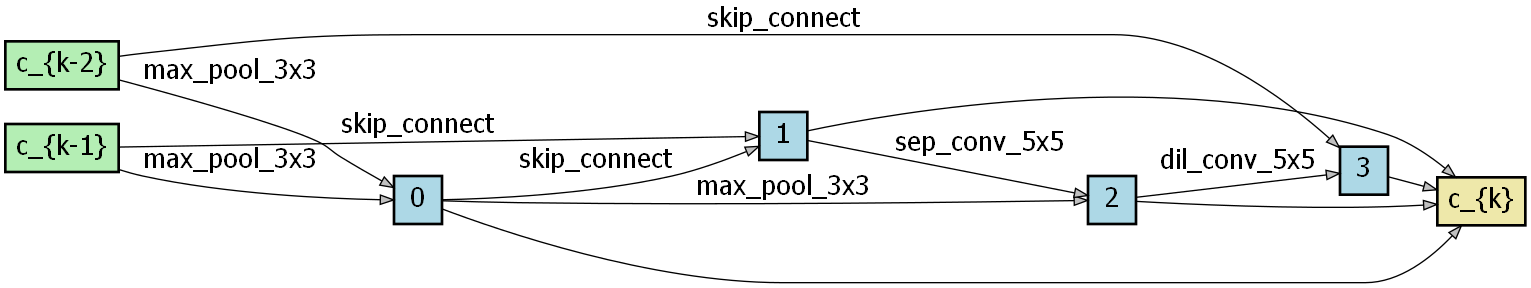}%
	}
	\subfloat[Normal Cell in Weight Sharing based Random$\_$Search$\_$Natural]{\includegraphics[width=2.35in]{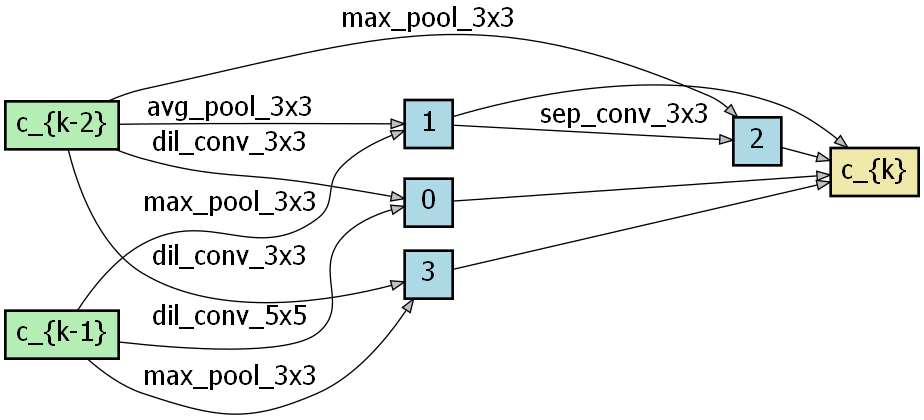}%
	}
	
	\subfloat[Reduction Cell in Weight Sharing based Random$\_$Search$\_$FGSM]{\includegraphics[width=2.35in]{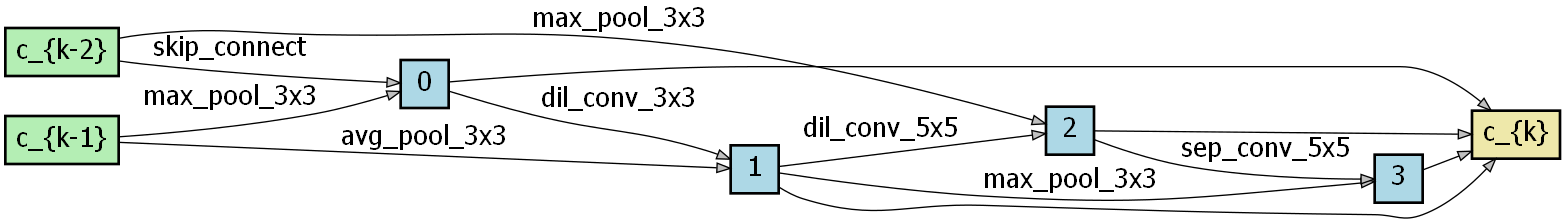}%
	}
	\hfil
	\subfloat[Reduction Cell in Weight Sharing based Random$\_$Search$\_$PGD]{\includegraphics[width=2.35in]{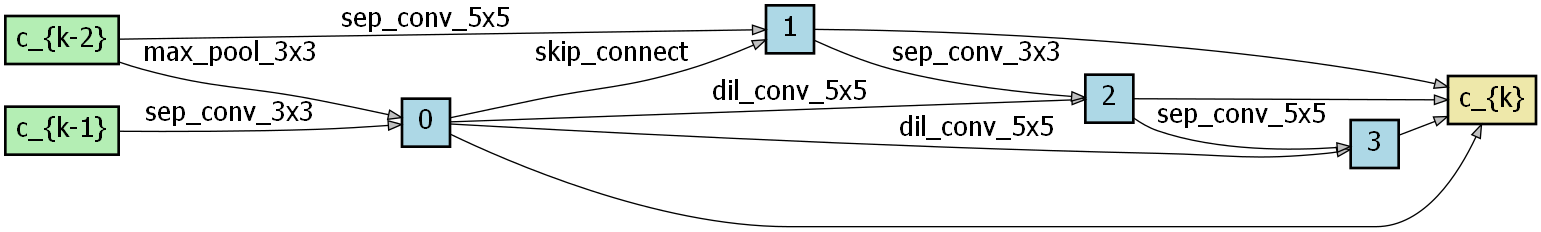}%
	}
	\subfloat[Reduction Cell in Weight Sharing based Random$\_$Search$\_$Natural]{\includegraphics[width=2.35in]{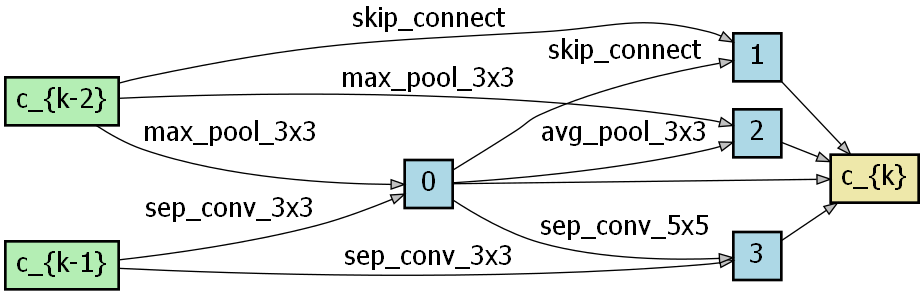}%
	}
	
	\subfloat[Normal Cell in Weight Sharing based Random$\_$Search$\_$System]{\includegraphics[width=2.35in]{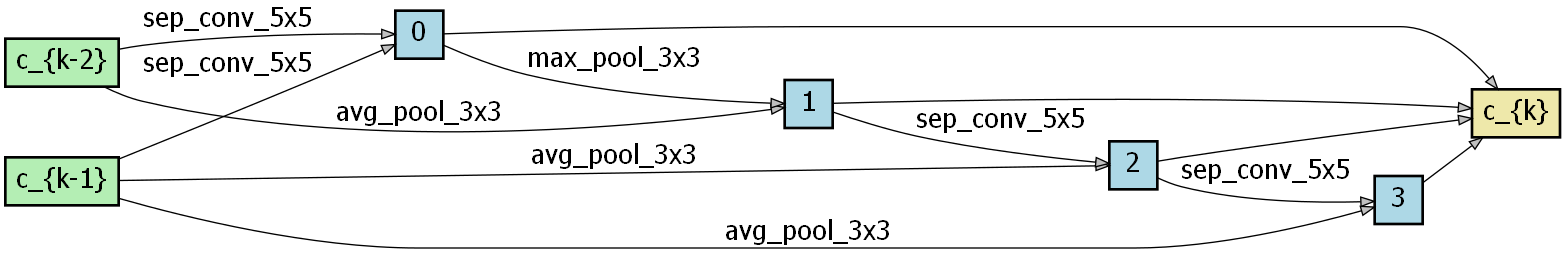}%
	}
	\hfil
	\subfloat[Normal Cell in Weight Sharing based Random$\_$Search$\_$Jacobian]{\includegraphics[width=2.35in]{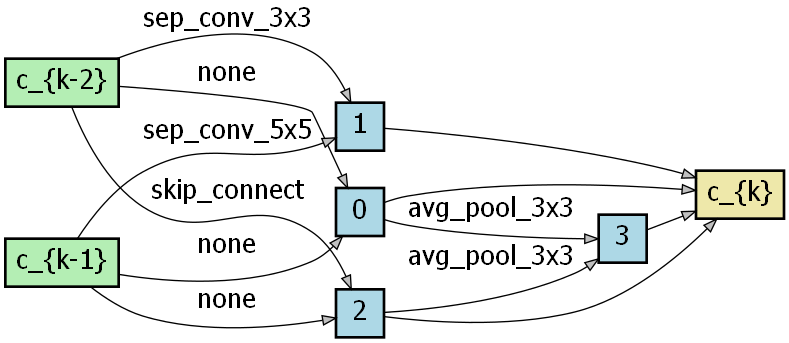}%
	}
	\subfloat[Normal Cell in Weight Sharing based Random$\_$Search$\_$Hessian]{\includegraphics[width=2.35in]{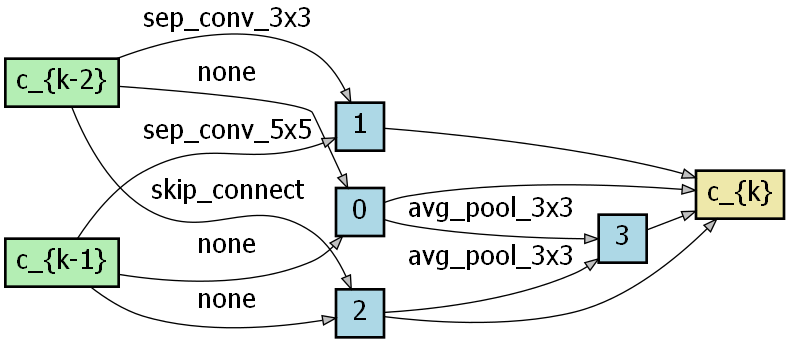}%
	}
	
	\subfloat[Reduction Cell in Weight Sharing based Random$\_$Search$\_$System]{\includegraphics[width=2.35in]{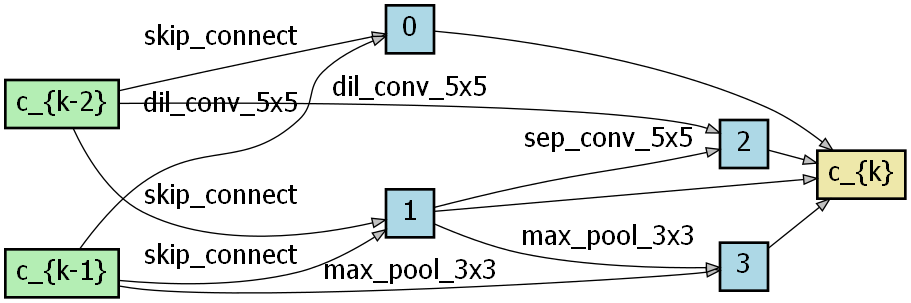}%
	}
	\hfil
	\subfloat[Reduction Cell in Weight Sharing based Random$\_$Search$\_$Jacobian]{\includegraphics[width=2.35in]{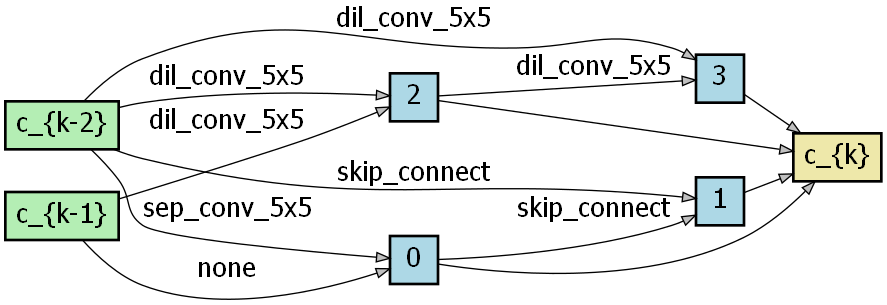}%
	}
	\subfloat[Reduction Cell in Weight Sharing based Random$\_$Search$\_$Hessian]{\includegraphics[width=2.35in]{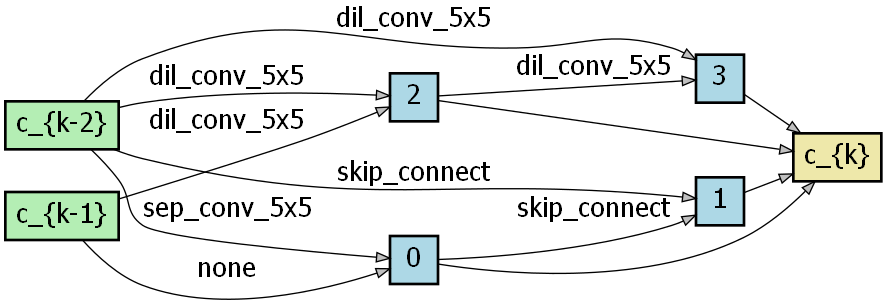}%
	}
	\caption{The visualization of the searched architectures using Weight-sharing based Random Search under different robustness evaluation}
	\label{arch7}
\end{figure*}

\end{document}